\documentclass[10pt,twocolumn,letterpaper]{article}

\usepackage[pagenumbers]{cvpr} 

\usepackage{siunitx}

\definecolor{cvprblue}{rgb}{0.21,0.49,0.74}
\usepackage[pagebackref,breaklinks,colorlinks,allcolors=cvprblue]{hyperref}

\usepackage{url}
\usepackage{booktabs}
\usepackage{multirow}
\usepackage{xcolor}
\usepackage{colortbl}
\usepackage{subcaption}
\usepackage{makecell}

\usepackage[many]{tcolorbox}
\tcbset{
  colback=gray!3,
  colframe=black!40,
  colbacktitle=gray!10,
  coltitle=black,        
  boxrule=0.4pt,
  arc=1pt,
  outer arc=1pt,
}

\usepackage{siunitx}
\usepackage[table]{xcolor}
\usepackage{enumitem} 

\usepackage{dblfloatfix}   

\definecolor{twboxbg}{HTML}{F6FAFF}
\definecolor{twboxframe}{HTML}{2F6FED}

\makeatletter
\newsavebox{\tw@box}
\newenvironment{takeawaybox}[1][Key Takeaways]{%
  \par\noindent\setlength{\fboxsep}{8pt}%
  \begin{lrbox}{\tw@box}%
    \begin{minipage}{\dimexpr\linewidth-2\fboxsep-2\fboxrule\relax}%
    \small 
      \textbf{#1}\par\vspace{0.35em}%
}{%
    \end{minipage}%
  \end{lrbox}%
  \fcolorbox{twboxframe}{twboxbg}{\usebox{\tw@box}}%
  \par\vspace{0.9em}%
}
\makeatother

\definecolor{latentblue}{HTML}{E6F2FF}

\newcommand{\survival}{\raisebox{-0.3ex}{\includegraphics[height=1.2em]{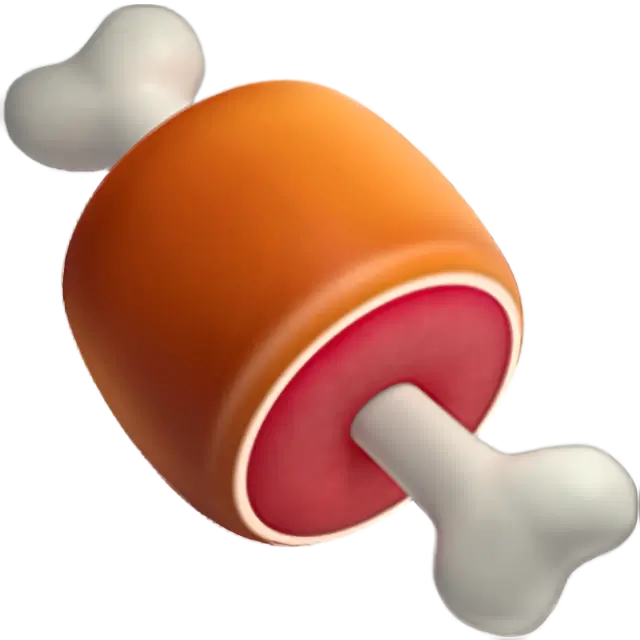}}}
\newcommand{\curiosity}{\raisebox{-0.3ex}{\includegraphics[height=1.2em]{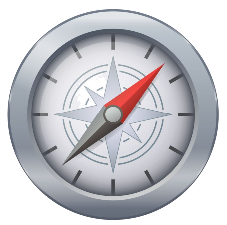}}}
\newcommand{\utility}{\raisebox{-0.3ex}{\includegraphics[height=1.2em]{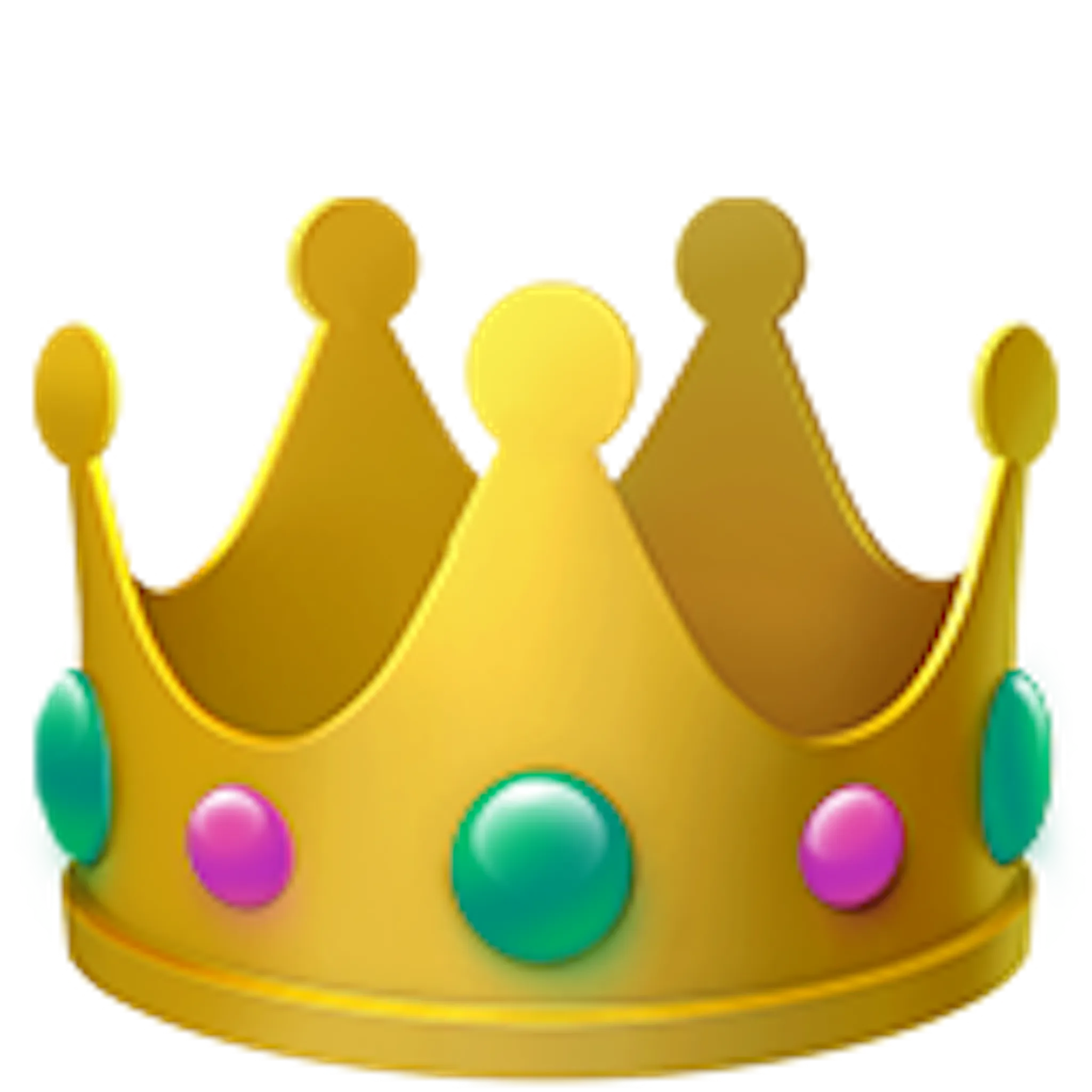}}}

\definecolor{colControl}{HTML}{FDFBE5}
\definecolor{colIL}{HTML}{FEF1E6}
\definecolor{colPPO}{HTML}{FFE8E9}
\definecolor{colDQN}{HTML}{F8DFF3}
\definecolor{colDreamer}{HTML}{E7DDF8}
\definecolor{colVJEPA2}{HTML}{D1E2F9}
\definecolor{colGenie}{HTML}{c6edf9}
\definecolor{col4o}{HTML}{cbfaf0}
\definecolor{col5}{HTML}{dafdde}
\definecolor{colQwen}{HTML}{fdfbe5}
\definecolor{colIPR}{HTML}{fef1e6}

\def\paperID{1816} 
\def\confName{CVPR}
\def\confYear{2026}

\title{IPR-1: Interactive Physical Reasoner}

\author{
    Mingyu Zhang$^{1,*}$ \quad Lifeng Zhuo$^{1,*}$ \quad Tianxi Tan$^{1}$ \quad Guocan Xie$^1$ \quad Xian Nie$^1$ \quad Yan Li$^1$ \\ Renjie Zhao$^{1,**}$ \qquad Zizhu He$^1$ \qquad Ziyu Wang$^1$ \qquad Jiting Cai$^{1,2,**}$ \qquad Yong-Lu Li$^{1,\dagger}$ \\
    RHOS \\
    $^1$Shanghai Jiao Tong University \qquad 
    $^2$Carnegie Mellon University \\
    {\tt\small sjtuzmy2003@sjtu.edu.cn \qquad yonglu\_li@sjtu.edu.cn}
}

\begin{document}
\maketitle

\begingroup
\renewcommand\thefootnote{}\footnotetext{
$^*$Equal contribution. \quad
$^\dagger$Corresponding author. 

$^{**}$Conducted during an internship at Shanghai Jiao Tong University.
}
\addtocounter{footnote}{-1}
\endgroup

\begin{abstract}
Humans learn by observing, interacting with environments, and internalizing physics and causality. Here, we aim to ask whether an agent can similarly acquire human-like reasoning from interaction and keep improving with more experience.
To study this, we introduce a \emph{Game-to-Unseen} (G2U) benchmark of 1,000+ heterogeneous games that exhibit significant visual domain gaps.
Existing approaches, including VLMs and world models, struggle to capture underlying physics and causality since they are not focused on core mechanisms and overfit to visual details.
VLM/VLA agents reason but lack look-ahead in interactive settings, while world models imagine but imitate visual patterns rather than analyze physics and causality.
We therefore propose \textbf{IPR} (\textbf{Interactive Physical Reasoner}), using world-model rollouts to score and reinforce a VLM’s policy, and introduce \textbf{PhysCode}, a physics-centric action code aligning semantic intent with dynamics to provide a shared action space for prediction and reasoning.
Pretrained on 1,000+ games, our IPR performs robustly on levels from primitive intuition to goal-driven reasoning, and even surpasses GPT-5 overall. 
We find that performance improves with more training games and interaction steps, and that the model also zero-shot transfers to unseen games.
These results support physics-centric interaction as a path to steadily improving physical reasoning. 
Further demos and project details can be found at \url{https://mybearyzhang.github.io/ipr-1}.
\end{abstract}    
\section{Introduction}

Humans do not learn physics and causality from labels; we earn them through \emph{interaction}.
As experience accumulates with age, our prediction sharpens, our reasoning stabilizes, and our abilities scale.
This motivates a central question for embodied AI: \emph{what learning paradigm enables human-like reasoning to learn through interactive experience, and to improve steadily with more interaction?}

\begin{figure}
    \centering
    \includegraphics[width=\linewidth]{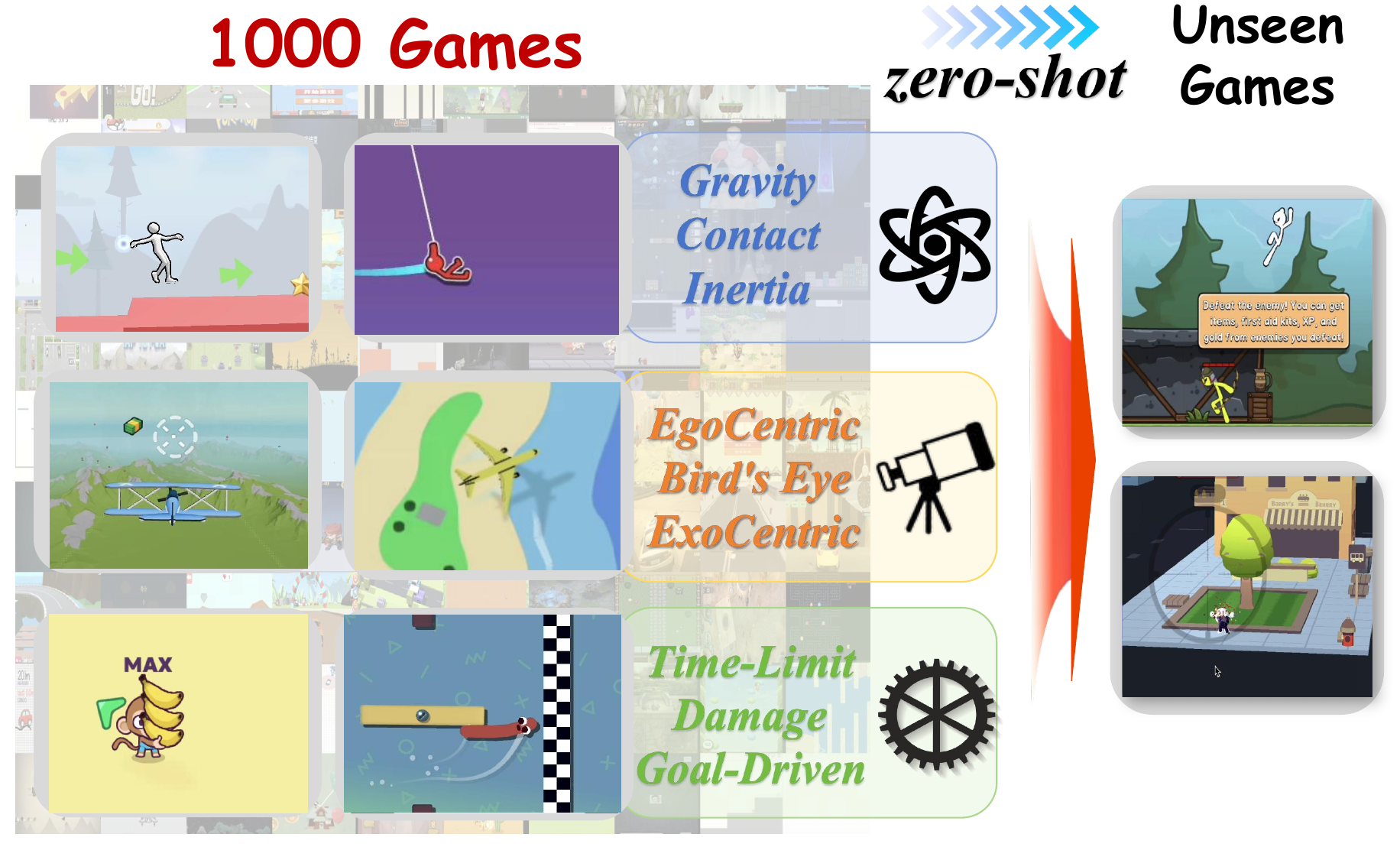}
    \caption{\textbf{Game-to-Unseen (G2U) problem.} Humans accumulate interactive experience and rapidly adapt to new games. Despite different visuals and interfaces, many games share underlying physical/causal mechanisms. We pretrain on 1{,}000+ visually and physically diverse games to test whether an agent can internalize these shared mechanisms and generalize to \emph{unseen} games.}
    \vspace{-5pt}
    \label{fig:teaser}
\end{figure}

\begin{figure*}[t]
    \centering
    \includegraphics[width=0.7\linewidth]{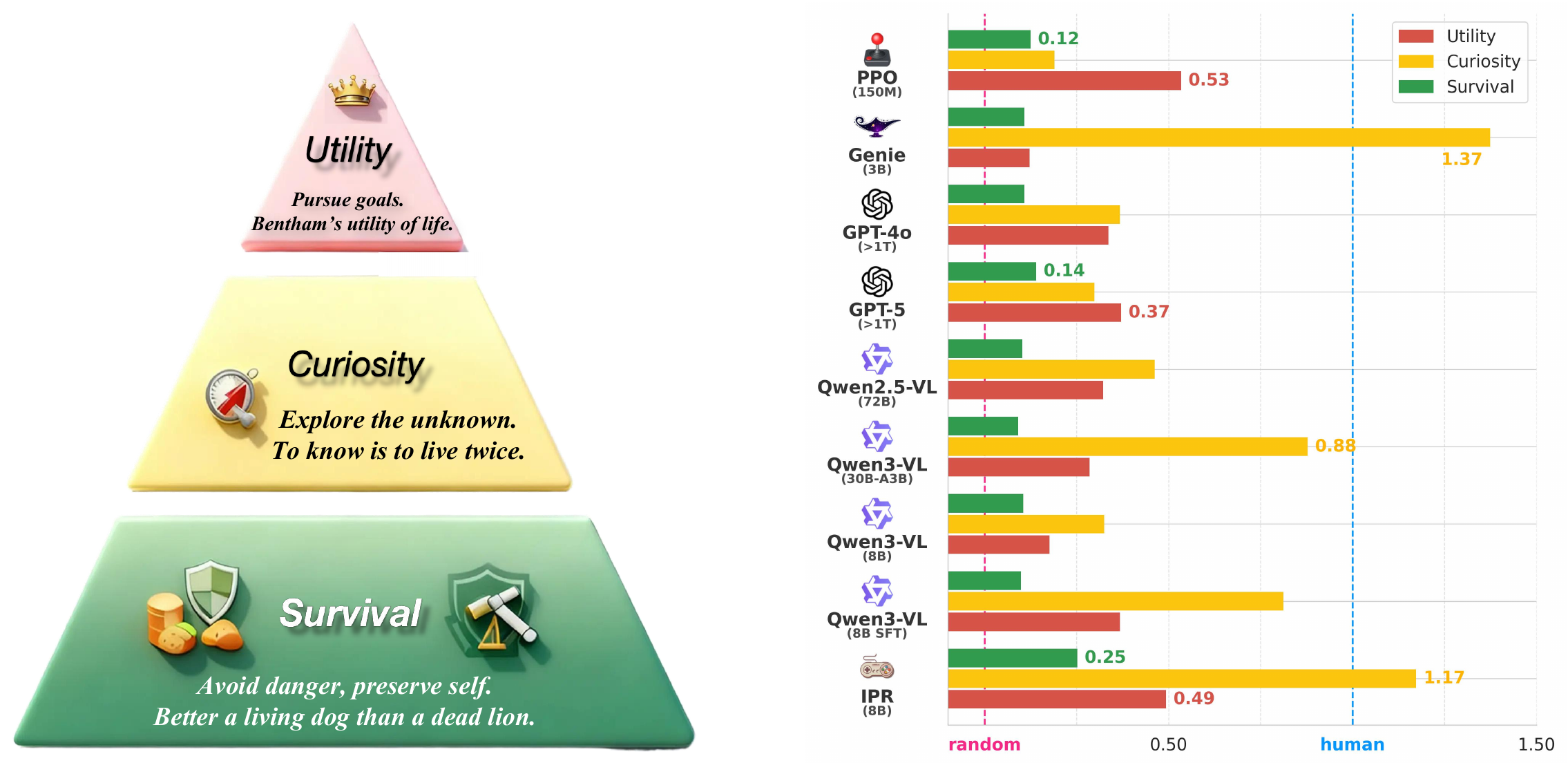}
    \vspace{-5pt}
    \caption{
    \textbf{Three-level evaluation inspired by Maslow's hierarchy of needs.}
    We organize tasks into a pyramid of Survival, Curiosity, and Utility.
    \textbf{Survival} measures how long the agent can stay alive by avoiding risks.
    \textbf{Curiosity} measures how broadly it visits novel states;
    and \textbf{Utility} measures how well it achieves downstream goals.
    The three levels progress from physical intuition to goal-driven reasoning.
    Our IPR performs robustly across the entire pyramid.
    }
    \label{fig:pyramid}
\end{figure*}

We assume that, if an agent is exposed to \emph{diverse, interactive worlds} and trained to distill \emph{shared physical and causal mechanisms}, rather than domain-specific appearance or action interfaces, 
it would scale its physical reasoning ability reliably and \emph{transfer} to new scenarios.
This view resonates with prior reasoning works~\cite{Bengio+chapter2007,huang2023reasoninglargelanguagemodels,zhang2024stepbackrethinkingstages}.
Pretrained VLMs~\cite{hu2024survey, wang2025gametarspretrainedfoundationmodels}, despite broad semantic priors from large-scale pretraining, mainly acquire static, pattern-matching behaviors as open-loop planning; SFT strengthens text-based declarative reasoning but not the predictive grounding required for interactive physical tasks. 
Behavior-cloned VLAs~\cite{wang2024jarvis, bu2025univlalearningacttaskcentric} are upper-bounded by the quality of demonstrations; relying on rote memorization of trajectories, they lack adaptability and prone to failure under environmental variations.
Model-based approaches~\cite{wiener2019cybernetics, 
moerland2023model}, including control theory and model-based RL, ensure stability with accurate dynamics but struggle in complex environments where faithful modeling is infeasible.
Model-free RL~\cite{yu2020meta, osband2016deep} avoids explicit modeling but demands massive samples and dense rewards. It often overfits to \emph{task-specific shortcuts} rather than causal mechanisms, hindering transfer to complex tasks.
Recent world models and other prediction-based approaches~\cite{bruce2024genie,assran2025v,hafner2025trainingagentsinsidescalable} scale effectively by learning latent dynamics and enabling agents to imagine futures.
They can optimize actions interactively toward goal-aligned representations, but in practice they often collapse into short-horizon target chasing or imitation of surface correlations, lacking robust causal reasoning and suffering from compounding errors in complex environments.

Collectively, these limitations highlight a fundamental gap: while existing paradigms exhibit partial success, they tend to overfit to superficial visual details rather than capturing the underlying physical and causal mechanisms. Appropriately approximating these invariant dynamics, which are pivotal for robust transfer across interactive environments, requires leveraging diverse domains to disentangle core mechanisms from visual appearance.
Reinforcement Learning (RL) excels at optimizing by interaction but relies on sparse, task-entangled signals that hinder generalization; Generative World Models capture dynamics but often \emph{over-models} the full sensory space; and VLMs, despite offering rich semantic priors, lack the predictive grounding required for precise physical consistency. 
This motivates a ``blended'' perspective: instead of committing entirely to exploration (RL), full-scene prediction (world models), or static priors (VLMs), we should reconsider the ratio to absorb these components. 
To operationalize this, our approach aligns with the Latent World Model paradigm~\cite{assran2025v}, which serves as the structural backbone to integrate these strengths. 
Specifically, a scalable reasoner should (i) model only the \textbf{essential latent dynamics} necessary for anticipating consequences---discarding high-fidelity pixel reconstruction; (ii) interact with raw multimodal signals through a policy enriched by \textbf{VLM-based semantic priors}; and (iii) reinforce this policy using predictive feedback that reflects physical feasibility. 
By shifting the prediction target from raw observations to abstract representations, the system filters out task-irrelevant perceptual noise, allowing the agent to capture the ``essence'' of physical and causal mechanisms rather than the ``appearance'' of the world.

In this way, we propose \emph{IPR} (\textbf{I}nteractive \textbf{P}hysical \textbf{R}easoner), a paradigm where world model \emph{prediction} reinforces a VLM policy to adapt its physical reasoning in interactive environments (Fig.~\ref{fig:pipeline}).
To evaluate this paradigm at scale, we curate over 1,000 heterogeneous games spanning diverse visual styles, control interfaces, physics configurations, and causal structures.
Games provide an ideal testbed for physical reasoning: they offer rich interaction, realistic physics, and effectively \emph{unlimited} rollouts at low cost.
Crucially, their heterogeneous visual appearances introduce substantial domain gaps that typically break traditional agents trained environment-by-environment.
For IPR, however, these diverse worlds share the same underlying physical and causal principles, allowing it to learn a representation focused enough to transfer across radically different domains.

\begin{figure}[t]
    \centering
    \includegraphics[width=\linewidth]{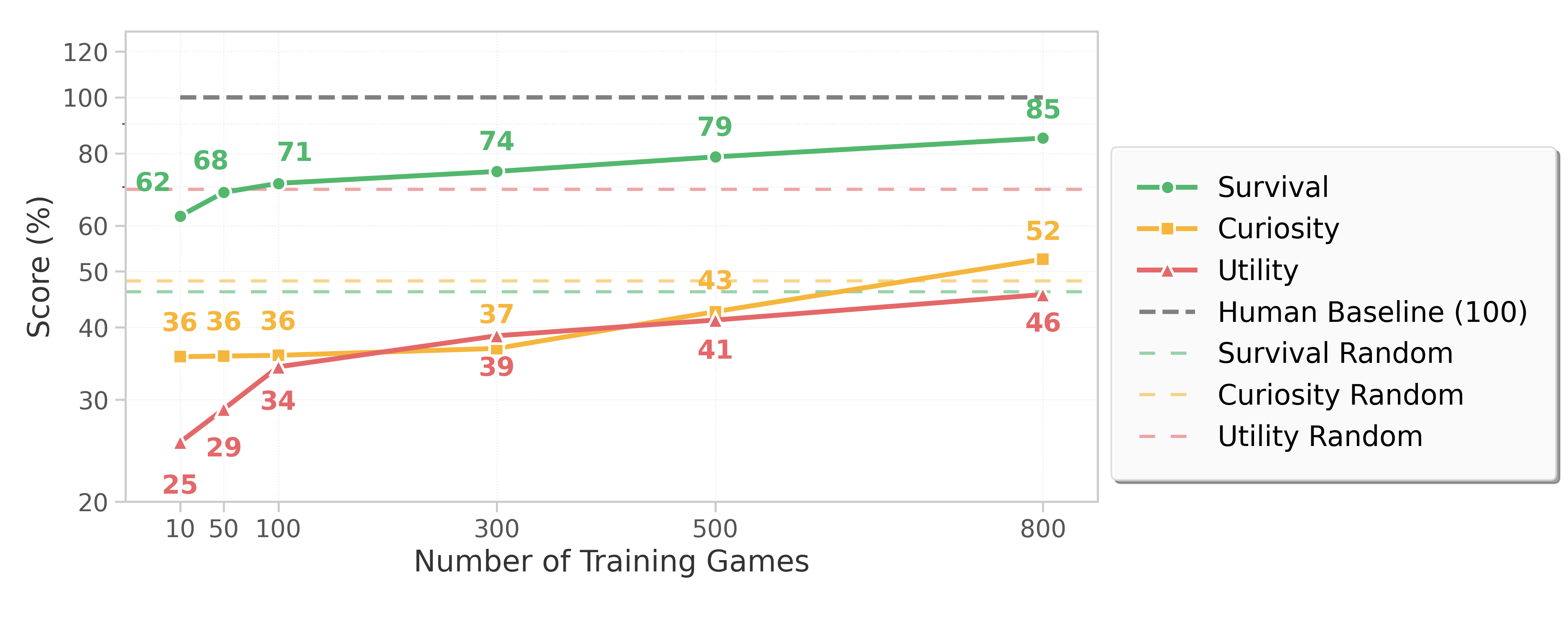}
    \caption{\textbf{G2U zero-shot scaling on 50 held-out games.}
    As the number of training games $N$ increases, zero-shot performance on \survival, \curiosity, and \utility~improves steadily on the unseen set $\mathcal{T}_{\mathrm{U}}$.}
    \label{fig:scaling}
\end{figure}

We further organize evaluation into three levels inspired by Maslow’s hierarchy of needs~\cite{huitt2007maslow}: \emph{Survival}, \emph{Curiosity}, \emph{Utility}, covering a spectrum from physical intuition to goal-directed reasoning (Fig.~\ref{fig:pyramid}).
The result on three levels verifies two failure modes: reasoning-based VLM/VLA lack forward consequence prediction to explore (Curiosity), while prediction-based world models explore broadly yet fail at goal-driven tasks (Utility).
Across the full suite, our \emph{IPR} remains robust on all three levels, while RL-based and prediction-based baselines often collapse on one or more of them.
With an 8B backbone, IPR even \emph{surpasses} GPT-5 overall.
Moreover, competence scales with the number of training games and interaction steps (Fig.~\ref{fig:scaling}) and \emph{zero-shot transfers} to novel environments, highlighting the potential of interactive learning for physical reasoning at scale.
We will further extend this paradigm to real-world interactive environments and perform on robotic tasks.

In general, our contributions are:
(1) We formulate the \textbf{G2U} problem and curate 1,000+ heterogeneous games with a hierarchical evaluation (\emph{Survival/Curiosity/Utility}), diagnosing the strengths and weaknesses of prevalent prediction-based, RL-based, and VLM-based methods. 
(2) We propose \textbf{IPR}: world-model rollouts \emph{score} and \emph{reinforce} VLM in the same action space, 
enabling interactive experience to steadily build up physical reasoning ability.
(3) We introduce \textbf{PhysCode}, a physics-centric action code fusing action semantics with visual dynamics, bridging WM prediction and VLM reasoning.
\section{Related Works}

\begin{figure*}[t]
    \centering
    \includegraphics[width=\linewidth]{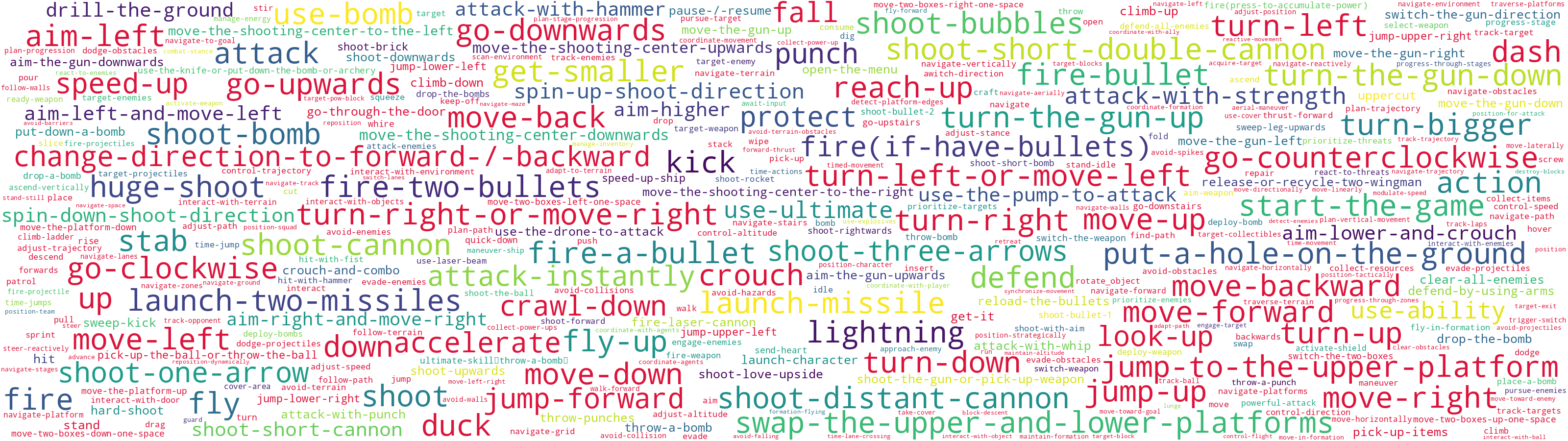}
    \caption{\textbf{Word cloud of action semantics across thousands of game worlds.} These shared semantics provide the structural foundation for cross-domain transfer. Actions highlighted in red represent those shared with general robotic operations, while the size of each word reflects its frequency in our data recipe.}
    \label{fig:wordcloud}
\end{figure*}

\paragraph{Action space discovery.}
Research on action spaces spans hand-designed controls, language-based interfaces, and learned latent representations. Early embodied agents operated over environment-specific key bindings, torques, or joystick signals~\cite{mnih2015human,schulman2017proximal,haarnoja2018softactorcriticoffpolicymaximum,brockman2016openaigym}, which offer precise control but entangle behavior with platform-specific layouts and hinder cross-domain transfer. A second line adopts \emph{language}-based action spaces, issuing natural-language commands or tool calls~\cite{shridhar2020alfredbenchmarkinterpretinggrounded,fan2022minedojobuildingopenendedembodied,wang2024jarvis,wang2023voyager,ahn2022icanisay}; while language affords semantic generality, it abstracts away timing, force, and perception–action couplings, often leading to imprecise or under-grounded control~\cite{savva2019habitat,reed2022generalist}.  
A complementary direction learns \emph{latent} action spaces directly from interaction data. Discrete or continuous latent codes, via VQ-VAE~\cite{oord2018neuraldiscreterepresentationlearning} or sequence models, have been explored for planning, control, and world models~\cite{bruce2024genie,chi2024diffusionpolicyvisuomotorpolicy,sharma2020emergentrealworldroboticskills,lynch2021languageconditionedimitationlearning,okada2021dreamingmodelbasedreinforcementlearning}. Recent VLM/VLA systems integrate such latent tokens into large multimodal models~\cite{sapkota2025vision,huang2024embodiedgeneralistagent3d}, but these codes often remain entangled across domains and lack mechanisms to capture shared physical principles versus environment-specific affordances.  
Our work addresses this gap by learning a \emph{physics-centric} latent action space that captures reusable dynamical patterns across games, instead of binding actions to domain-specific visuals and control layouts.
Fig.~\ref{fig:wordcloud} shows that different worlds share some semantic actions, validating the design of shared latent action space.

\begin{figure}[!t]
    \centering
    \includegraphics[width=\linewidth]{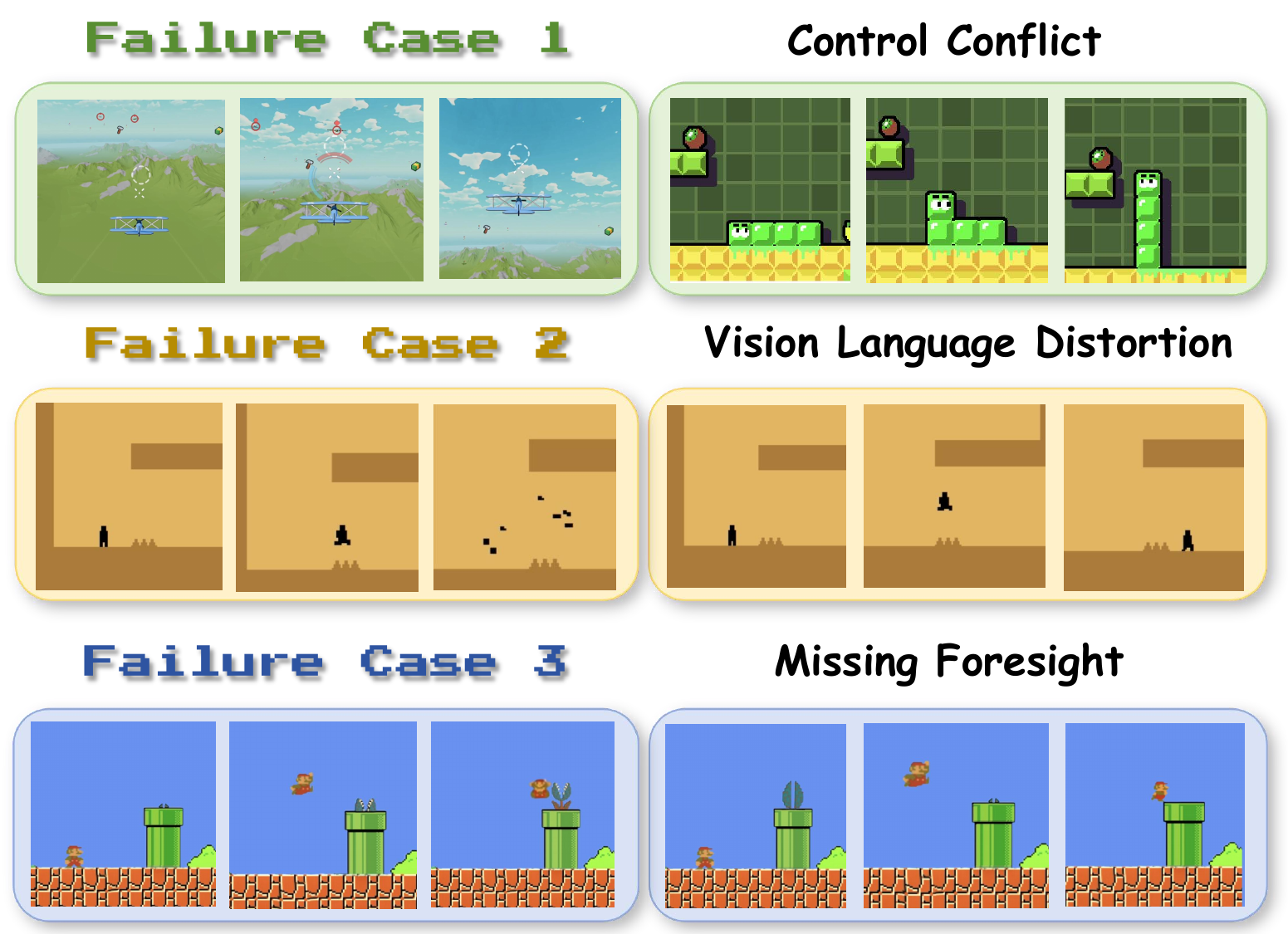}
    \caption{\textbf{Motivating failure cases in control semantics, language grounding, and prediction.}
    (1) \emph{Control conflict:} the same key (\eg, \texttt{UP}) triggers different semantics across games (camera tilt up v.s.\ character move up), causing console aliasing.
    (2) \emph{Vision-language distortion:} text-only actions cannot specify precise visual magnitudes (\eg, jump height/speed), leading to systematic amplitude errors.
    (3) \emph{Missing foresight:} without imagination, the agent cannot anticipate upcoming hazards during interaction (\eg, spikes, moving enemies).}
    
    \label{fig:failure}
\end{figure}

\paragraph{Agents in interactive environments.}
Research on game-playing agents has largely followed three threads. 
\textit{RL-based} agents, from DQN and PPO/SAC to large-scale systems like AlphaStar and OpenAI Five~\cite{mnih2013playing,vanhasselt2015deepreinforcementlearningdouble,schulman2017proximal,haarnoja2018softactorcriticoffpolicymaximum,Vinyals2019GrandmasterLI,openai2019dota2largescale}, learn policies directly from pixels and rewards and achieve strong title-specific performance, but remain sample-inefficient, brittle to interface changes, and struggle with long-horizon credit assignment and cross-game transfer. 
\textit{Prediction-based} (world-model) agents such as World Models, PlaNet, the Dreamer family, and Genie~\cite{https://doi.org/10.5281/zenodo.1207631,hafner2019learninglatentdynamicsplanning,hafner2020dreamcontrollearningbehaviors,hafner2023mastering,bruce2024genie} first learn latent dynamics and then plan or optimize in imagination, improving exploration and sparse-reward learning, yet degrade when learned dynamics or action semantics drift from the test environment and typically optimize task or pixel losses rather than reasoning quality. 
\textit{VLM/VLA-based} agents like Gato, RT-2, Voyager, MineDojo, and recent VLA frameworks~\cite{reed2022generalist,brohan2023rt2visionlanguageactionmodelstransfer,wang2023voyager,fan2022minedojobuildingopenendedembodied} cast acting as sequence modeling over images, text, and actions and excel at zero-shot instruction following, but rely heavily on static corpora, heuristic wrappers, and weakly grounded forward prediction (Fig.~\ref{fig:failure}).
Our IPR paradigm aims to inherit the strengths of these lines by using a physics-centric latent action space where a world model provides imagination-based value estimates and a reasoning VLM policy is reinforced through interactive experience in the \emph{same} latent space.

\paragraph{Benchmarks and evaluation.}
Interactive environments have long served as testbeds for learning control, exploration, and generalization: Atari/ALE provided dense stepwise rewards for RL training and evaluation~\cite{mnih2013playing,bellemare2013arcade}, while later platforms such as \textit{Minecraft}, \textit{VizDoom}, and \textit{StarCraft} introduced long-horizon goals, partial observability, and sparse rewards~\cite{kempka2016vizdoom,vinyals2017starcraft,fan2022minedojobuildingopenendedembodied,wang2023voyager}.
With the rise of VLM/VLA agents, web-based benchmarks and browser environments have been proposed to test generalization to novel tasks and interfaces~\cite{yuan2023rl,qi2024webrl}.
Following this line, we evaluate agents on a diverse suite of games and adopt simple game-agnostic metrics grouped into three levels: \emph{survival}, \emph{curiosity}, and \emph{utility}, to provide their preformance from physical intuition to reasoning and their scaling with experience.
\section{Preliminaries}
\label{sec:prelim}

\begin{figure*}[th]
    \centering
    \includegraphics[width=\linewidth]{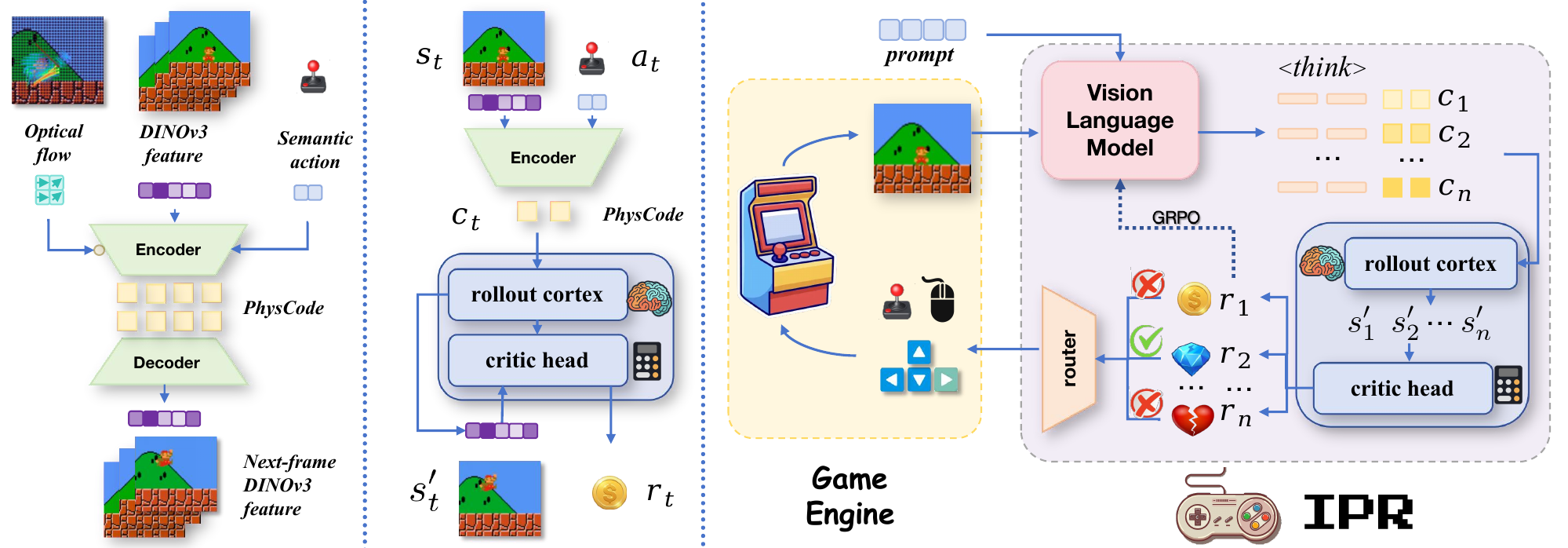}
\caption{\textbf{IPR training pipeline.}
\textbf{Stage~1: PhysCode pre-training.} Video clips with optical flow and action semantics are fed to a VQ-based latent action model to learn discrete codes (\textit{PhysCode}) that represent  dynamics.
\textbf{Stage~2: Latent-conditioned world model.} Given current features and PhysCode sequences, a world model is trained to predict future features and rewards under latent actions.
\textbf{Stage~3: Prediction-reinforced reasoning.} A VLM reasons over the scene and generates candidate PhysCode sequences. The world model rolls them out in imagination, and the predicted rewards/values are used to select the best actions and to optimize the VLM policy.}

    \vspace{-10pt}
    \label{fig:pipeline}
\end{figure*}

\subsection{Problem Setting}
We consider a family of interactive environments $\{\mathcal{E}_m\}_{m=1}^M$, each formalized as a POMDP:
\begingroup
\setlength{\abovedisplayskip}{6pt}
\setlength{\belowdisplayskip}{6pt}
\begin{equation}
\mathcal{M}_m=\big(\mathcal{S},\mathcal{A},T_m,R_m,\mathcal{O},\gamma;\,\varphi_m\big),
\end{equation}
\endgroup
where $\varphi_m$ are latent \emph{physics parameters} (\eg, gravity $g$, friction $\mu$, mass $M$). At time $t$, the environment emits an image $x_t\!\sim\!\mathcal{O}(\cdot\!\mid\! s_t)$, which we encode as $z_t=\phi_{\mathrm{enc}}(x_t)$; the agent executes $a_t\!\in\!\mathcal{A}$ and transitions according to
\begingroup
\setlength{\abovedisplayskip}{6pt}
\setlength{\belowdisplayskip}{6pt}
\begin{equation}
s_{t+1}\sim T_m\!\big(s_{t+1}\mid s_t,a_t;\varphi_m\big),\qquad r_t=R_m(s_t,a_t),
\end{equation}
\endgroup
where physics resides in $T_m$, and causality in $R_m$.

Control may use one of several interfaces $A\in\{\textsc{keyboard},\textsc{language},\textsc{latent}\}$;
a goal-conditioned VLM selects actions in the chosen space via
\begingroup
\setlength{\abovedisplayskip}{6pt}
\setlength{\belowdisplayskip}{6pt}
\begin{equation}
a_t^{(A)}\sim \pi_{\omega}^{(A)}\!\big(\cdot \mid z_t,\text{prompt}_t\big),\qquad a_t\equiv a_t^{(A)}\in\mathcal{A}.
\end{equation}
\endgroup
A feature-level world model $f_\theta$ then rolls out imagined futures under selected action sequences in the same action space $A$.
Given a horizon $H\in\mathbb{N}$, initialize $\hat z_t := z_t$ and choose an action sequence $\{a_{t+k}^{(A)}\}_{k=0}^{H-1}$. The rollout is defined by
\begingroup
\setlength{\abovedisplayskip}{6pt}
\setlength{\belowdisplayskip}{6pt}
\begin{equation}
\hat z_{t+k+1} = f_\theta\!\big(\hat z_{t+k},\, a_{t+k}^{(A)}\big),\quad k=0,1,\ldots,H-1,
\end{equation}
\endgroup
where $k$ indexes the step inside the imagined trajectory from time $t$ to $t+H$.

\subsection{PhysCode: Physics-centric Action Code}
\label{sec:physcode}

Motivated by the issues of raw-key semantic aliasing and the distortion of fine-grained visual dynamics when expressed in language, we propose
\textit{PhysCode}, a discrete latent action representation built on a VQ codebook $\mathcal{C}=\{v_k\}_{k=1}^K$. At step $t$, an action is a short code sequence $a_t^{\textsc{lat}}=\langle c_{t,1:L}\rangle$ with embedding obtained by looking up and pooling $\{v_{c_{t,\ell}}\}$.

Each code is conditioned on three cues: 
(i) \emph{domain-specific} visual appearance via DINOv3~\cite{siméoni2025dinov3} features $\phi_{\mathrm{img}}(x_t)$, 
(ii) \emph{domain-agnostic} motion via optical flow~\cite{fischer2015flownetlearningopticalflow} $\phi_{\mathrm{flow}}(\mathrm{Flow}(x_t,x_{t+1}))$, 
and (iii) lightweight semantic hints extracted by a T5 encoder~\cite{raffel2023exploringlimitstransferlearning}, with $\phi_{\mathrm{sem}}(y_t)=\mathrm{Enc}_{\text{T5}}(y_t)$.
Since natural language alone cannot express fine-grained dynamics (\eg, impulse magnitude, frictional slip), we rely on flow and visual features to carry these details while keeping semantics as guidance. By design, the resulting codes capture \emph{physics-relevant} intervention primitives that \emph{share} across environments with similar underlying physics and \emph{separate} when physics differ, enabling consistent reuse under matched physics and discrimination under shifted dynamics.

\section{Method}
\label{sec:method}

In this section, we introduce three components of \textbf{IPR} (Fig.~\ref{fig:pipeline}): 
(1) learning a \emph{physics-centric action code vocabulary} across diverse physical principles and causal mechanisms; 
(2) training a \emph{latent-conditioned world model} that predicts future features and rewards under sequences of latent actions; 
and (3) \emph{reinforcing VLM with world model rollout prediction} in the interactive environment, using aligned latent action code.
In inference, the VLM proposes candidate latent actions, queries the world model for short-horizon imagination and value estimates to score them, and executes the highest-scoring action.

\paragraph{Inducing the latent action vocabulary.}
Using the cues in Sec.~\ref{sec:physcode} (DINOv3 appearance $f_t,f_{t+\Delta}$, optical flow $u_t$, and lightweight semantics $e_t$), a small gated fusion module forms a fused representation $h_t$.
A spatio-temporal encoder $E_\psi$ maps $h_t$ to a continuous code $z_t$, which is vector-quantized to an index $a_t\!\in\!\{1,\ldots,K\}$ with codebook $\mathcal{C}=\{c_k\}_{k=1}^K$, and a decoder $D_\psi$ predicts the future feature $\hat f_{t+\Delta}$ from $(f_t,c_{a_t})$.
We train with a standard VQ-VAE objective
\begin{equation}
\begin{aligned}
\mathcal{L}_{\mathrm{LA}}
&= \big\|\hat f_{t+\Delta}-f_{t+\Delta}\big\|_2^2 \\
&\quad + \beta\big\|\mathrm{sg}[z_t]-c_{a_t}\big\|_2^2
+ \gamma\big\|z_t-\mathrm{sg}[c_{a_t}]\big\|_2^2,
\end{aligned}
\end{equation}
augmented with modality dropout on flow and a mild gate-sparsity regularizer to avoid over-reliance on optional cues.
Since optical flow is only available during pretraining, it acts as privileged information that helps shape a physics-centric codebook, while dropout and gate sparsity distill this structure into an encoder that, at test time, relies only on appearance and semantic cues.
At inference, we disable the flow gate and reuse the same encoder to obtain $z_t$ and its quantized index $a_t$ from appearance+semantics only.
The resulting discrete vocabulary yields temporally predictive tokens that cluster under matched physics and separate under different dynamics, providing a shared interface for VLM reasoning and world-model prediction.

\paragraph{Training the latent-level world model with a critic.}
With the latent action vocabulary fixed, we train a feature-level world model to predict future features conditioned on latent actions, replacing raw controls with their \textit{PhysCode} indices. For triples \((f_t,a_t,f_{t+\Delta})\), we embed \(a_t\) to \(e_{a_t}\) and compute
\begin{equation}
(\hat f_{t+\Delta},\, V_\theta(f_t,a_t)) \;=\; P_\theta\big(f_t,\, e_{a_t}\big).
\end{equation}
We predict in the \textit{latent space}, since features compress appearance variance and rendering noise, making dynamics more shareable across games. 
Concretely, we first train the world model with a feature-prediction loss
$
\mathcal{L}_{\text{pred}}
= \big\|\hat f_{t+\Delta}-f_{t+\Delta}\big\|_1,
$
and then learn a critic head with a Q-learning–style objective
$
\mathcal{L}_{\text{value}}
= \ell_{\text{Q}}\!\big(V_\theta(f_t,a_t),\, y_t\big),
$
where \(y_t\) is a target value computed from rollout returns via standard TD backups.

\paragraph{Prediction-reinforced interactive reasoning.}
We strengthen interactive reasoning with prediction: a world model imagines rollouts, and a VLM plans in the same latent action space. 
We adopt Qwen3-VL-8B~\cite{yang2025qwen3} as the backbone and extend its tokenizer with \textit{PhysCode} tokens so the VLM can directly emit discrete latent actions while preserving its language ability.

We first align perception and action by supervised training on $(f_t,c_t)$ pairs, where $f_t$ is the DINOv3 feature of the current frame and $c_t$ the latent action learned in Stage~1. 
Given the current context and goal $g$, the VLM samples $B$ candidate \textit{PhysCode} sequences $\{\mathbf{a}^{(b)}\}_{b=1}^B$, and the world model runs short-horizon imagined rollouts to assign each a predicted return, from which we compute advantages $A^{(b)}$. 
We then update the policy with GRPO~\cite{shao2024deepseekmath}:
\begin{equation}
\mathcal{L}_{\text{GRPO}}
=
\frac{1}{B}\sum_{b=1}^{B}
A^{(b)}\log \pi_\phi(\mathbf{a}^{(b)}\!\mid f_t,g)
\;-\;
\beta\,\mathrm{KL}\!\big(\pi_\phi \,\|\, \pi_0\big),
\end{equation}

In inference, the VLM proposes latent action candidates, the world model scores and prunes them via short-horizon rollouts, and a router $T_{\text{env}}$ maps the selected \textit{PhysCode} to environment controls.
Through repeated interaction under this prediction-in-the-loop scheme, the experience collected from imagined and executed trajectories reinforces the VLM, improving its physical reasoning in interactive environments.

\section{Experiments}
\label{sec:exp}

In this section, we aim to answer three questions: 
(1) Why is PhysCode necessary compared with raw keyboard inputs or language instructions?
(2) How would world model prediction reinforce VLM reasoning?
(3) Would IPR show scaling potential to transfer to unseen games?

\subsection{Setup: Datasets, Tasks, and Metrics}
\label{sec:exp:setup}

\paragraph{Sources.}
We curate a multi-source benchmark covering \textbf{863} open-source retro titles (via \texttt{stable-retro}~\cite{stable-retro}), \textbf{134} lightweight HTML/Canvas games, and \textbf{3} commercial games. 
This breadth exposes agents to heterogeneous visuals, action interfaces, and underlying physics/causal mechanisms, encouraging models to capture shared physical–causal regularities rather than overfit to domain-specific biases.

\begin{figure}
    \centering
    \includegraphics[width=\linewidth]{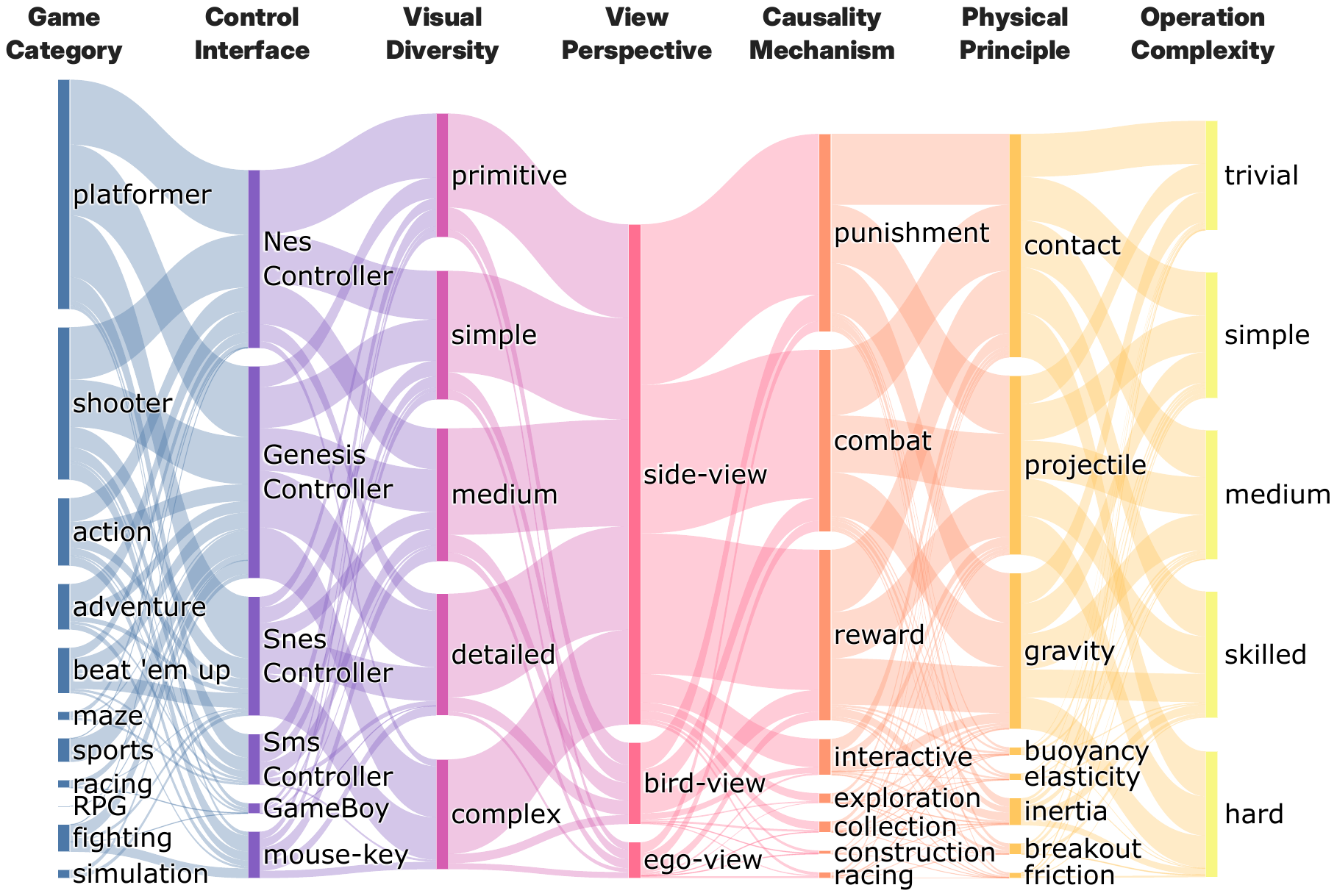}
    \caption{\textbf{Game data distribution.} 
Our dataset spans over 1,000 games categorized by \textit{game category}, \textit{control interface}, \textit{operation and visual complexity}, \textit{physical and causal mechanisms}. 
This wide coverage enables agents to experience diverse domains and learn transferable physical and causal understanding.}
    \label{fig:data_distribution}
\end{figure}

\paragraph{Diversity axes.}
We characterize each environment along seven axes to enable structured generalization analysis: 
\emph{(1) Game category}, with emphasis on physical interaction (e.g., platformer, shooter, sports); 
\emph{(2) Control interface}, such as GameBoy–style discrete keys, keyboard–mouse combinations, and high-dimensional hybrids; 
\emph{(3) Visual complexity}, ranging from low-resolution pixel art to high-fidelity 3D; 
\emph{(4) View perspective}, \eg ego-centric, top-down, and side views;
\emph{(5) Causal mechanism}, \eg damage/health dynamics, collection, punishment;
\emph{(6) Physical principle}, \eg gravity, contact, and inertia; 
\emph{(7) Operational difficulty}, approximated by the entropy and frequency of human control actions, reflecting how precisely and how often players must operate to succeed; 
Fig.~\ref{fig:data_distribution} summarizes the distributions over sources, game types, and these axes; detailed per-environment statistics are provided in the \textit{supplementary}.

\begin{table*}[t]
\centering
\caption{\textbf{PhysCode validation.} 
\textbf{Left:} Joint training across heterogeneous-physics games reveals cross-game conflicts for keyboard/mouse; language partially alleviates this via semantics, while \textit{PhysCode} separates actions by dynamics, reducing interface aliasing and showing minimal degradation under physics shifts. 
\textbf{Middle:} Leave-$n$-out transfer: training on all but 10 titles and evaluating zero-shot on the held-out set, \textit{PhysCode} transfers more reliably than keyboard or language interfaces. \textbf{Right:} Physics-conditioned transfer: zero-shot performance is relatively higher when target environments \emph{match} the training set’s physical mechanisms, indicating that \textit{PhysCode} captures reusable physical principles rather than game-specific bindings. 
}

\label{tab:exp1_all}

\begin{subtable}[t]{0.29\textwidth}
\centering
\caption{Confusion test for joint training.}
\label{tab:exp1:conf}
\resizebox{\textwidth}{!}{
\begin{tabular}{lccc}
\toprule
\textbf{Latent-Predict} & \textbf{Cosine} $\uparrow$ & \textbf{MSE} $\downarrow$ & \textbf{L1} $\downarrow$ \\
\midrule
Ad-hoc   & 0.9939 & 0.0121 & 0.0495   \\
\hline
Keyboard & 0.9894 & 0.0211 & 0.0772   \\
Language & 0.9892 & 0.0216 & 0.0758   \\
\rowcolor{latentblue}
\textit{PhysCode} & \textbf{0.9919} & \textbf{0.0204} & \textbf{0.0737}   \\
\bottomrule
\toprule
\textbf{Pixel-Predict} & \textbf{FID} $\downarrow$ & \textbf{SSIM} $\uparrow$ & \textbf{PSNR} $\uparrow$ \\
\midrule
Ad-hoc   & 87.83 & 0.7062 & 23.86   \\
\hline
Keyboard & 110.9 & 0.6110 & 20.82   \\
Language & 82.51 & 0.6960 & 23.52   \\
\rowcolor{latentblue}
\textit{PhysCode}   & \textbf{80.35} & \textbf{0.7240} & \textbf{23.82}   \\
\bottomrule
\end{tabular}
}
\end{subtable}
\hfill
\begin{subtable}[t]{0.29\textwidth}
\centering
\caption{Leave-$n$-out transfer.}
\label{tab:exp1:trans}
\resizebox{\textwidth}{!}{
\begin{tabular}{lccc}
\toprule
\textbf{Latent-Predict} & \textbf{Cosine} $\uparrow$ & \textbf{MSE} $\downarrow$ & \textbf{L1} $\downarrow$ \\
\midrule
Pre-trained & 0.9856 & 0.0230 & 0.0846     \\
\hline
Keyboard & 0.9784 & 0.0430 & 0.1153   \\
Language & 0.9790 & 0.0418 & \textbf{0.1132}   \\
\rowcolor{latentblue}
\textit{PhysCode}   & \textbf{0.9798} & \textbf{0.0403} & 0.1212   \\
\bottomrule 
\toprule
\textbf{Pixel-Predict} & \textbf{FID} $\downarrow$ & \textbf{SSIM} $\uparrow$ & \textbf{PSNR} $\uparrow$ \\
\midrule
Pre-trained   & 127.3 & 0.7438 & 22.11   \\
\hline
Keyboard & 315.0 & 0.3340 & 12.46   \\
Language & 320.2 & 0.1670 & 9.389   \\
\rowcolor{latentblue}
\textit{PhysCode}   & \textbf{297.0} & \textbf{0.3533} & \textbf{13.04}   \\
\bottomrule
\end{tabular}
}
\end{subtable}
\hfill
\begin{subtable}[t]{0.4\textwidth}
\centering
\caption{Physics-conditioned transfer.}
\label{tab:exp1:phys} 
\includegraphics[width=\linewidth]{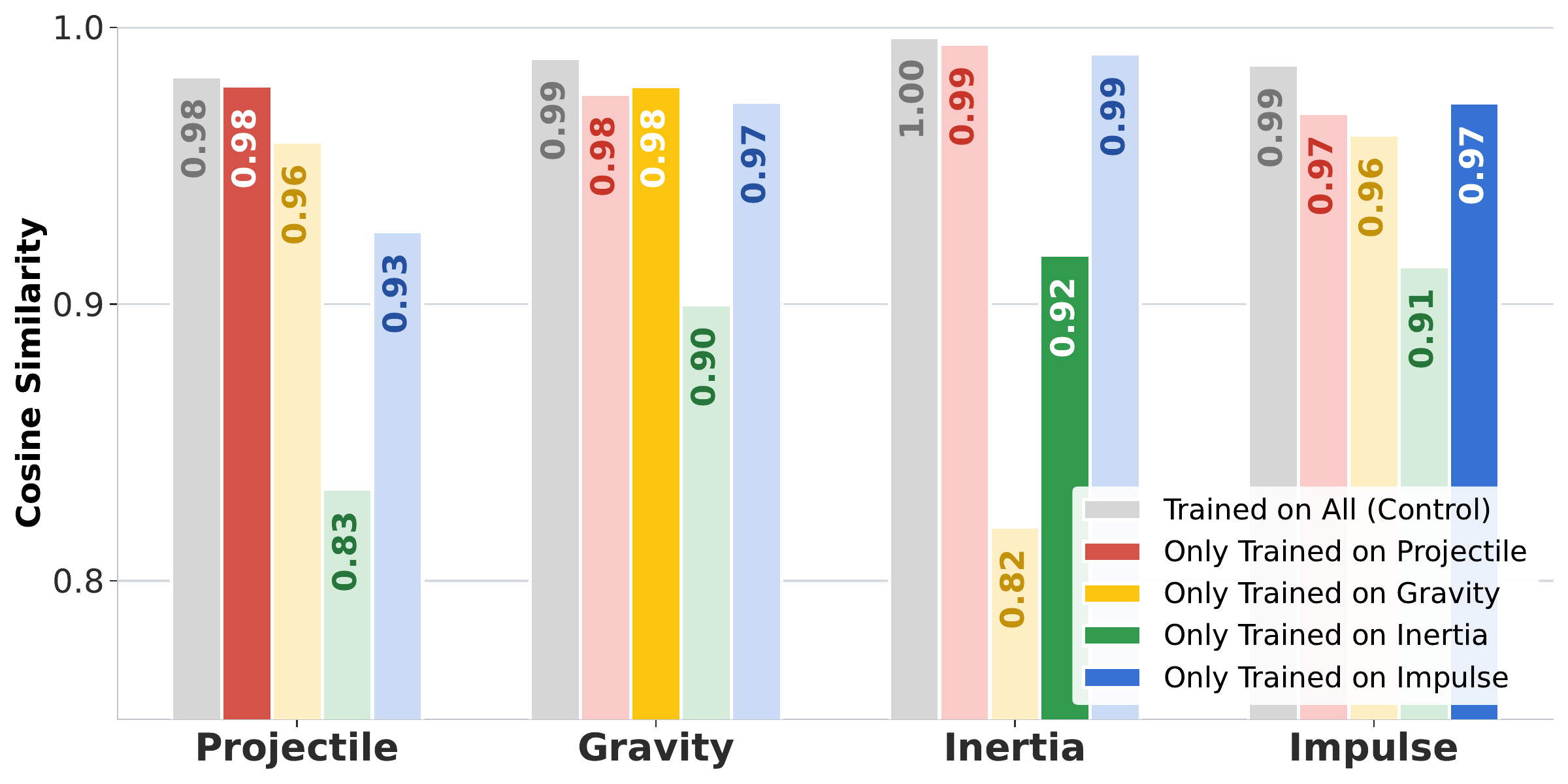}
\end{subtable}

\end{table*}

\paragraph{Data collection and preprocessing.}
Across the \textbf{1{,}000}-game corpus, we record human play at 60 FPS for 4 minutes per title and obtain per-game annotations covering \emph{physical principles}, \emph{causal mechanisms}, \emph{action semantics}, and \emph{game instructions}.
We perform a series of preprocessing, including normalizing time intervals, removing non-interactive segments, rebalancing extended idle/no-op periods, \etc. More details are in the \textit{supplementary}.

\paragraph{Hierarchical level design.}
Inspired by Maslow’s hierarchy of needs~\cite{huitt2007maslow}, we treat gameplay as a three-level progression: \emph{Survival} $\rightarrow$ \emph{Curiosity} $\rightarrow$ \emph{Utility} (Fig.~\ref{fig:pyramid}), from intuition to reasoning.

\textbf{\survival \emph{Survival.}}
The objective is to remain alive as long as possible, ignoring the original goal and avoiding risks. 
We report \emph{survival time} normalized per game,
\(H=\mathbb{E}[T]/T_{\text{typ}}\), where \(T\) is episode length (steps) and \(T_{\text{typ}}\) is a per-game reference horizon (\eg, median survival under a random policy).

\textbf{\curiosity \emph{Curiosity.}}
The goal is to visit \emph{novel states} like a baby to uncover regularities in the environment’s dynamics and causal mechanisms.
Following Magnipy~\cite{limbeck2024metric}, we embed frames with a pretrained CLIP visual encoder~\cite{radford2021learningtransferablevisualmodels}, compute the trajectory’s multi-scale \emph{metric-space magnitude} curve \(M(\tau)\), and define the exploration score as the area under this curve: $
E=\mathrm{AUC}(M(\tau)),
$
where larger \(E\) indicates broader state-space coverage.

\textbf{\utility \emph{Utility.}}
Utility measures how well an agent \emph{realizes Bentham's utility of life}~\cite{bentham1789morals}: devoting itself to goal completion with higher reward and shorter time.
We evaluate downstream goals according to the game types (completion, score, checkpoint time) and report the \emph{human-normalized score (HNS)}~\cite{bellemare2013ale} per game:
\begin{equation}
\mathrm{HNS}=\frac{m-m_{\text{rnd}}}{m_{\text{hum}}-m_{\text{rnd}}}\,,
\end{equation}
where \(m\) is the agent metric, \(m_{\text{rnd}}\) the random baseline, and \(m_{\text{hum}}\) human performance.

\subsection{Why is PhysCode Necessary}

We first investigate whether \textbf{PhysCode} is necessary compared with raw keyboard/mouse inputs and natural-language instructions.
First, we assess robustness under mixed-game joint training with heterogeneous physics (Tab.~\ref{tab:exp1:conf}), examining which action space best performs in diverse physical mechanisms and different console/game interfaces.
Second, we test transfer (Tab.~\ref{tab:exp1:trans}, Tab.~\ref{tab:exp1:phys}): a \emph{shared} PhysCode learned on source games improves zero-shot performance in unseen environments with \emph{matched} physics, demonstrating genuine physics grounding rather than interface memorization.

First, we examine how different action spaces
behave when trained jointly across a mixture of games with heterogeneous physics (Tab.~\ref{tab:exp1:conf}).
In this regime, raw keyboard/mouse inputs exhibit cross-game conflicts (the same key triggers different behaviors across environments).
Language interfaces partially alleviate this via explicit semantics.
\textit{PhysCode} separates actions by dynamics, reducing interface aliasing and showing minimal degradation under physics shifts.

Next, we ask whether sharing the latent space supports transfer.
In a leave-$n$-out protocol (Tab.~\ref{tab:exp1:trans}), we train on all but 10 games and evaluate zero-shot on the held-out titles.
We find that PhysCode transfers more reliably than keyboard or language instructions.

Moreover, we condition transfer on the physics of the environment.
We group games by their dominant physical mechanism, train under one principle (\eg, gravity), and evaluate zero-shot on held-out games with matching or different mechanisms.
When targets \emph{match} the training physics, zero-shot performance is \emph{typically} higher (Tab.~\ref{tab:exp1:phys}), with notable exceptions such as \textit{inertia}, which may already be covered by projectile/impulse.
This suggests that \textit{PhysCode} captures reusable physical mechanisms rather than game-specific bindings, even though our coarse physics taxonomy does not perfectly align with the agent’s internal abstractions.

\begin{table*}[th]
    \caption{\textbf{Comprehensive comparison across \survival, \curiosity, and \utility.}
    ``@'' denotes the optimization objective.
    Scores are normalized individually for each game, scaled between random ($0$) and human ($1$) benchmarks.
    \textbf{Mean} is the average of these normalized scores, indicating overall competence.
    \textbf{Avg.~Rank} is the average relative rank among 30 methods across all games (lower is better).
    \textbf{Ratio@Top-3(\%)} is the proportion of games where the method ranks within the top-3.
    Our IPR demonstrates robust performance across all metrics.}
    \label{tab:exp:main}
    \centering
    \scriptsize
    \vspace{-8.5pt}
    \resizebox{\linewidth}{!}{
\begin{tabular}{l c S[table-format=2.1] S[table-format=2.1] c S[table-format=2.1] S[table-format=2.1] c S[table-format=2.1] S[table-format=2.1] | S[table-format=2.1]}
\toprule
\multirow{3}{*}{\textbf{Methods}} & \multicolumn{3}{c}{\textbf{\survival Survival}} & \multicolumn{3}{c}{\textbf{\curiosity Curiosity}} & \multicolumn{3}{c}{\textbf{\utility Utility}} & \textbf{Overall} \\
\cmidrule(lr){2-4} \cmidrule(lr){5-7} \cmidrule(lr){8-10}

& \textbf{(Overall)} & \textbf{(Robustness)} & \textbf{(Competitiveness)} & \textbf{(Overall)} & \textbf{(Robustness)} & \textbf{(Competitiveness)} & \textbf{(Overall)} & \textbf{(Robustness)} & \textbf{(Competitiveness)} & \textbf{Avg.}\\

& \textbf{Mean $\uparrow$} & \textbf{Avg. Rank $\downarrow$} & \textbf{Ratio@Top-3(\%) $\uparrow$} & \textbf{Mean $\uparrow$} & \textbf{Avg. Rank $\downarrow$} & \textbf{Ratio@Top-3(\%) $\uparrow$} & \textbf{Mean $\uparrow$} & \textbf{Avg. Rank $\downarrow$} & \textbf{Ratio@Top-3(\%) $\uparrow$} & \textbf{Rank $\downarrow$}\\
\midrule
\rowcolor{gray!20}
\multicolumn{11}{c}{\textbf{Control Group}} \\
\midrule
\rowcolor{colControl}
Random                         & 0.000 & 16.2 &  6.7 & 0.000 & 18.1 &  3.0 & 0.000 & 12.3 & 12.8 & 26.9\\
\rowcolor{colControl}
Human                          & 1.000 &  5.7 & 46.3 & 1.000 &  7.9 & 14.0 & 1.000 &  2.9 & 61.6 & 2.8\\
\midrule

\rowcolor{gray!20}
\multicolumn{11}{c}{\textbf{Imitation Learning (IL) Group}} \\
\midrule
\rowcolor{colIL}
ACT-BC                         & 0.088 & 14.3 & 17.1 & 0.793 & 15.1 & 12.8 & 0.255 & 12.0 & 13.4 & 16.6\\
\rowcolor{colIL}
Qwen3-VL-8B-BC                 & 0.099 & 12.9 & 14.0 & 0.812 & 12.8 &  9.1 & 0.368 &  9.6 & 12.8 & 13.3\\
\midrule

\rowcolor{gray!20}
\multicolumn{11}{c}{\textbf{Reinforcement Learning (RL) Group}} \\
\midrule
\rowcolor{colPPO}
PPO@survival                   & 0.125 & 14.0 & 14.0 & 0.233 & 16.5 &  3.7 & \underline{0.588} & \underline{ 7.3} & \textbf{30.5} & 12.0\\
\rowcolor{colPPO}
PPO@curiosity                  & 0.114 & 14.9 & 11.6 & 0.190 & 17.3 &  2.4 & \textbf{0.609} & \textbf{ 6.9} & \underline{29.3} & 14.8\\
\rowcolor{colPPO}
PPO@utility                    & 0.120 & 15.0 & 12.2 & 0.220 & 16.8 &  3.0 & \textit{0.534} &  8.0 & 25.6 & 14.7\\
\rowcolor{colDQN}
DQN@survival                   & 0.121 & 14.4 & 15.9 & 0.856 & 14.4 &  8.5 & 0.497 & 10.8 & 15.2 & 12.2\\
\rowcolor{colDQN}
DQN@curiosity                  & \textit{0.131} & 13.2 & \textit{18.3} & 0.772 & 13.4 &  7.9 & 0.424 & 10.9 & 15.9 & 10.6\\
\rowcolor{colDQN}
DQN@utility                    & 0.125 & 13.7 & 16.5 & 0.620 & 14.2 &  4.9 & 0.445 & 10.8 & 17.1 & 11.4\\
\midrule

\rowcolor{gray!20}
\multicolumn{11}{c}{\textbf{World Model Group}} \\
\midrule
\rowcolor{colDreamer}
DreamerV3@survival             & 0.102 & 15.8 & 15.2 & 1.120 & \textit{12.5} & 16.5 & 0.298 & 11.3 & 16.5 & 13.1\\
\rowcolor{colDreamer}
DreamerV3@curiosity            & 0.108 & 14.5 & 17.7 & 1.161 & 13.1 & 14.0 & 0.235 & 10.0 & 20.1 & 10.7\\
\rowcolor{colDreamer}
DreamerV3@utility              & 0.097 & 14.9 & 17.7 & 0.964 & 15.4 & 11.0 & 0.139 & 11.4 & 18.3 & 15.4\\
\rowcolor{colVJEPA2}
V-JEPA2@survival               & 0.102 & 17.4 &  4.9 & 1.150 & 15.6 & \underline{17.7} & 0.191 & 13.9 & 16.5 & 18.3\\
\rowcolor{colVJEPA2}
V-JEPA2@curiosity              & 0.100 & 17.8 &  2.4 & \textbf{1.402} & 15.6 & 16.5 & 0.146 & 14.0 & 11.6 & 20.8\\
\rowcolor{colVJEPA2}
V-JEPA2@utility                & 0.102 & 17.5 &  1.8 & 1.136 & 14.5 & \textbf{22.6} & 0.152 & 14.1 & 11.6 & 20.2\\
\rowcolor{colGenie}
GenieRedux@survival            & 0.108 & 13.7 & 15.9 & 1.198 & \underline{12.5} & 11.0 & 0.128 & 12.7 & 12.8 & 14.2\\
\rowcolor{colGenie}
GenieRedux@curiosity           & 0.104 & 14.3 & 14.0 & \underline{1.374} & 12.5 &  9.8 & 0.100 & 12.8 & 12.8 & 16.1\\
\rowcolor{colGenie}
GenieRedux@utility             & 0.110 & 13.7 & 16.5 & \textit{1.248} & \textbf{12.4} & 14.6 & 0.122 & 13.5 & 14.6 & 12.4\\
\midrule

\rowcolor{gray!20}
\multicolumn{11}{c}{\textbf{Multimodal Large Language Model (MLLM) Group}} \\
\midrule
\rowcolor{col4o}
GPT-4o@survival                & 0.108 & \textit{12.6} & 13.4 & 0.039 & 17.2 &  0.6 & 0.302 &  9.2 & 19.5 & 16.4\\
\rowcolor{col4o}
GPT-4o@curiosity               & 0.079 & 16.8 & 11.6 & 0.368 & 15.3 &  5.5 & 0.186 & 10.6 & 17.7 & 19.4\\
\rowcolor{col4o}
GPT-4o@utility                 & 0.087 & 15.8 & 10.4 & 0.319 & 14.6 &  3.7 & 0.337 & 10.0 & 17.1 & 18.8\\
\rowcolor{col5}
GPT-5@survival                 & \underline{0.140} & \underline{10.5} & \underline{24.4} & 0.127 & 18.3 &  1.8 & 0.263 &  8.0 & 23.8 & 13.3\\
\rowcolor{col5}
GPT-5@curiosity                & 0.093 & 15.3 & 12.2 & 0.298 & 16.4 &  7.3 & 0.333 &  9.8 & 16.5 & 17.9\\
\rowcolor{col5}
GPT-5@utility                  & 0.108 & 15.2 & 11.0 & 0.185 & 16.5 &  0.6 & 0.371 & \textit{ 7.8} & \textit{26.2} & 16.8\\
\rowcolor{colQwen}
Qwen3-VL-30B-A3B@survival      & 0.091 & 14.3 & 11.0 & 0.325 & 23.0 &  0.0 & 0.289 & 12.0 & 14.0 & 22.7\\
\rowcolor{colQwen}
Qwen3-VL-30B-A3B@curiosity     & 0.086 & 15.8 & 11.6 & 0.878 & 20.5 &  2.4 & 0.155 & 11.7 & 15.2 & 22.4\\
\rowcolor{colQwen}
Qwen3-VL-30B-A3B@utility       & 0.108 & 13.5 & 12.2 & 0.528 & 21.3 &  4.9 & 0.285 & 11.6 & 14.6 & 17.6\\
\midrule

\rowcolor{gray!20}
\multicolumn{11}{c}{\textbf{Interactive Physical Reasoner}} \\
\midrule
\rowcolor{colIPR}
Qwen3-VL-8B w/o IPR            & 0.105 & 13.7 & 14.0 & 0.325 & 15.0 &  4.3 & 0.176 & 11.6 & 12.8 & 18.2\\
\rowcolor{colIPR}
\textbf{Qwen3-VL-8B w/ IPR}    & \textbf{0.252} & \textbf{ 2.6} & \textbf{72.0} & 1.173 & 13.1 & 13.4 & 0.493 &  8.5 & 22.0 & {\textbf{\hphantom{0}4.9}}\\
\rowcolor{colIPR}
(IPR ranking w/o control group)
  & \multicolumn{1}{c}{\scriptsize (\textbf{1}/28)}
  & \multicolumn{1}{c}{\scriptsize (\textbf{1}/28)}
  & \multicolumn{1}{c}{\scriptsize (\textbf{1}/28)}
  & \multicolumn{1}{c}{\scriptsize (\textbf{5}/28)}
  & \multicolumn{1}{c}{\scriptsize (\textbf{6}/28)}
  & \multicolumn{1}{c}{\scriptsize (\textbf{7}/28)}
  & \multicolumn{1}{c}{\scriptsize (\textbf{5}/28)}
  & \multicolumn{1}{c}{\scriptsize (\textbf{6}/28)}
  & \multicolumn{1}{c}{\scriptsize (\textbf{6}/28)}
  & \multicolumn{1}{c}{\scriptsize (\textbf{4.9}/28)} \\
\bottomrule
\end{tabular}}
\end{table*}

\begin{figure*}[!b] 
\centering
\vspace{-36.5pt}
\begin{minipage}{0.97\textwidth}
\begin{takeawaybox}[Key Takeaways across \survival Survival, \curiosity Curiosity, and \utility Utility]

\begin{itemize}[leftmargin=1.25em,itemsep=0.25em,topsep=0.1em]
  \item \textbf{Prediction-based Methods (WM).}
  Strong at \curiosity, but weaker at \survival\ and \utility. 
  Trained on broad exploratory trajectories, latent rollouts broaden coverage and reveal dynamics, but tend to imitate visually-alike futures rather than reliably pursue goals. So prediction is useful as a look-ahead prior for risk and candidate actions.

    \item \textbf{RL-based Methods (PPO, DQN).}
  Strong at \survival and \utility\ when rewards are well-shaped, but weaker on \curiosity\ and tasks without explicit goals. 
  Reward gradients enable effective credit assignment under the right signal, yet sparsity and partial observability induce instability and interface overfitting, so RL works best as an optimization method.

    \item \textbf{Experience-based Methods (Behavior Cloning).}
  Strong at human-like \survival, but weaker on \curiosity\ and \utility. Deliberately imitate human trajectories and thus excel at low-risk survival, but struggle once tasks require precise control or exploration, and their performance depends strongly on the coverage and quality of the demonstrations.

    \item \textbf{Reasoning-based Pretrained VLMs.}
  Strong at goal-conditioned \survival and \utility; weaker on \curiosity. 
  They excel at instruction-driven reasoning but cannot predict consequences in the visual state space, so they work best as high-level reasoners that need auxiliary prediction modules for outcome-aware decisions.

    \item \textbf{Interactive Physical Reasoner (Ours).}
  Robust across \survival, \curiosity, and \utility. We combine the strengths of all three paradigms: VLMs provide goal-driven causal reasoning, the world model supplies rollout prediction, and RL optimizes decisions using imagined rewards, yielding consistently strong performance across all three levels.

    \item \textbf{Summary.} Prediction-based world models understand dynamics but cannot reliably plan toward long-horizon goals, while reasoning-based VLMs can plan semantically but lack grounded prediction of physical outcomes.
    IPR combines them by using WM rollouts as physical priors and VLM reasoning to select and pursue feasible futures, surpassing GPT-5 with an 8B backbone.

\end{itemize}

\end{takeawaybox}
\end{minipage}
\vspace{-0.4em} 
\end{figure*}

\subsection{Playing in Diverse Physical Worlds}
\label{sec:exp:main}

We evaluate IPR against prevalent baselines on 200 games, chosen to match the full dataset’s distribution of types, action spaces, and physics/causality.
The baselines include:

\begin{itemize}
    \item \textbf{RL.} We utilize Multitask PPO~\cite{yu2020meta} \textit{(policy-based)} and shared-parameter DQN~\cite{osband2016deep} \textit{(value-based)} as standard reinforcement learning approaches.
    \item \textbf{VLM.} We employ a range of vision-language models, including closed-source models such as GPT-4o and GPT-5~\cite{openai2024gpt4ocard}, as well as open-source models like Qwen3-VL-30B-A3B~\cite{yang2025qwen3}.
    \item \textbf{World Model.} We compare three different world models: DreamerV3~\cite{hafner2023mastering} \textit{(latent-based)}, V-JEPA2~\cite{assran2025v} \textit{(pretrained latent-based prediction)}, and Genie~\cite{bruce2024genie} \textit{(pixel-based prediction)} (we follow GenieRedux implementation~\cite{kazemi2024learninggenerativeinteractiveenvironments}).
    \item \textbf{IL.} We apply imitation learning (IL) models, including ACT~\cite{zhao2023learning} \textit{(end-to-end model)} and Qwen3-VL-8B~\cite{yang2025qwen3} \textit{(VLM-based model)}.
\end{itemize}

We assess every model on the three hierarchical objectives, instantiating level-specific training or prompting. Further implementation details are provided in the \textit{supplementary}. The key results are reported in Tab.~\ref{tab:exp:main}.
Takeaways are summarized below the table.

\subsection{Zero-shot Transfer to Unseen Games}
\label{sec:zero_shot_unseen}

To validate our \textit{Games-to-Unseen (G2U)} setting, we construct a held-out target set $\mathcal{T}_{\mathrm{U}}$ of 50 games that are \emph{never} used for training.
From the remaining pool, we form stratified training subsets $\{\mathcal{S}_N\}$ of increasing size $N$, balanced by physics and causal mechanisms 
to control for domain bias.
For each $N$, we train our \textit{IPR} paradigm end-to-end on $\mathcal{S}_N$ and \emph{directly} evaluate zero-shot on $\mathcal{T}_{\mathrm{U}}$ without any adaptation or reward re-scaling.

Across all three objectives, performance increases steadily with $N$, with the steepest early gains on \utility, followed by sustained improvements on \curiosity~and \survival~as more diverse interactions are observed.
This suggests that training in \emph{physically and causally related} environments helps \textit{IPR} move beyond domain-specific quirks (visual style, control interface) and focus on \emph{shared physical and causal patterns} (\eg, gravity, contact, momentum).
In other words, as interactive experience accumulates, \textit{IPR} behaves more \emph{human-like}: it carries over physical priors and causal expectations rather than memorizing domain appearance or controls, demonstrating potential to further scale in richer interactive domains.

\subsection{Ablations and Analysis}
\label{sec:exp:ablation}

\paragraph{Does prediction help VLM reasoning?}
Table~\ref{tab:exp3_ablation} compares variants on the same Qwen3-VL-8B backbone.
Starting from the pretrained VLM, naive BC barely changes survival (0.62$\to$0.63) but \emph{hurts} curiosity and utility, suggesting that low-quality demonstrations can overwrite useful priors instead of improving control.
PPO on top of the VLM achieves the best survival (1.00) and higher utility (1.23), but further suppresses curiosity, and combining PPO with BC degrades all three metrics, indicating that RL alone tends to overfit short-term rewards under biased data.
In contrast, our IPR, which augments the VLM with world-model prediction and GRPO updates, attains the highest curiosity (2.77) while keeping strong survival and utility, showing that prediction-based reinforcement is key to strengthening long-horizon physical reasoning rather than simply pushing for higher immediate scores.

\begin{table}[t]
\centering
\caption{Ablation study results for IPR components of World Model prediction and GRPO.}
\label{tab:exp3_ablation}
\resizebox{0.9\linewidth}{!}{
\begin{tabular}{lccc}
\toprule
\textbf{Method} & \textbf{\survival Survival} & \textbf{\curiosity Curiosity} & \textbf{\utility Utility} \\
\midrule
VLM (pretrained) & 0.62 & 2.14 & 0.89 \\
VLM + BC & 0.63 &1.88  & 0.87 \\
VLM + PPO &\textbf{1.00} &1.79 &1.23 \\
VLM + GRPO &0.95 &1.78 &1.22 \\
VLM + BC + PPO &0.57 &1.86 &0.77 \\
VLM + BC + GRPO &0.55 &1.84 &0.79 \\
\rowcolor{latentblue}
\textbf{IPR} &0.76 &\textbf{2.77} &\textbf{1.34} \\
\bottomrule
\end{tabular}
}
\end{table}

\section{Discussion}

We study an interactive physical reasoner paradigm in which a general-purpose VLM reasons in language, acts through a physics-centric latent interface (PhysCode), and is reinforced by imagined rewards from a world model, asking whether such agents can internalize physical and causal regularities from heterogeneous games and show clear scaling as experience grows.
From this perspective, latent-action world models (\eg Genie, UniVLA~\cite{bruce2024genie,bu2025univlalearningacttaskcentric}) learn discrete action abstractions and latent dynamics for controllable rollouts; imagination-based control methods (\eg Dreamer, V-JEPA2-AC~\cite{hafner2024masteringdiversedomainsworld,assran2025vjepa2selfsupervisedvideo}) optimize policies inside learned world models over device-level actions; and large-scale VLM-based game agents (\eg Game-TARS~\cite{wang2025gametarspretrainedfoundationmodels}) scale vision--language--action models with massive human demonstrations and auxiliary multimodal tasks.
Yet, from a physics-centric perspective, these approaches do not explicitly organize actions by shared physical mechanisms across hundreds of games or align VLM's reasoning ability with prediction competence in a common latent space.
IPR combines their advantages to study how physical knowledge and transfer emerge under the unified Survival-Curiosity-Utility evaluation, though it is still limited to game environments and short-horizon imagination, leaving real-world transfer and longer-horizon reasoning to future work.

\section{Conclusion}
\label{sec:conclusion}
In this work, we introduced \textit{IPR}, a paradigm that \emph{reinforces physical reasoning with prediction} by coupling a physics-centric latent action space (\textit{PhysCode}) with prediction-guided VLM optimization, so that physical and causal regularities are distilled directly from interactive consequences rather than static corpora.
On a curated suite of 1{,}000+ heterogeneous games with \textit{Survival/Curiosity/Utility} evaluation, IPR yields robust gains over VLM-based, prediction-based, and RL-based baselines, and shows strong zero-shot transfer to unseen games (\emph{survive the 1001$^{\text{st}}$ night}).
These results suggest that a general-purpose VLM, when grounded in a physics-organized latent interface and trained with imagined rewards, can indeed \emph{learn} and \emph{scale} its physical reasoning ability purely through interaction, providing a step toward interactive agents that acquire reusable physical and causal knowledge.

{
    \small
    \bibliographystyle{ieeenat_fullname}
    \bibliography{main}
}

\clearpage
\setcounter{page}{1}
\maketitlesupplementary

In this supplementary, we further provide the additional
contents as follows:

Sec. \ref{sec:discussion}: Further Discussion.

Sec. \ref{sec:bench_detail}: Benchmark Details.

Sec. \ref{sec:impl_detail}: Implementation Details.

Sec. \ref{sec:abl}: Additional Ablation Study.

Sec. \ref{sec:case}: Case Study.

\section{Further Discussion}
\label{sec:discussion}

Recent progress in world models and interactive agents has produced systems that can predict future states, learn latent dynamics, and act across large numbers of games.
While we share certain design choices with these systems—such as learning latent dynamics, adopting multimodal interfaces, and scaling across diverse environments—our motivation is fundamentally different. Rather than optimizing for task performance within a single domain, we aim to approximate cross-domain invariants: the physical and causal structures that remain stable across heterogeneous worlds.
Below, we situate our paradigm relative to representative systems, organized by methodological families.

\paragraph{World-model-centric approaches.}

Early world-model work and the Dreamer series~\cite{https://doi.org/10.5281/zenodo.1207631,hafner2020dreamcontrollearningbehaviors,hafner2022masteringataridiscreteworld,hafner2024masteringdiversedomainsworld,hafner2025trainingagentsinsidescalable} show that learning a latent dynamics model and updating a policy from imagined rollouts can master diverse control tasks from pixels. Dreamer progressively strengthens this paradigm: DreamerV1/V2 introduce latent RSSM dynamics with imagination-based actor--critic updates; DreamerV3 demonstrates that a single configuration can reliably solve over 150 tasks across Atari, continuous control, and Minecraft; Dreamer4 further improves robustness and exploration in harder, long-horizon domains. Collectively, these results establish that prediction in latent space is a powerful tool for efficient RL and long-horizon control.

The Genie family~\cite{bruce2024genie,parkerholder2024genie2,genie3} takes a complementary step by treating the world model itself as a \emph{generative environment}. Genie learns a latent action interface from Internet videos and uses it to drive a video world model that converts text or frame prompts into interactive, playable environments. Subsequent versions (Genie-2/3) extend this idea to longer, higher-resolution, and partially 3D worlds with persistent object state and richer user interaction, suggesting that latent world models can serve as general-purpose sandboxes for training and evaluating agents rather than only internal simulators.

V-JEPA and V-JEPA~2~\cite{bardes2024revisitingfeaturepredictionlearning,assran2025v} further push prediction into the feature space: instead of reconstructing pixels, they learn joint-embedding predictive encoders on Internet-scale video. V-JEPA~2-AC augments this with a latent action-conditioned head trained on a small amount of robot interaction data, showing that purely self-supervised video pretraining can be post-hoc adapted into an actionable world model capable of zero-shot manipulation without per-task finetuning. This line of work highlights that high-quality dynamics for physical reasoning do not require pixel-level supervision.

SIMA and SIMA-2~\cite{simateam2024scalinginstructableagentssimulated,sima2} focus on building scalable, instructable multiworld agents in 3D games. SIMA trains a vision–language–action system that follows free-form language instructions across many commercial titles via keyboard-and-mouse control, demonstrating that a single agent can generalise across heterogeneous game interfaces and tasks. SIMA-2 upgrades this framework with a stronger backbone and richer virtual worlds, improving instruction following and in-context learning of new tasks. However, both SIMA variants largely treat the environment as a black box: they rely on language-driven policy learning rather than explicit latent dynamics for imagination or planning.

\paragraph{VLM/VLA-centric agents.}
A parallel line of work builds \emph{vision–language(-action)} agents that treat the game as a black-box interface and learn to map instructions and screen pixels directly to high-level actions. Early VLA-based game agents~\cite{chen2025combatvlaefficientvisionlanguageactionmodel,wang2025uitars2technicalreportadvancing,tan2024cradleempoweringfoundationagents} explore this direction by combining pretrained VLMs with keyboard–mouse or GUI control, often wrapping the environment through OS- or browser-level APIs. These systems demonstrate that a single pretrained backbone can drive diverse games and applications with minimal task-specific finetuning, but typically rely on scripted tools, slow deliberation, or narrow benchmarks.

Game-TARS~\cite{wang2025gametarspretrainedfoundationmodels} pushes this paradigm to scale. It trains a generalist game agent with a unified, human-aligned keyboard–mouse action space, pretraining on hundreds of billions of multimodal tokens collected from OS, web, and simulation games. This large-scale pretraining, together with continual-loss scheduling and sparse-thinking strategies, yields strong performance across open-world Minecraft, web-based 3D games, and FPS benchmarks, often surpassing general-purpose VLMs of comparable size. The key insight is that anchoring the action space to a human-native interface enables broad reuse of trajectories and supports scalable cross-domain training.

Lumine~\cite{tan2025lumineopenrecipebuilding} provides an open recipe for building real-time generalist agents in 3D open worlds. Powered by a VLM backbone, Lumine processes raw pixels at low frequency while emitting precise 30\,Hz keyboard–mouse actions, and adaptively invokes heavier reasoning only when necessary. Trained primarily in a single but rich title (Genshin Impact), it completes hours-long storylines, handles diverse tasks such as exploration, combat, and puzzle solving, and zero-shot transfers to other games with different graphics and interaction dynamics. This line of work underscores that strong semantic reasoning, combined with human-like interaction loops, can already produce impressive in-domain and cross-game competence.

$\pi^\star_{0.6}$~\cite{intelligence2025pi06vlalearnsexperience} is an RL-enhanced large model trained with preference optimization and long-horizon interactive rollouts, producing a general policy that exhibits strong cross-environment competence in web tasks, games, and interactive reasoning. The model benefits heavily from scale, both data scale and model capacity, and demonstrates that sufficiently large policies can generalize to unseen tasks with minimal task-specific engineering. However, $\pi^\star_{0.6}$ does not expose an explicit latent dynamics model, nor does it articulate how prediction or physical regularities structure the policy; its improvements originate primarily from reinforcement tuning on massive interaction data rather than structured cross-domain abstractions.

\paragraph{Our interactive physical reasoner.}

In contrast to the above lines of work, our motivation is explicitly \emph{cross-domain}. Games differ dramatically in appearance, controls, and reward structures, yet we observe that many of them instantiate a small set of shared \emph{physical and causal mechanisms}: gravity, collisions, momentum exchange, and contact-driven state changes. Crucially, these mechanisms tend to be expressed not in pixels but in the \emph{action space}: actions are the agent’s only means to interactively induce physical effects, and different domains often implement similar effects (jump, move, dash, interact) even under mismatched key layouts and visuals.

This suggests that a domain-invariant interface should be built not from raw controls but from a latent action space that captures \emph{what the action does to the world}. Inspired by the intuition-based action extraction in Genie, we learn such a space—\textit{PhysCode}—by encoding visual cues around hand–object–scene interactions and letting a VQ codebook automatically cluster domains whose actions induce similar physical outcomes. PhysCode, therefore, materializes the shared causal structure across heterogeneous games.

With a unified action space in hand, the next question is how to model dynamics. We evaluate both pixel-space and latent-space prediction in the ablation study~\ref{sec:abl_latent_pixel}, and consistent with V-JEPA–style findings, latent dynamics are substantially more efficient and more stable. However, a pure world model—even with accurate latent rollouts—remains confined to intuitive physics and short-horizon prediction; it lacks the high-level reasoning and cross-domain abstraction needed for complex tasks.

To close this gap, we bring in a pretrained VLM that already exhibits some cross-domain generalization in games, as evidenced by works like Game-TARS, but Game-TARS typically relies on costly human prompts and annotations. Instead, we aim to let the agent learn \emph{directly} from interactive environments, using prediction inside the loop of action selection. Our IPR framework couples the two components through PhysCode: the VLM observes the current visual context and task description, then proposes candidate latent actions; the world model performs short-horizon rollouts in PhysCode space to forecast their physical consequences; and a GRPO-style objective reinforces VLM policies whose imagined futures are safe, physically consistent, and task-aligned. In this way, prediction is no longer just an exploration aid—it becomes an in-the-loop imagination mechanism that continuously sharpens the VLM’s physical and causal reasoning across domains.

\section{Benchmark Details}
\label{sec:bench_detail}

\subsection{Game Sources}
\label{sec:bench_source}

\begin{figure*}
    \centering
    \includegraphics[width=0.8\linewidth]{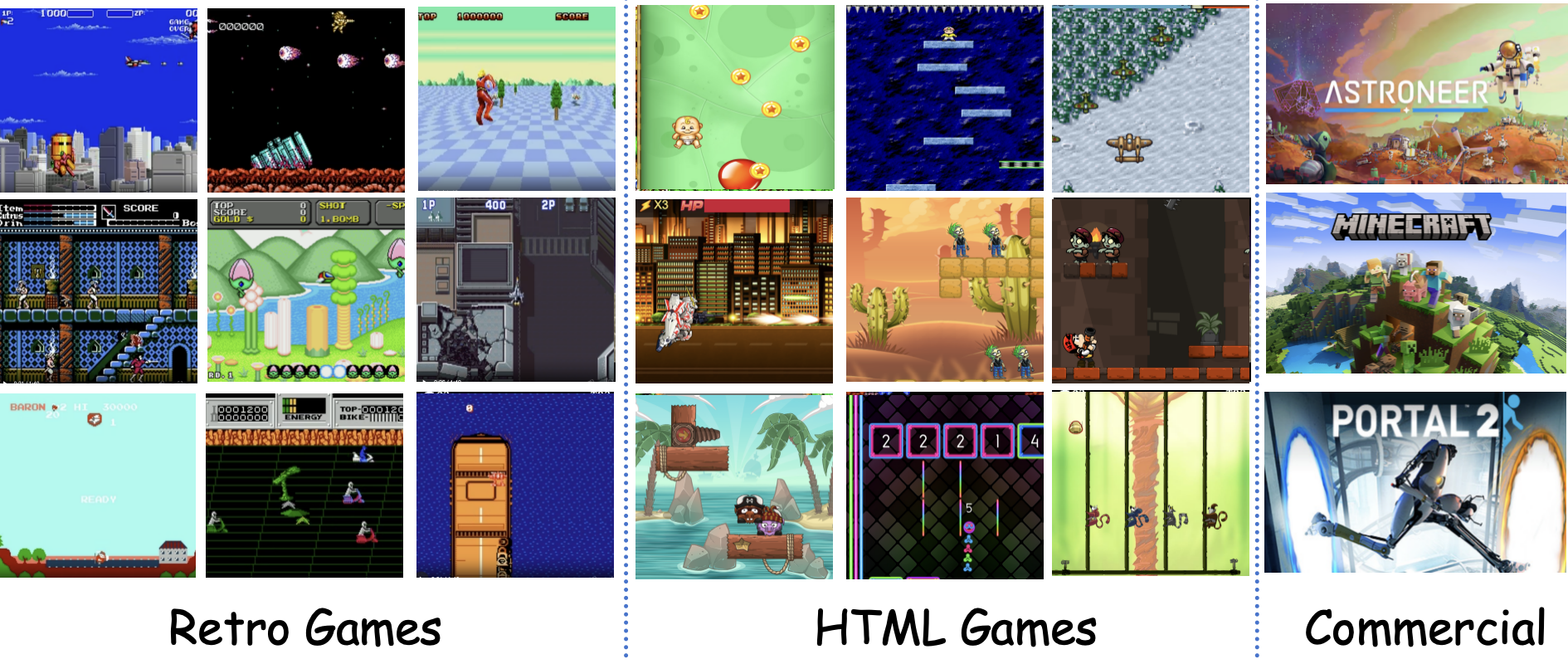}
    \caption{\textbf{Overview of our 1,000 games,} containing old-fashioned retro games, HTML/canvas games, and modern commercial games.}
    \label{fig:overview}
\end{figure*}

\begin{figure*}[t]
    \centering
    \includegraphics[width=\linewidth]{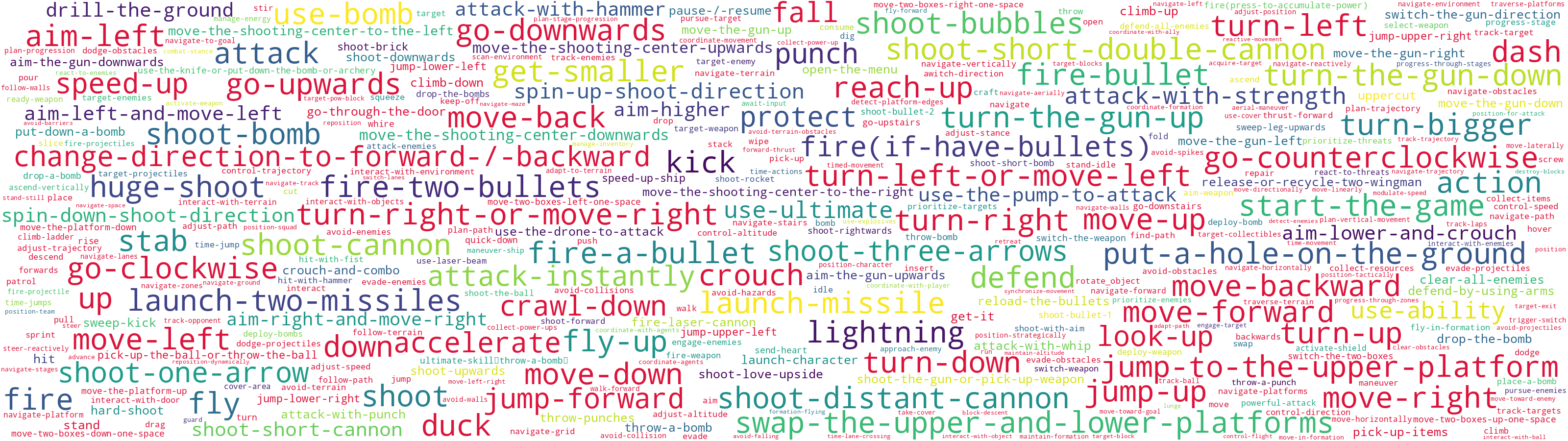}
    \caption{\textbf{Word cloud of action semantics across thousands of game worlds.} These shared semantics provide the structural foundation for cross-domain transfer. Actions highlighted in red represent those shared with general robotic operations, while the size of each word reflects its frequency in our data recipe.}
    \label{fig:wordcloud2}
\end{figure*}

\paragraph{Retro games.}
We curate 863 open-source retro titles via \textsc{stable-retro}~\cite{stable-retro}, covering \textsc{NES}, \textsc{SNES}, \textsc{Genesis}, \textsc{SMS} consoles, \etc.
These environments provide frame-perfect emulation with discrete controller actions (D-pad directions, up to four face buttons, and start/select), and span a wide range of genres including \textit{platformers}, \textit{shooters}, \textit{sports}, \textit{racing}, \etc.
We focus on titles where motion and interaction are governed by clear physical rules (jumping under gravity, rigid-body collisions, projectile–enemy interactions, kinematics, \etc).
For each game, we annotate the dominant \emph{physical} mechanism (\eg, platformer gravity, rigid-body contact, projectile motion, kinematic logic) and \emph{causal} structure (\eg, resource-accumulation objectives, score-based progression, survive-as-long-as-possible tasks, shortest-time-to-goal objectives, or unlocking mechanisms to obtain rewards).
This diversity encourages agents to capture shared physical–causal regularities rather than overfit to title-specific sprites, textures, or control layouts.

\paragraph{HTML games.}
We additionally include 134 HTML/Canvas games collected from public web repositories, comprising both license-free and permissively licensed titles.
Compared to retro consoles, these games rely heavily on mouse and mixed mouse+keyboard interaction (click, drag, hold, scroll), often with modern 2D physics engines (\eg, Box2D-style rigid-body dynamics).
We instrument a Chromium-based browser with a lightweight JavaScript/Playwright wrapper to (i) capture rendered canvas frames at a fixed frame rate and (ii) log low-level input events (mouse position, button state, and keyboard presses) together with timestamps.
When available, we also read a small set of game variables exposed in JavaScript (\eg, score, level, remaining lives) as auxiliary state.
To unify control across heterogeneous HTML titles, we define a hybrid action space consisting of:
(1) a discrete keyboard state vector (one-hot over pressed keys);
(2) a continuous mouse position represented by normalized coordinates $(x, y) \in [0,1]^2$, automatically scaled to each canvas size; and
(3) mouse interaction flags, including left/right click, hold (for continuous dragging), and scroll.
This representation covers the vast majority of HTML/Canvas interaction patterns while remaining compatible with the discrete and low-dimensional interfaces used by other environments in our benchmark.

\paragraph{Commercial games.}
Finally, we include 3 lightweight commercial games using properly licensed PC builds: \textsc{Portal 2}, \textsc{Astroneer}, and \textsc{Minecraft}.
We respect copyright and treat these games as black-box applications, and do not modify their binaries or access internal source code.
Instead, the agent interacts through a virtual desktop: we capture RGB frames from a virtual display, and wrap human-like input via an emulator layer that maps keyboard events to discrete action indices and converts mouse movement/clicks into the same grid-based macro-actions used for HTML games.
For real-time titles, we run the game in a frame-stepped mode: the environment advances one step only after an action is issued, and remains effectively paused while the agent performs reasoning, so decision latency does not affect in-game timing.
To expose the higher-level semantic state of the game, each commercial game is paired with a pre-defined list of target objects and goals.
We record rich-perspective gameplay videos for each title and uniformly sample $\sim$150 frames per game.
On these frames, we manually annotate bounding boxes for all target categories (\eg, cubes, buttons, and portals in \textsc{Portal 2}), and use them to train a lightweight YOLOv11~\cite{khanam2024yolov11overviewkeyarchitectural} detector.
At run time, this detector provides object-centric cues on top of raw RGB, which we use to enrich prompts and evaluation without changing the underlying game binaries.
Across retro, HTML, and commercial games, we use a unified logging interface to record consistent $(x_t, a_t, r_t, x_{t+1})$ trajectories ($r_t$ from the extracted state and rules), enabling joint training and evaluation under a shared interaction format.

\begin{figure*}
    \centering
    \includegraphics[width=0.8\linewidth]{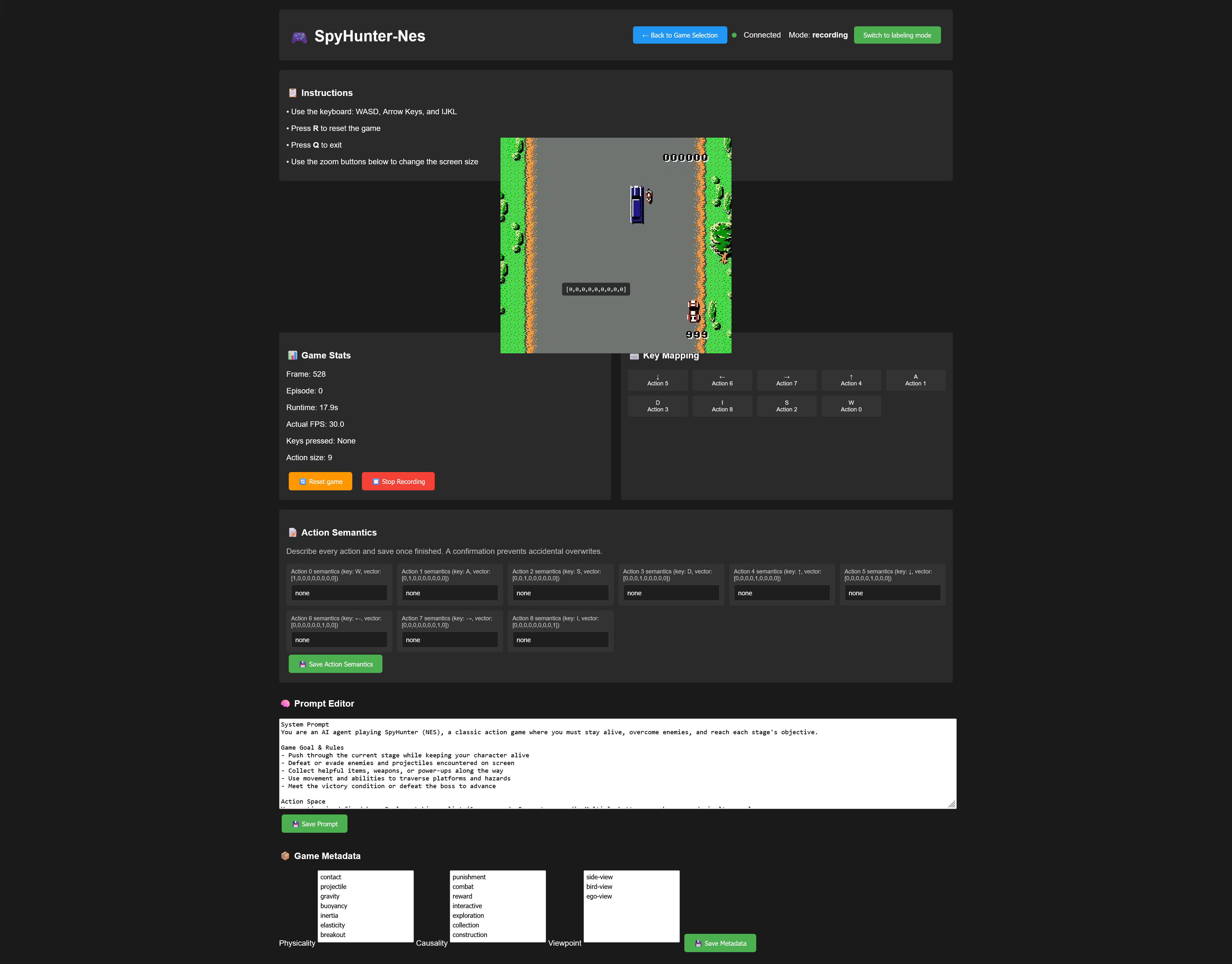}
    \caption{\textbf{Overview of our game-recording website tools.}}
    \label{fig:recording}
\end{figure*}

\subsection{Data Collection and Preprocessing}
\label{sec:data_collect}

\paragraph{Human gameplay recording.}
For each game, we collect \textbf{4 minutes} of human interaction data from \textbf{at least two independent players} to balance individual skill biases.
If the two players' performance exhibits a large score gap (typically $>1.5\times$ difference), we recruit a \textbf{third annotator} to provide additional trajectories, ensuring stable coverage of feasible strategies and reducing overfitting to a single playstyle.
Each trajectory is stored as a sequence of stepwise tuples
\begin{equation}
(x_t, a_t, r_t, x_{t+1}),
\end{equation}
where $x_t$ is the rendered frame, $a_t$ the human action, $r_t$ the instantaneous game reward (if available), and $x_{t+1}$ the next frame.

\paragraph{Frame rates and control logging.}
For \textbf{retro} titles (NES, SNES, Genesis, SMS), we adopt the \textbf{native system frame rate} provided by the emulator and record the console's discrete button events.
For \textbf{HTML/Canvas} and \textbf{commercial games}, we instead capture frames at a unified \textbf{60 FPS}, together with full logs of keyboard events, mouse deltas, and console states, so that all sources can be brought to a common temporal resolution.

\paragraph{Semantic action and physics/causality annotation.}
During recording, annotators additionally provide \textbf{lightweight semantic tags} for each short action segment.
These tags describe both \emph{what} characters are doing: they include action semantics (\eg, \emph{jump}, \emph{dodge}, \emph{charge}, \emph{aim}, \emph{grab}), local physical principles (\eg, \emph{gravity-driven fall}, \emph{sliding under friction}, \emph{momentum carry-over}), and simple causal relations (\eg, \emph{hit switch $\rightarrow$ open door}, \emph{push object $\rightarrow$ block hazard}), as well as short goal/instruction snippets describing the intended skill or sub-task.
These semantics are later used in Sec.~\ref{sec:zero_shot_unseen} in the main paper as grounding signals when inducing semantics-aware actions.

\paragraph{Data preprocessing.}
To ensure uniform sequence quality across heterogeneous sources, we apply a series of preprocessing steps.
First, we \textbf{normalize time intervals}: retro games provide fixed-step transitions through the emulator, while HTML and AAA titles may exhibit variable render intervals.
We resample all trajectories to an aligned 60\,Hz timeline and interpolate missing states when necessary, so that downstream models can assume a fixed time step.
Second, we \textbf{remove non-interactive segments} such as cut-scenes, loading screens, menus, and extended full-idle periods.
These segments are automatically detected using simple motion statistics and input-entropy thresholds over recent frames and key/mouse events.
Third, we \textbf{rebalance idle or no-op periods}: players often hold still or wait for environmental cycles, which would otherwise dominate the dataset.
We therefore downsample long idle windows (\eg, keeping only 1 out of every $k$ idle frames) while explicitly preserving the beginning and end of each idle episode to maintain temporal context.
Finally, we apply \textbf{action smoothing and deduplication}: for mouse movement and other analog-like controls, we smooth out small jitter to avoid spurious micro-movements; for discrete actions, we collapse repeated no-ops or very short flicks that do not meaningfully change the game state.

\subsection{Evaluation Metrics}
\label{sec:eval_metrics}

\paragraph{\survival Survival.}
\survival\ measures the average number of environment steps an agent survives before an irreversible failure (\eg, losing all lives, falling into a death pit, running out of health, or entering a terminal game-over state).
It captures the agent’s ability to avoid risk, prevent collisions, and keep the episode alive.

Game horizons vary widely: in some titles, a random policy dies within tens of steps, while in others it can wander for thousands.
To obtain comparable scores, we derive a per-game \emph{step scale} from the order of magnitude of a random policy’s lifetime.
For each game $m$, we estimate
\(
L^{(m)}_{\text{rand}} = \mathbb{E}[\text{steps}_{\text{rand}}],
\)
and define
\begin{equation}
H^{(m)} = 10^{\lfloor \log_{10} L^{(m)}_{\text{rand}} \rfloor + 1}.
\end{equation}
Given an agent with average lifetime $L^{(m)}_{\text{agent}} = \mathbb{E}[\text{steps}_{\text{agent}}]$, the normalized score is
\begin{equation}
\text{SurvivalScore}^{(m)} = \frac{L^{(m)}_{\text{agent}}}{H^{(m)}}.
\end{equation}
Here $H^{(m)}$ acts as a game-dependent ``step unit'' (\eg, $100$, $1000$, $10000$), keeping survival values in a comparable range without directly normalizing by the exact random baseline.

\paragraph{\curiosity Curiosity.}
\curiosity\ is designed to capture how broadly an agent explores the environment, beyond merely staying alive.
We measure exploration as the \emph{area} of the state space that an agent visits, computed in a learned representation space using \textsc{Magnipy}.
Concretely, we subsample frames from each evaluation episode and embed them with a pretrained vision encoder CLIP to obtain feature vectors $\{f_t\}$.
We then apply \textsc{Magnipy}~\cite{andreeva2023metric,limbeck2024metric} to these features to estimate the volume of the region covered by the agent’s trajectory: \textsc{Magnipy} treats each feature as a point in the embedding space and approximates the union of local neighborhoods around these points, yielding a scalar coverage score that increases when the agent visits new, diverse states and saturates when it revisits already explored regions.
We compute this coverage per episode and average across episodes for each game.

\paragraph{\utility Utility.}
\utility\ measures progress toward explicit task goals, such as maximizing score, winning matches, or completing puzzles.
Because different games expose different reward signals, we unify them into a scalar \emph{game score} before normalization.
In practice, our score types include: (i) raw in-game numerical scores for arcade-style titles; (ii) binary or fractional success indicators for win/loss and puzzle-completion tasks; (iii) time- or step-based objectives where finishing earlier yields a higher score (we invert and rescale time so that ``faster is better''); and (iv) resource-based objectives (\eg, items collected, checkpoints reached) that reflect underlying causal goals such as \emph{collect resources to unlock new areas} or \emph{clear all enemies to progress}.
To compare utility across games, we report a human-normalized score (HNS).
For each game, we measure the average score of a random policy, $\text{score}_\text{rand}$, and the average score of human players, $\text{score}_\text{human}$.
Given an agent with average score $\text{score}_\text{agent}$, we define
\begin{equation}
\text{UtilityScore}
=
\frac{\text{score}_\text{agent} - \text{score}_\text{rand}}
     {\text{score}_\text{human} - \text{score}_\text{rand} + \varepsilon},
\end{equation}
optionally clipped to a reasonable range for robustness.
A value of $0$ indicates random-level performance, $1$ roughly corresponds to human-level performance, and values above $1$ reflect super-human success on the game’s causal objectives.

\section{Implementation Details}
\label{sec:impl_detail}

\subsection{PhysCode}
\label{sec:physcode_supp}

\paragraph{Inputs and temporal windowing.}
Given a gameplay video with per-step controls, we construct short clips of length $T{=}8$.
For each time index $t$, we form a triplet $(x_t, x_{t+\Delta}, y_t)$, where $x_t$ and $x_{t+\Delta}$ are RGB frames and $y_t$ is a lightweight textual description of the executed control (\eg, ``move right and jump'').
We extract three cues:
(i) DINOv3 appearance features $f_t, f_{t+\Delta} \!=\! \phi_{\text{DINO}}(x_t), \phi_{\text{DINO}}(x_{t+\Delta})$ from the final patch tokens (global-pooled to a single 1024-d vector),
(ii) dense optical flow $u_t \!=\! \mathrm{Flow}(x_t, x_{t+\Delta})$ computed by a FlowNet-style network and downsampled to match the DINOv3 patch grid, and
(iii) semantic embeddings $e_t \!=\! \phi_{\text{T5}}(y_t)$ from a frozen T5 encoder (we use the [CLS] token as a 768-d vector).
For efficiency, we precompute $f_t$, $f_{t+\Delta}$ and $u_t$ offline and only store the compact intermediate representations.

\paragraph{Gated modality fusion.}
To form a physics-centric token at time $t$, we first project each modality to a shared $d$-dimensional space ($d{=}512$ by default):
\begin{equation}
\tilde f_t = W_f f_t,\quad
\tilde u_t = W_u\,\mathrm{Pool}(u_t),\quad
\tilde e_t = W_e e_t,
\end{equation}
where $\mathrm{Pool}(\cdot)$ is a spatial average pooling over the flow field.
A small gating MLP $g(\cdot)$ outputs unnormalized gates $(\alpha_f,\alpha_u,\alpha_e)$ conditioned on the concatenation $[\tilde f_t;\tilde u_t;\tilde e_t]$.
We then form normalized gates via a softmax:
\begin{equation}
w_m = \frac{\exp(\alpha_m)}{\sum_{m'\in\{f,u,e\}}\exp(\alpha_{m'})},\quad m\in\{f,u,e\},
\end{equation}
and obtain the fused representation
\begin{equation}
h_t = w_f \tilde f_t + w_u \tilde u_t + w_e \tilde e_t.
\end{equation}
To avoid over-reliance on privileged motion cues, we apply \emph{flow dropout} with probability $p{=}0.5$:
when dropped, the flow feature $\tilde u_t$ is replaced by zero and the gates are renormalized over $\{f,e\}$.
We further add an $\ell_1$ penalty $\lambda_{\text{gate}}\!\sum_t\!\sum_m \lvert w_m - \bar w_m\rvert$ to discourage degenerate single-modality solutions, where $\bar w_m$ is a uniform prior.

\paragraph{Spatio-temporal encoder and codebook.}
A spatio-temporal encoder $E_\psi$ maps the fused sequence $\{h_{t-k}\}_{k=0}^{T-1}$ to a continuous latent $z_t\in\mathbb{R}^d$.
We instantiate $E_\psi$ as a lightweight 6-layer Transformer with hidden size $d{=}512$, 8 attention heads, and a temporal positional embedding; only the last token (corresponding to $t$) is used for code assignment.
We maintain a VQ codebook $\mathcal{C}=\{c_k\}_{k=1}^K$ with $K{=}256$ codes of dimension $d$, updated with EMA.
The continuous latent $z_t$ is quantized to the nearest entry
\begin{equation}
a_t = \arg\min_{k\in\{1,\dots,K\}} \|z_t - c_k\|_2^2,\qquad
\hat z_t \equiv c_{a_t},
\end{equation}
and an \emph{action} is defined as a short sequence $a_t^{\textsc{lat}} = \langle c_{t,1:L}\rangle$ by taking a sliding window of $L$ consecutive indices (we use $L{=}4$ by default).
The sequence representation is obtained by averaging pooling the corresponding embeddings $\{c_{c_{t,\ell}}\}_{\ell=1}^L$.

\paragraph{Training objective and prediction head.}
Given $(f_t, f_{t+\Delta}, h_t)$, the decoder $D_\psi$ predicts the future feature $\hat f_{t+\Delta} = D_\psi(f_t, \hat z_t)$.
We use a standard VQ-VAE loss with a feature-prediction target:
\begin{equation}
\mathcal{L}_{\mathrm{LA}}
= \big\|\hat f_{t+\Delta} - f_{t+\Delta}\big\|_2^2
+ \beta \big\|\mathrm{sg}[z_t] - \hat z_t\big\|_2^2
+ \gamma \big\|z_t - \mathrm{sg}[\hat z_t]\big\|_2^2,
\end{equation}
where $\mathrm{sg}[\cdot]$ denotes stop-gradient and $(\beta,\gamma)$ are codebook and commitment weights (set to $0.25$ and $0.25$ respectively).
We additionally attach a light temporal head that predicts $\Delta f_{t+\Delta} = f_{t+\Delta} - f_t$ from $\hat z_t$, encouraging codes to align with dynamical changes rather than static appearance.

\paragraph{Optimization and data usage.}
PhysCode is pretrained on the entire 1,000+ game corpus.
We randomly sample 4-minute human trajectories per title and extract $(x_t, x_{t+\Delta}, y_t)$ with $\Delta\in\{1,2,4\}$, balancing short- and medium-term dynamics.
We train for 500k steps with AdamW (learning rate $1\!\times\!10^{-4}$, weight decay $0.05$, cosine schedule with 5k warmup), batch size 1024 clips, and gradient-norm clipping at 1.0.
DINOv3, FlowNet, and T5 encoders are frozen; only the fusion module, Transformer, codebook, and decoder are learned.
We found that smaller $K$ (\eg, 128) collapses dynamics from distinct physics into shared codes, while much larger codebooks ($K{\geq}2048$) hurt sample efficiency and lead to under-used codes.

\paragraph{Inference-time usage.}
At test time, optical flow is not available.
We therefore disable the flow gate by fixing $w_u{=}0$ and renormalizing over $\{f,e\}$, and reuse the same $E_\psi$ and codebook to obtain $a_t$ from appearance+semantics only.
The resulting discrete tokens form a temporally predictive vocabulary that (i) clusters trajectories with matched physics (\eg, gravity+contact) and (ii) stays separable under physics shifts, and are used as the shared action interface for both the world model and the VLM in IPR.

\begin{figure*}
    \centering
    \includegraphics[width=0.9\linewidth]{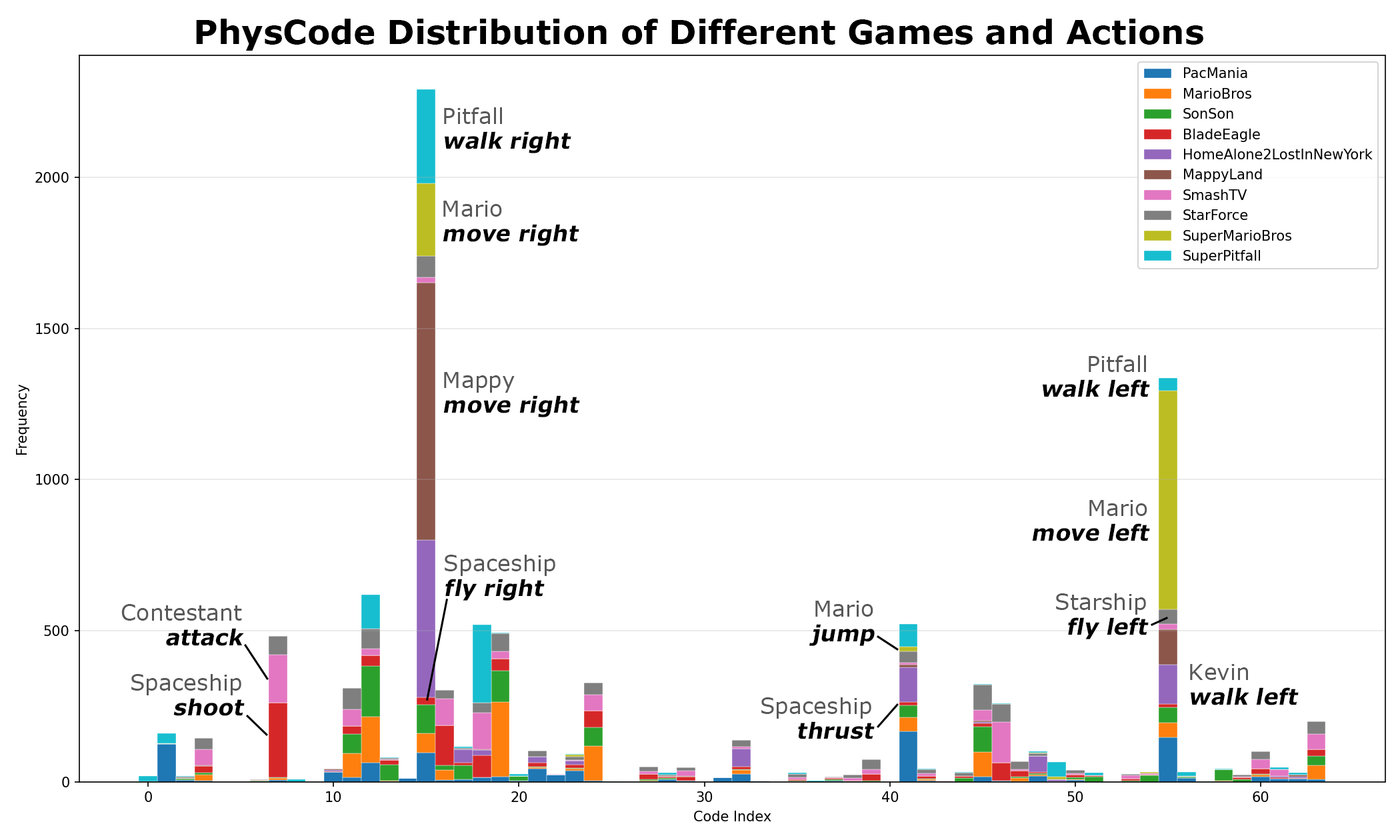}
    \caption{\textbf{Distribution of \textit{PhysCode} in different game domains.} Some action codes share across games, typically \texttt{move right}, \texttt{jump}, while others are separated according to different physical domains.}
    \label{fig:placeholder}
\end{figure*}

\begin{figure*}
    \centering
    \includegraphics[width=0.43\linewidth]{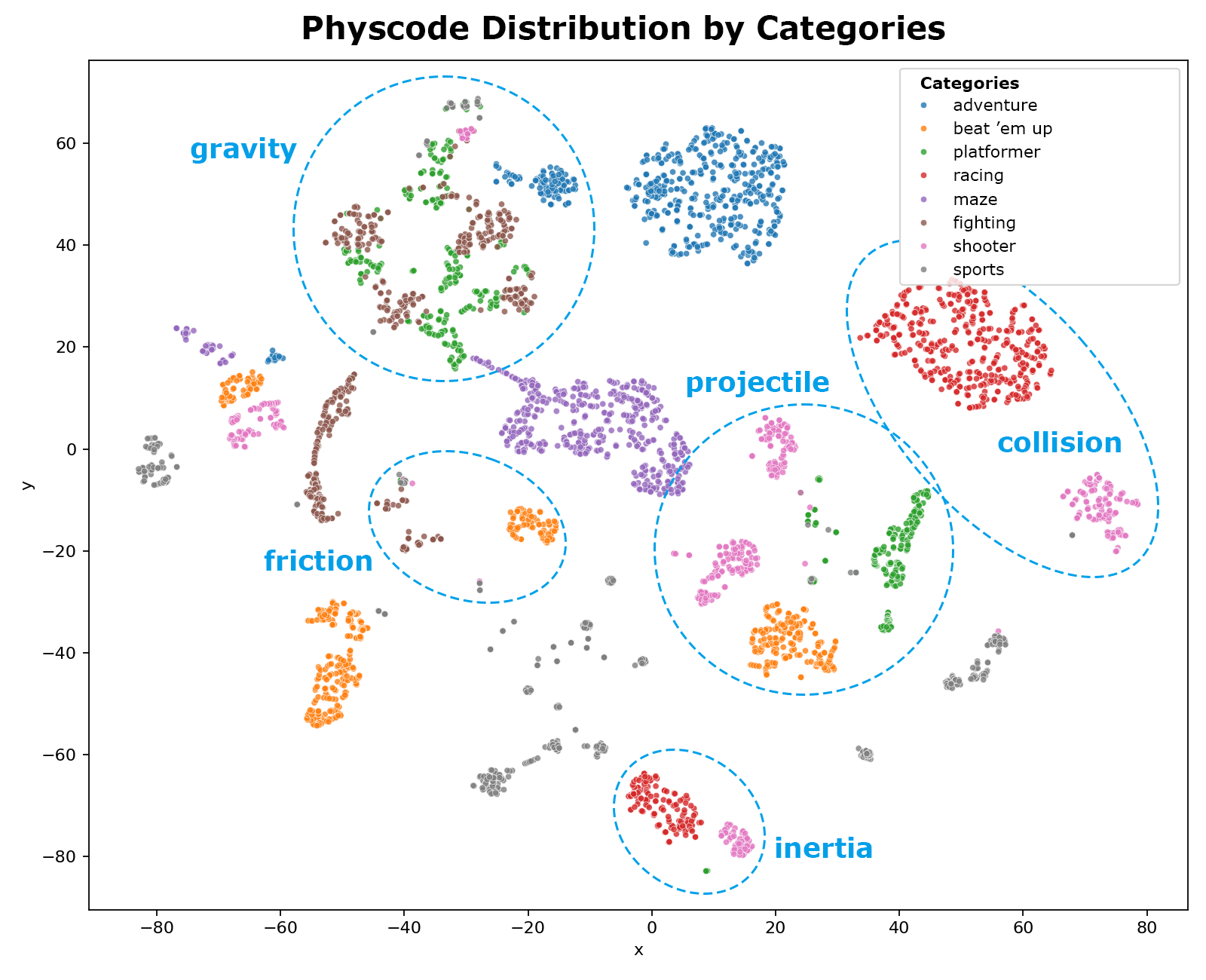}
    \includegraphics[width=0.54\linewidth]{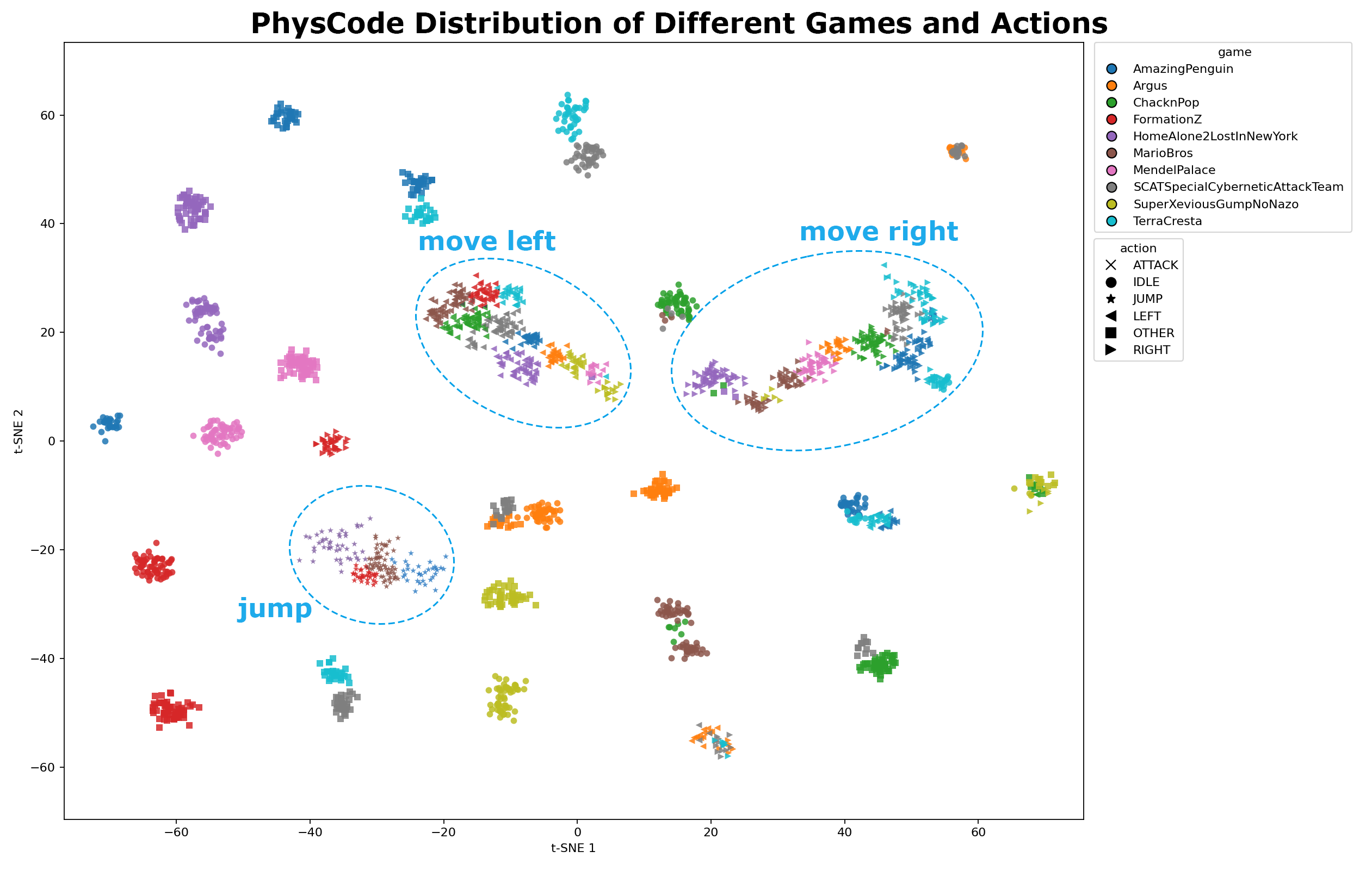}
    \caption{\textbf{Qualitative analysis of PhysCode latent space.} The visualization demonstrates that the learned representations effectively cluster by both high-level physical concepts (left) and fine-grained action-game pairs (right).}
    \label{fig:placeholder2}
\end{figure*}

\subsection{Interactive Physical Reasoner}
\label{sec:ipr_supp}

We detail the three stages of IPR and the training protocol used in our experiments.

\paragraph{Stage 1: PhysCode pretraining.}
IPR builds on the PhysCode vocabulary described in Sec.~\ref{sec:physcode_supp}.
We first pretrain PhysCode on human gameplay across all 1{,}000+ games.
Environments are sampled uniformly over titles and replay segments, and we enforce a balanced mixture over physical mechanisms (gravity, projectile, contact, etc.) to avoid overfitting to a single physics family.
The resulting codebook and encoder are frozen for all subsequent stages.

\paragraph{Stage 2: Latent-conditioned world model with a critic.}
Given fixed PhysCode indices, we replace raw controls by latent action tokens.
For each transition $(f_t, a_t, r_t, f_{t+\Delta})$ we embed $a_t$ into an action embedding $e^a_t \in \mathbb{R}^d$, concatenate with the current feature, and feed into a feature-level predictor $P_\theta$:
\begin{equation}
(\hat f_{t+\Delta}, V_\theta(f_t, a_t)) = P_\theta\big([f_t; e^a_t]\big),
\end{equation}
where $V_\theta$ is a scalar value head sharing all but the last layer with the feature predictor.
We implement $P_\theta$ as an 8-layer Transformer over short latent trajectories (length $H$) to capture multi-step interactions; during training, we unroll on real data segments of length $H{=}5$.

World model training is split into two phases:
(i) a pure prediction phase with
\begin{equation}
\mathcal{L}_{\text{pred}} = \big\|\hat f_{t+\Delta} - f_{t+\Delta}\big\|_1,
\end{equation}
using only pre-collected trajectories and no reward information, followed by
(ii) a value-learning phase, where we freeze the dynamics layers and only train the critic head using a TD-style loss
\begin{equation}
\mathcal{L}_{\text{value}}
= \ell_Q\!\left(
V_\theta(f_t, a_t),\;
r_t + \gamma \max_{a'} V_{\theta^-}(f_{t+\Delta}, a')
\right),
\end{equation}
with a slowly updated target network $\theta^-$.
This separation stabilizes learning: the dynamics focus on physics-consistent feature evolution, while the critic adapts to level-specific reward scales.

\paragraph{Stage 3: VLM alignment to PhysCode.}
We adopt Qwen3-VL-8B as the backbone and extend its tokenizer with $K$ special PhysCode tokens $\{\langle\texttt{PC\_k}\rangle\}_{k=1}^K$.
Each PhysCode index $a_t$ is mapped to its corresponding token, enabling the VLM to produce latent actions as part of its normal autoregressive decoding.

We first perform a perception–action alignment stage on 10k human frame–action pairs.
For each pair $(x_t, a_t^{\textsc{lat}}, g)$, where $g$ is a textual goal or instruction, we format the input as an interleaved image–text prompt and the target as the PhysCode sequence:
\begin{equation*}
\texttt{[IMG($x_t$)] ``Goal: $g$''} \rightarrow \langle\texttt{PC}_{c_{t,1}}\rangle \dots \langle\texttt{PC}_{c_{t,L}}\rangle.
\end{equation*}
We train with a standard teacher-forced cross-entropy loss only on the PhysCode tokens, keeping most of the language parameters close to their initialization via a small learning rate and weight decay.
This stage teaches the VLM to (i) parse visual context, (ii) understand goals, and (iii) output correctly structured PhysCode sequences.

\paragraph{Stage 4: Prediction-reinforced GRPO.}
After alignment, we place the world model in the loop and train the VLM with GRPO using imagined rollouts.
At each real environment step:
\begin{enumerate}
\item Encode the current frame $x_t$ to $f_t$ using the same DINOv3 encoder as in PhysCode.
\item Condition Qwen3-VL on $x_t$ and the current task prompt $g$, and \emph{sample} $B$ candidate latent action sequences $\{a^{(b)}_t\}_{b=1}^B$ (we use $B{=}8$, temperature $0.7$, and top-$p{=}0.9$).
\item For each candidate $a^{(b)}_t$, unroll the world model for $H$ steps in feature space, obtaining a predicted return $\hat R^{(b)}_t$ from the critic head (discount factor $\gamma{=}0.99$).
\item Normalize returns within the candidate set to compute advantages $A^{(b)} = (\hat R^{(b)} - \bar R)/(\sigma_R + \varepsilon)$.
\item Update the VLM with the GRPO objective
\begin{equation}
\begin{aligned}
\mathcal{L}_{\text{GRPO}} =
& - \frac{1}{B} \sum_{b=1}^B A^{(b)} \log \pi_\phi(a^{(b)}_t \mid x_t, g) \\
& + \beta_{\text{KL}}\, \mathrm{KL}\!\big(\pi_\phi(\cdot\mid x_t,g)\,\|\,\pi_{\phi_0}(\cdot\mid x_t,g)\big),
\end{aligned}
\end{equation}
where $\pi_{\phi_0}$ is the initial aligned VLM and $\beta_{\text{KL}}$ controls a conservative trust region.
\end{enumerate}

We interleave real environment interaction and imagination-based updates in a $1{:}k$ ratio (one real step followed by $k{=}4$ imagination-only updates sampled from a replay buffer of recent contexts), which significantly improves data efficiency.

\paragraph{Inference and control routing.}
At test time, the world model remains in the loop but no longer updates.
Given $(x_t, g)$, the VLM proposes $B$ candidate PhysCode sequences as during training; the world model scores them via short-horizon imagination, and we execute the highest-scoring candidate.
A lightweight router $T_{\text{env}}$ maps the selected PhysCode sequence to environment-specific controls (keyboard/mouse macros or gamepad buttons) using a per-game lookup table learned from human trajectories and short calibration episodes.
This keeps the reasoning and prediction in a unified latent space while adapting only a small mapping layer to each new game.

Overall, these stages realize IPR as a \emph{prediction-reinforced reasoning} loop: PhysCode provides a physics-organized latent action interface, the world model supplies imagination and value estimates in this interface, and the VLM is continually refined to prefer actions whose imagined consequences lead to safer survival, broader exploration, and higher utility.

\subsection{Exp 1: PhysCode Validation Setup}
\label{sec:exp1_setup}

\noindent\textbf{Data and model implementation.} 
To investigate how different action spaces influence the learning of shared physical dynamics across heterogeneous environments and their ability to generalize to unseen games, we curate a representative subset of 200 games from our benchmark. For each game, we collect a dataset of 1 million frames paired with ground-truth actions generated by a random policy. We employ \textbf{V-JEPA 2-AC}~\cite{assran2025v} and \textbf{GenieRedux}~\cite{savov2025explorationdrivengenerativeinteractiveenvironments} as the backbone world models. For V-JEPA 2-AC, we first train a ViT-L image encoder from scratch on the combined 200-game dataset (refer to Table~3 in the main paper for encoder ablations). We then post-train the predictive components conditioned on three inputs: (i) the previous frame's image latent $z_{t-1}$, encoded from the raw pixel frame; (ii) the previous action $a_{t-1}$, the representation of which varies by experimental setting (detailed below); and (iii) an auxiliary state vector $s_{t-1}$, which is set to a zero vector for all experiments in this section.

\vspace{1mm}
\noindent\textbf{Action conditioning variants.} 
The core variable in this experiment is the representation of the action input $a_{t-1}$. We compare four distinct configurations:
\begin{enumerate}
    \item \textbf{Keyboard (raw shared).} We train a single model jointly across all games using raw control inputs. We determine the maximum button configuration size within this 200-game domain ($D_{\max}=12$) and pad the multi-hot vectors of simpler controllers with zeros to match this dimension. This represents a naive union of hardware interfaces, where all inputs are normalized to a fixed $1 \times D_{\max}$ vector.
    
    \item \textbf{Language (semantic shared).} We construct a unified semantic action space to resolve the aliasing of raw keys (\eg, key \texttt{A} may trigger \texttt{Jump} in one game but \texttt{Attack} in another). We manually annotate the function of every button in every game using natural language and create a superset of all unique semantics, resulting in a global semantic vector of size $D_{\text{sem}}=173$ (covering actions such as \texttt{move left}, \texttt{jump}, \texttt{shoot}). We generate a static mapping matrix for each game that projects its raw multi-hot vector into this sparse, 173-dimensional global vector.
    
    \item \textbf{PhysCode (Ours).} We use the discretized latent codes derived from our proposed method. As described in Sec.~\ref{sec:physcode} of the main paper, raw actions are replaced by quantized indices $a_t \in \{1, \dots, K\}$ from the learned codebook (we set $K=256$). These indices are projected via a learnable embedding layer before being fed into the world model. Unlike language, this aligns actions based on physical dynamics (\eg, momentum, contact) rather than human-defined semantics.
    
    \item \textbf{Ad-hoc (Single-game expert).} We train a separate world model for each game individually. The input $a_{t-1}$ is the raw game-specific multi-hot vector with dimension $1 \times D_{\text{game}}$. This serves as an oracle upper bound for intra-game prediction quality but lacks any cross-game generalization capabilities.
\end{enumerate}

\vspace{1mm}
\noindent\textbf{Evaluation protocols.} 
We evaluate these representations across three regimes, corresponding to the results reported in Table~1 of the main paper:
\begin{itemize}
    \item \textbf{Confusion test (joint training).} We train a single model on the union of all 200 games and evaluate it on the training set (Table~1a). This measures the model's ability to handle conflicting control schemes (interface aliasing) without performance degradation.
    \item \textbf{Leave-$n$-out transfer.} We then evaluate the same model from the joint training phase on a separate, held-out set of 10 unseen games that were not part of the training data (Table~1b). This protocol tests true zero-shot generalization to entirely new environments using the shared action interface.
    \item \textbf{Physics-conditioned transfer.} To disentangle semantic generalization from physical grounding, we categorize games into four dominant mechanisms (\eg, \textit{Gravity}, \textit{Inertia}). We train specialized models on subsets of 20 games sharing a single mechanism (using the Language-aligned model as a baseline) and evaluate them on held-out games that either match or mismatch the training physics (Tab.~1c in the main paper). This verifies whether the action space captures reusable physical laws or merely memorizes semantic bindings.
\end{itemize}

\subsection{Exp 2: One Model for All Games}
\label{sec:baseline_supp}

\paragraph{RL.}

To instantiate a unified, multi-task model for both \textsc{PPO} and \textsc{DQN} algorithms, we employ a dynamic parameterization scheme. This is achieved by integrating \emph{task embeddings} with a \emph{hypernetwork} architecture. The core idea is to condition the parameters of the policy and value functions directly on the task identity, enabling a single model to specialize its behavior across different tasks.

\noindent The training procedure for a given task is as follows:
\begin{enumerate}
    \item \textbf{Data collection}: Agent interacts with the environment to collect trajectory data $\tau = (s_t, a_t, r_t, s_{t+1})$.
    \item \textbf{Task conditioning}: The current task ID $z$ is mapped to a continuous vector representation $\mathbf{e}_z$ (the task embedding).
    \item \textbf{Parameter generation}: The task embedding $\mathbf{e}_z$ is fed into a hypernetwork $h_{\phi}$, which outputs the parameters $\theta_z$ for the target network:
    \begin{itemize}
        \item For \textsc{DQN}: $\theta_z$ defines the weights of the Q-network.
        \item For \textsc{PPO}: $\theta_z$ defines the weights of the actor $\pi(a|s; \theta_z)$ and critic $V(s; \theta_z)$ heads.
    \end{itemize}
    \item \textbf{Loss computation \& optimization}: The agent's loss (\eg, TD-error for \textsc{DQN}, clipped surrogate objective for \textsc{PPO}) is computed using the generated parameters $\theta_z$. Gradients are backpropagated through both the primary loss and the hypernetwork $h_{\phi}$ to update the shared parameters $\phi$.
\end{enumerate}

\paragraph{VLM.}

We evaluate several strong vision--language policies as prompt-only baselines:
\textbf{GPT-5} and \textbf{GPT-4o}~\cite{openai2024gpt4ocard} (closed-source, accessed through their official APIs),
and two high-capacity open-source models,
\textbf{Qwen3-VL-30B-A3B} and \textbf{Qwen2.5-VL-72B}~\cite{Qwen-VL}.
All models are used in a purely zero-shot manner without any task-specific fine-tuning.

Following the interaction format defined in \textit{videogamebench}~\cite{zhang2025videogamebenchvisionlanguagemodelscomplete},
each query consists of a structured prompt with four components:
(i)~\emph{Game overview} describing the environment type (NES/SNES/Genesis/HTML),
the available control interface, 
and major causal rules (\eg, hazards, damage, reward triggers) to facilitate understanding the target;
(ii)~\emph{Human-annotated action space}, where we provide the discrete actions extracted from human gameplay or emulator documentation, normalized to a canonical textual form;
(iii)~\emph{Task and goals}, summarizing human-labeled objectives
(survival, avoiding collisions, collecting items, defeating enemies, reaching exits);
(iv)~\emph{Step context}, including the current frame, a brief history of recent actions,
and (when available) high-level semantics such as ``the platform collapses after stepping on it'' or
``the projectile follows a parabolic trajectory''.

This format allows each VLM to reason with explicit physics- and causality-related cues
instead of relying solely on one-frame appearance.

\textbf{Required output structure.}
Each model is instructed to always return three fields:
\texttt{THOUGHT} (free-form situational analysis),
\texttt{MEMORY} (persistent long-horizon notes), and
\texttt{ACTION} (the chosen control from the provided action space).
We parse only the \texttt{ACTION} field and execute the corresponding environment action verbatim.
The remaining fields are logged for qualitative analysis and do not affect control.

\textbf{Inference loop.}
At every environment step $t$, the current frame $x_t$,
game description, and the last \(L\) steps of history are inserted into the template.
The model generates autoregressively, and the final \texttt{ACTION: [XXX]} token
is mapped directly to the environment’s action interface.
All baselines use identical prompting templates to ensure fairness across models.

\begin{tcolorbox}[title={Example Prompt for VLM Baselines}]
\small

\textbf{(1) System Prompt.} 
You are an AI agent playing \textit{BillAndTedsExcellentGameBoyAdventure} (Game Boy), 
a classic action game where you must stay alive, overcome enemies, and reach each stage's objective.

\medskip
\textbf{(2) Game Goal \& Rules.}
\begin{itemize}
  \item Push through the current stage while keeping your character alive.
  \item Defeat or evade enemies and projectiles encountered on screen.
  \item Collect helpful items, weapons, or power-ups along the way.
  \item Use movement and abilities to traverse platforms and hazards.
  \item Meet the victory condition or defeat the boss to advance.
\end{itemize}

\medskip
\textbf{(3) Action Space.} \\
Your action is defined by a 9-element binary list (1 = pressed, 0 = not pressed). 
Multiple buttons may be pressed simultaneously.
\begin{center}
\resizebox{\linewidth}{!}{
\begin{tabular}{@{}cll@{}}
\textbf{Index} & \textbf{Button} & \textbf{Meaning} \\
\midrule
0 & B      & Attack with weapons or sprint when held \\
1 & --     & Unused slot -- keep at 0 \\
2 & SELECT & Open sub-menus or cycle through inventory/options \\
3 & START  & Pause the game or open the main menu \\
4 & UP     & Move up, climb, or aim upward \\
5 & DOWN   & Move down, crouch, or drop through platforms \\
6 & LEFT   & Move or face left \\
7 & RIGHT  & Move or face right \\
8 & A      & Jump or confirm actions \\
\end{tabular}
}
\end{center}

\medskip
\textbf{(4) Action Combination Examples.}
\begin{itemize}
  \item Move right: \verb+[0,0,0,0,0,0,0,1,0]+
  \item Move left: \verb+[0,0,0,0,0,0,1,0,0]+
  \item Jump in place: \verb+[1,0,0,0,0,0,0,0,0]+
  \item Jump while moving right: \verb+[1,0,0,0,0,0,0,1,0]+
  \item Trigger a special ability: \verb+[0,0,0,0,0,0,0,0,1]+
  \item Climb or enter upward path: \verb+[0,0,0,0,1,0,0,0,0]+
\end{itemize}

\medskip
\textbf{(5) Output Format.} \\
You \emph{must} respond with \textbf{only} one valid JSON object in the exact format below. Do not include any other text, explanations, or markdown formatting.
{
  ``thought": ``reasoning about the current game state, strategy, and why you choose this action.",
  ``action": ``press\_key",
  ``action\_input": [1,0,0,0,0,0,0,0,0],
  ``memory": ``note about your current status."
}

\medskip
\textbf{(6) Critical Directives.}
\begin{itemize}
  \item \textbf{Fixed Length}: The array length must be exactly 9.
  \item \textbf{Binary Elements}: Elements must be either 0 or 1.
  \item \textbf{Concurrency}: Multiple \texttt{1}s are allowed.
  \item \textbf{Think Then Act}: Analyze internally, and output the JSON format above.
\end{itemize}

\medskip
\textbf{(7) User Prompt.} \\
Analyze the current gameplay frame and output the JSON format above.

\end{tcolorbox}

\paragraph{World model.}
We implement a latent-level video predictor based on V-JEPA2 / Genie-style architectures.
Given $(z_t, a_t)$, the model predicts $(\hat z_{t+1}, \hat r_t)$ using masked temporal transformers.
It performs 5–10 step rollouts for imagined optimization.

\textbf{Genie.}
Our Genie implementation includes the following key enhancements over the baseline \textsc{GenieRedux}:

\begin{itemize}
    \item \textbf{Increased visual fidelity:} The original model operated on low-resolution ($64\times64$) inputs and reconstructions, which we identified as a source of significant information loss due to aggressive downsampling. To mitigate this, we increased the input and output spatial resolution to $224\times224$, thereby preserving finer-grained visual details crucial for complex environments.
    
    \item \textbf{Multi-action embedding:} The baseline \textsc{GenieRedux} was limited to a small, fixed set of five semantically-aligned, one-hot encoded actions. To support a broader and more flexible action space, we designed a novel action processing module. This module takes a multi-discrete action vector (\eg, $[0,1,0,0,0,1,0,1,0]$), identifies the indices of activated actions, performs embedding lookups for each active index, and aggregates the resulting embeddings via mean pooling to produce a unified action representation for the world model.
    
    \item \textbf{Semantic action space alignment:} We extended the action space into a larger, semantically structured space. Furthermore, we performed cross-game semantic alignment on this space, enabling the model to interpret and utilize actions consistently across different tasks and environments.
\end{itemize}

\textbf{DreamerV3.}
We adopt the official DreamerV3 architecture~\cite{hafner2024masteringdiversedomainsworld}, utilizing a shared Recurrent State Space Model (RSSM)~\cite{doerr2018probabilistic} backbone with categorical latent states to capture universal physical dynamics. To handle the distinct objectives and reward scales across 1,000+ games (\eg, sparse survival signals v.s. dense score accumulation), we employ a \textit{multi-head} architecture: while the visual encoder and recurrent dynamics model are shared across all titles, we instantiate separate Actor and Critic heads for each game. We rely on symlog predictions to normalize reward magnitudes and train the shared backbone jointly on all environments.

\textbf{V-JEPA 2.}
Following the V-JEPA 2-AC formulation~\cite{assran2025v}, we employ a non-generative world model that predicts in the representation space rather than reconstructing pixels. The model is trained in two phases: First, a ViT-L video encoder is pre-trained on our dataset of game frames using the self-supervised masked modeling objective, learning to predict latent representations of masked regions. Second, we freeze the encoder and post-train a latent Action-Conditioned (AC) predictor on offline trajectories collected by a random policy. This predictor learns to autoregressively forecast the latent representations of future frames conditioned on the context and action sequence, capturing physical dynamics in the abstract feature space.

\paragraph{IL.}
We include Behavior Cloning (BC) on human frame–action pairs and SFT on VLM, where VLM predicts latent actions from visual tokens.

\textbf{ACT-BC.}
We implement a standard behavior-cloning visuomotor transformer following the ACT paradigm.
Each training sample consists of $(x_{t-H:t}, a_t)$, where $x_{t-H:t}$ are the last $H{=}4$ RGB frames and $a_t$ is either the discrete action vector or PhysCode latent action.
Frames are resized to $128{\times}128$ and encoded by a lightweight 3-layer ConvNet, whose output tokens are fed into a 12-layer transformer.
The model predicts $a_t$ with a cross-entropy loss and is trained jointly on all games without any domain-specific parameters, following the ``single policy for all games'' setting used in ACT.

\textbf{Qwen3-VL-8B-BC.}
We also evaluate a large-model BC baseline using Qwen3-VL-8B.
At each timestep, we construct a simple prompt containing:
(1) the current frame $x_t$ (encoded by the model’s native vision encoder),
and (2) a fixed instruction template, like prompts in the VLM part.
To avoid generating free-form language, we disable chain-of-thought decoding and restrict the output vocabulary to the action only.
BC supervision is applied using next-token prediction: the ground-truth action is appended after a \texttt{<action>} tag, and the model is trained to reproduce it exactly.
We do not use memory tokens, history text, or reasoning steps—Qwen3-VL operates purely as a frame-to-action predictor under teacher forcing.

\textbf{Training.}
Both ACT-BC and Qwen3-VL-8B are trained on the same human trajectories used throughout the paper.
We use AdamW with a learning rate of $2{\times}10^{-4}$ for ACT and $1{\times}10^{-5}$ for Qwen3-VL fine-tuning, batch size 64, and train for 300k steps.
As shown in Sec.~\ref{sec:abl_bc}, low-quality BC supervision may override pretrained priors and degrade long-horizon performance, making these IL baselines strong short-horizon solvers but weak in cross-domain reasoning.

\begin{figure*}[h]
    \centering
    \includegraphics[width=0.9\linewidth]{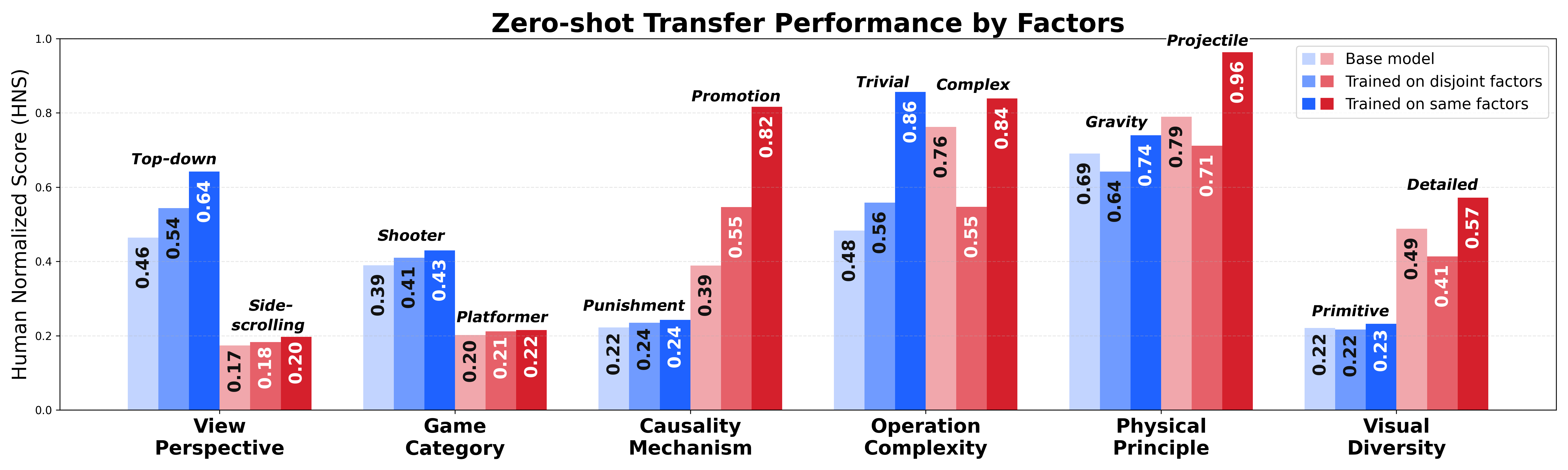}
    \caption{\textbf{Comparative analysis of factor-wise zero-shot transfer performance.} Disjoint training (lighter bars) facilitates generalization on structural factors (e.g., View Perspective) via disentanglement, while it incurs negative transfer on dynamic factors (e.g., Physical Principle) due to cognitive interference.}
    \label{fig:transfer}
\end{figure*}

\subsection{Exp 3: Zero-shot Transfer to Unseen Games}
\label{sec:supp_g2u}

We study Game-to-Unseen (G2U) transfer by splitting the full game pool into a training pool and a held-out target set $\mathcal{T}_U$.
The games in $\mathcal{T}_U$ (20 titles by default) are excluded from \emph{all} training stages, including PhysCode pretraining, world-model learning, and IPR optimization.

From the remaining games, we construct stratified subsets $\{S_N\}$ of increasing size $N$ (from tens to several hundred games).
Each $S_N$ approximately matches the full pool in terms of annotated physical and causal mechanisms (\eg, gravity-driven platformers, projectile shooters, frictional top-down motion, rigid-contact puzzles), control interfaces (NES/SNES/Genesis/SMS/HTML), and visual/genre styles.
This stratification keeps domain bias roughly fixed so that variations in performance mainly reflect the amount and diversity of interactive experience.

For each subset size $N$, we train a full IPR pipeline under a fixed configuration.
PhysCode is pretrained on $S_N$ with the same VQ-VAE setup as in the main paper (codebook size $K{=}512$, identical encoder/decoder, flow dropout $p{=}0.5$) and never sees data from $\mathcal{T}_U$.
The world model is then trained on trajectories from $S_N$ with a fixed architecture and rollout horizon, without any game-specific tuning.
Finally, we attach the learned PhysCode interface and world model to a Qwen3-VL-8B backbone and optimize IPR with GRPO on multi-step latent rollouts.
Learning rate, batch size, rollout length, GRPO sampling temperature, and other optimization hyperparameters are kept identical across all $N$.

Zero-shot evaluation is performed \emph{only} on the unseen titles in $\mathcal{T}_U$.
For a given $N$, we freeze PhysCode, the world model, and the IPR policy, and directly deploy the agent to these games without fine-tuning, reward re-scaling, or game-specific calibration.
We reuse the same PhysCode-to-environment router as in the seen games, and adopt a fixed decoding temperature and sampling scheme at inference.
For each unseen game, we roll out full episodes up to the native time limit or termination, and compute the three metrics defined in the main paper:
\survival\ (normalized survival time), \curiosity\ (normalized exploration coverage), and \utility\ (normalized task reward).
Reported scores are averaged over all episodes and all 50 games in $\mathcal{T}_U$.

Under this protocol, performance on all three objectives improves monotonically with $N$, with larger relative gains at small $N$ and continued, though diminishing, improvements as more diverse games are added.
This supports our claim that G2U behavior is driven by exposure to varied physical and causal environments rather than game-specific tuning.

\subsection{Exp 4: Prediction-Based Interactive Reasoner}
\label{sec:abl_bc}

We ablate four components on a shared Qwen3-VL-8B backbone with the PhysCode interface: world-model prediction, GRPO-based group-wise optimization, PPO-based optimization, and behavior cloning (BC) on 10k human frame–action pairs. We construct a training game set with 200 games, containing all kinds of games. And we then construct one validation game set with 20 never-trained games, balanced by difficulty and novelty. Adding WM prediction and GRPO on top of the pretrained VLM consistently improves survival, curiosity, and utility, indicating that imagination-guided updates strengthen long-horizon reasoning. In contrast, inserting a low-quality BC stage before RL hurts performance: the model overfits to suboptimal demonstrations, its original pretrained reasoning is partially overwritten, and even after GRPO or PPO on optimizations, it underperforms the no-BC variant.

\begin{table}[t!]
\centering
\caption{Latent-action v.s. pixel-based prediction.}
\label{tab:abl_latent_pixel}
\resizebox{0.56\linewidth}{!}{
\begin{tabular}{l c c}
\toprule
\textbf{Agent Type} & \textbf{L1} $\downarrow$ & \textbf{MSE} $\downarrow$ \\
\midrule
Pixel-based & 0.0259 & 0.5622 \\
Latent-based & \textbf{0.0195} & \textbf{0.3821} \\
\bottomrule
\end{tabular}
}
\end{table}

\section{Additional Ablation Study}
\label{sec:abl}

\subsection{Value Prediction in Latent v.s. Pixel Space}
\label{sec:abl_latent_pixel}

To justify our design choice of performing imagination and planning within a compact latent space, we investigate the fidelity of reward prediction when operating on learned representations versus raw sensory inputs. We frame this as a value estimation task using Temporal Difference (TD) learning on offline datasets. Specifically, we curate a dataset of trajectories generated by a random policy and train a value function to predict the expected return (TD target) from a given state-action pair.

We compare two distinct architectures:
(1) Pixel-based Predictor, a convolutional network that takes the raw RGB frame $x_t$ and action $a_t$ as input to directly regress the value; and
(2) Latent-based Predictor, a lightweight MLP that operates on the frozen visual embedding $z_t$ (extracted via the V-JEPA ViT-L encoder used in our main pipeline) concatenated with $a_t$.

We evaluate both models on a held-out test set using Mean Squared Error (MSE) and L1 loss against the computed TD targets. As shown in Table~\ref{tab:abl_latent_pixel}, the latent-based model significantly outperforms the pixel-based baseline.

This substantial performance gap highlights the difficulty of extracting sparse reward signals directly from high-dimensional pixel space, which is often dominated by high-frequency noise, shifting textures, and task-irrelevant background details. The pixel-based model must simultaneously learn to parse visual geometry and estimate value, leading to slower convergence and overfitting to visual nuisance. In contrast, the V-JEPA latent space—pretrained to capture structural and dynamical consistency—effectively filters out these distractions. It provides a compact abstraction of the physical state, allowing the value head to focus entirely on causal associations between states and rewards. This result empirically validates our architecture: performing reasoning and imagination in a semantic latent space is not only computationally efficient but also yields more accurate physical and value predictions than operating in raw pixels.

\subsection{PhysCode Codebook Size}
\label{sec:abl_codebook}

We ablate the PhysCode codebook size with $K\!\in\!\{32,64,128,256,512,1024\}$.
Our goal is twofold: (i) the codebook should be \emph{compact enough} so that codes are effectively used, rather than wasted on rare patterns, and (ii) it should be \emph{expressive enough} to separate distinct control behaviors and their induced physics.

For each $K$, we pretrain PhysCode under the same protocol and evaluate two properties:
(1) \emph{code usage}, computed as $U=\frac{\#\text{main codes}}{K}$ on a held-out split, where \emph{main codes} are those whose empirical usage exceeds $0.1\%$ of all assignments, reflecting how effectively the codebook capacity is utilized;
and (2) \emph{action separation}, measuring the alignment between codes and human action labels (\eg, normalized mutual information between $a_t^{\textsc{lat}}$ and discrete key configurations).

\begin{table}[t]
\centering
\caption{Ablation on PhysCode codebook size $K$.
Code usage is the fraction of codes visited on a held-out split; action separation measures the alignment between codes and human actions (higher is better).}
\label{tab:abl_codebook}
\resizebox{0.8\linewidth}{!}{
\begin{tabular}{c c c}
\toprule
\textbf{$K$} & \textbf{Code usage (\%)} & \textbf{Action separation}  \\
\midrule
32   & 3.1 & 0.006  \\
64   & 3.1 & 0.006  \\
128  & 1.6 & 0.006  \\
\textbf{256}  & \textbf{10.5} & \textbf{0.063}  \\
512  & 0.4 & 0.011  \\
1024 & 0.5 & 0.021  \\
\bottomrule
\end{tabular}
}
\end{table}

We observe a clear trade-off as $K$ varies.
Very small codebooks ($K\!\leq\!128$) over-compress behavior: only a few codes are actually used, and they mix heterogeneous key patterns, leading to poor action separation.
Extremely large codebooks ($K\!\geq\!512$) suffer from the opposite issue: capacity is badly under-utilized (less than $1\%$ of codes are active), and similar behaviors get fragmented across many rarely visited entries.
$K\!=\!256$ strikes a favorable balance, achieving the highest code usage and a substantially stronger alignment with human actions, which in turn yields the best downstream performance.
We therefore adopt $K\!=\!256$ as the default codebook size in all main experiments.

\subsection{Latent Encoder Pretraining}
\label{sec:abl_vjepa}

Our IPR's world model requires an effective visual encoder to ground its predictions in the visual dynamics of the environment. While V-JEPA 2 offers a powerful foundation pretrained on millions of hours of internet videos~\cite{assran2025v}, our work operates in the visually distinct domain of games. This introduces a potential domain gap, where features learned from real-world videos may not be optimal for capturing the specific appearance of game environments.

We compare three ViT-L encoder configurations, all followed by a V-JEPA 2-AC-style predictor trained on our game data: (1) a frozen, off-the-shelf pretrained encoder, (2) the same pretrained encoder fine-tuned on our game trajectories, and (3) a ViT-L encoder trained from scratch using only our game dataset.

The results in Table~\ref{tab:vjepa_encoder_ablation} demonstrate a clear performance hierarchy. The frozen pretrained encoder performed the worst, confirming a significant domain gap. Fine-tuning offered a moderate improvement, but the best predictive accuracy was unequivocally achieved by the encoder trained from scratch on in-domain data. This indicates that for specialized visual domains such as retro games, a tailored feature extractor is more effective than adapting a general-purpose one, as it avoids the potentially confounding inductive biases from out-of-domain pretraining.

\subsection{Transfer Factors across Categories}
To further validate the robustness of our representations across diverse environments, we conduct a fine-grained factor-wise transfer analysis as illustrated in Fig.~\ref{fig:transfer}, which reveals two distinct generalization regimes contingent on the nature of the underlying factors. For structural attributes such as \textit{view perspective}, \textit{causality mechanism}, and \textit{game category}, our results show that training on disjoint factor sets facilitates positive transfer, as varying these surface-level configurations forces the model to disentangle intrinsic task dynamics from visual layouts. Conversely, factors tied to latent mechanics, specifically \textit{physical principles} and \textit{operation complexity}, often exhibit negative transfer where strong priors learned from source environments conflict with target mechanics. This phenomenon mirrors human \textbf{cognitive interference}, suggesting that while structural diversity promotes invariant representations, mechanical discrepancies remain a primary bottleneck for zero-shot adaptation.

\begin{table}[t!]
\centering
\caption{Ablation on ViT-L encoder configurations for the world model.}
\label{tab:vjepa_encoder_ablation}
\resizebox{0.4\textwidth}{!}{
\begin{tabular}{@{}lccc@{}}
\toprule
\textbf{Encoder Setup} & \textbf{Cosine} $\uparrow$ & \textbf{MSE} $\downarrow$ & \textbf{L1} $\downarrow$ \\
\midrule
Pretrained & 0.8888 & 0.2223 & 0.3262 \\
Fine-tuned & 0.8924 & 0.2150 & 0.2705 \\
Trained from Scratch & \textbf{0.9891} & \textbf{0.0216} & \textbf{0.0758} \\
\bottomrule
\end{tabular}
}
\end{table}

\section{Case Study}
\label{sec:case}

\subsection{Experience-based Methods}

\begin{figure*}
    \centering
    \includegraphics[width=0.9\linewidth]{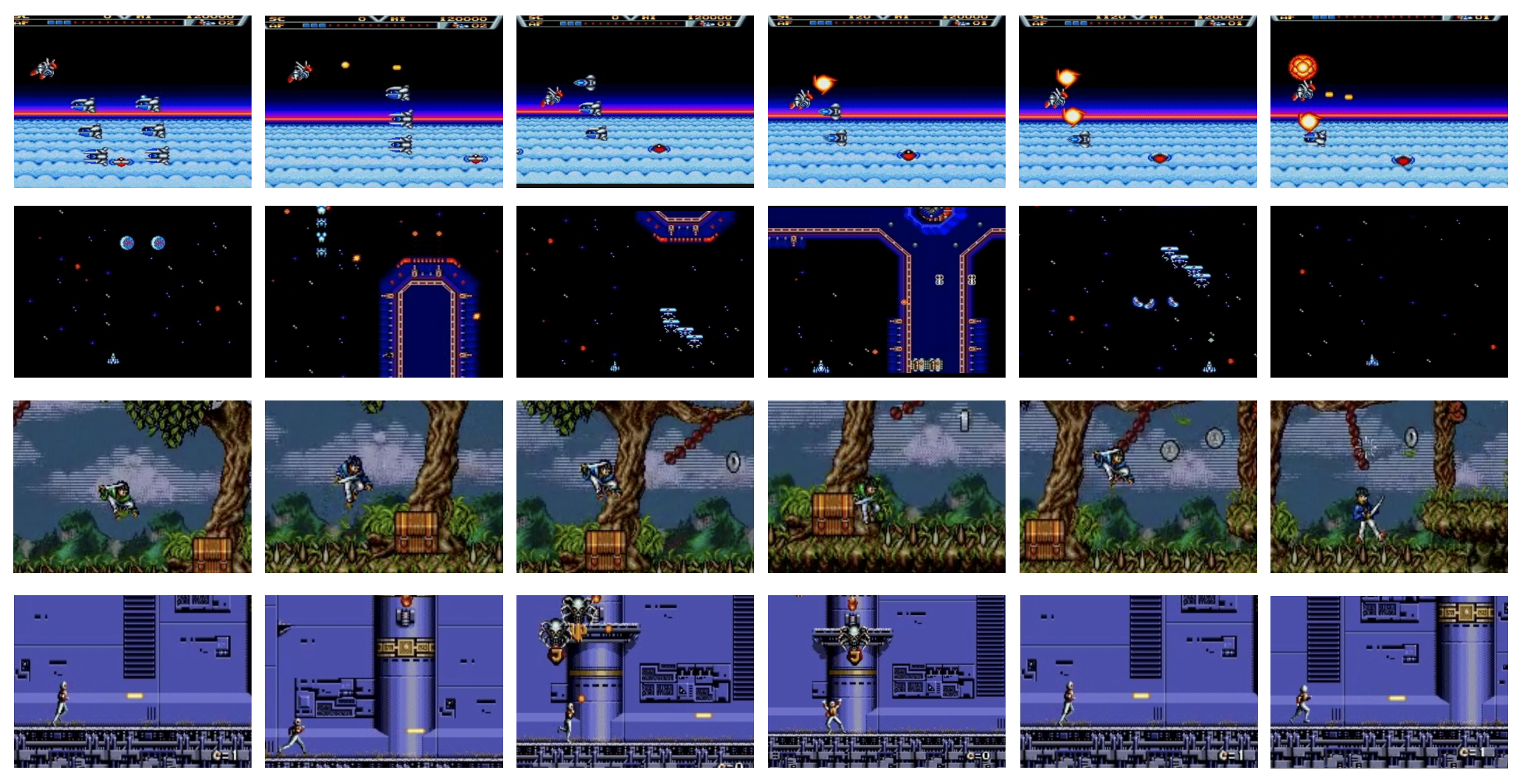}
\caption{
\textbf{ACT Case Study.}
The figure highlights four representative behaviors of ACT: 
(1) Line~1 shows that ACT can solve difficult segments by leveraging human demonstrations and extracting effective strategies; 
(2) Line~2 illustrates that imitation enables high scores on tasks with stable, low-variance dynamics; 
(3) Line~3 reveals that ACT also absorbs human failure patterns, reproducing suboptimal attempted actions; and 
(4) Line~4 shows that its behavior is highly sensitive to environmental noise, often leading to unstable or inconsistent actions.
}
    \label{fig:act}
\end{figure*}

\begin{figure*}
    \centering
    \includegraphics[width=0.9\linewidth]{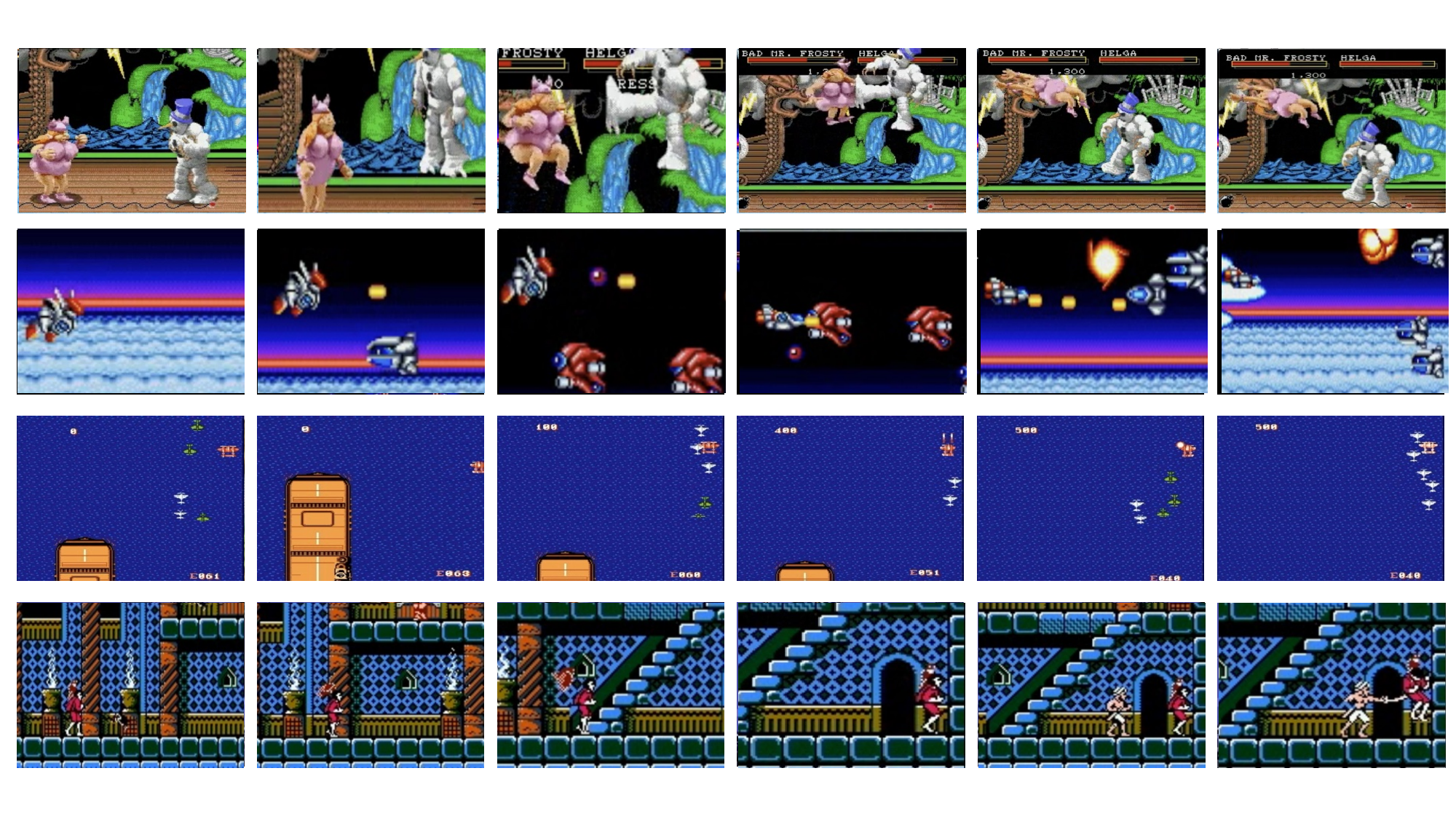}
    \caption{
\textbf{Qwen-BC Case Study.}
The figure illustrates four characteristic behaviors of the BC-trained Qwen agent: 
(1) Line~1 shows that the agent can faithfully reproduce high-difficulty actions; 
(2) Line~2 demonstrates its strong temporal stability and highly consistent action repetition; 
(3) Line~3 reveals its poor generalization to novel or perturbed situations; and 
(4) Line~4 shows its tendency to collapse into passive idling or a single repeated action when the action distribution is imbalanced.
}
    \label{fig:qwen-bc}
\end{figure*}

Experience-based agents such as ACT and Qwen-BC show clear strengths in learning from human demonstrations, but also exhibit systematic limitations.
ACT (Figure~\ref{fig:act}) benefits strongly from expert trajectories: it can extract effective strategies for difficult segments and achieves high scores on tasks with stable, low-variance dynamics. However, its imitation-heavy nature makes it prone to inheriting suboptimal human behavior and to degrading under environmental noise.
Similarly, Qwen-BC (Figure~\ref{fig:qwen-bc}) excels at reproducing high-difficulty actions with high fidelity and maintains very stable action sequences, yet its generalization is weak. When facing novel situations or imbalanced action distributions, the policy often collapses into passive idling or repetitive single-action loops.
These behaviors collectively show that experience-based policies are powerful in familiar, low-variance regimes but struggle to extrapolate, exposing the limits of purely demonstration-driven learning in interactive environments.

\subsection{RL-based Methods}
\begin{figure*}
    \centering
    \includegraphics[width=0.85\linewidth]{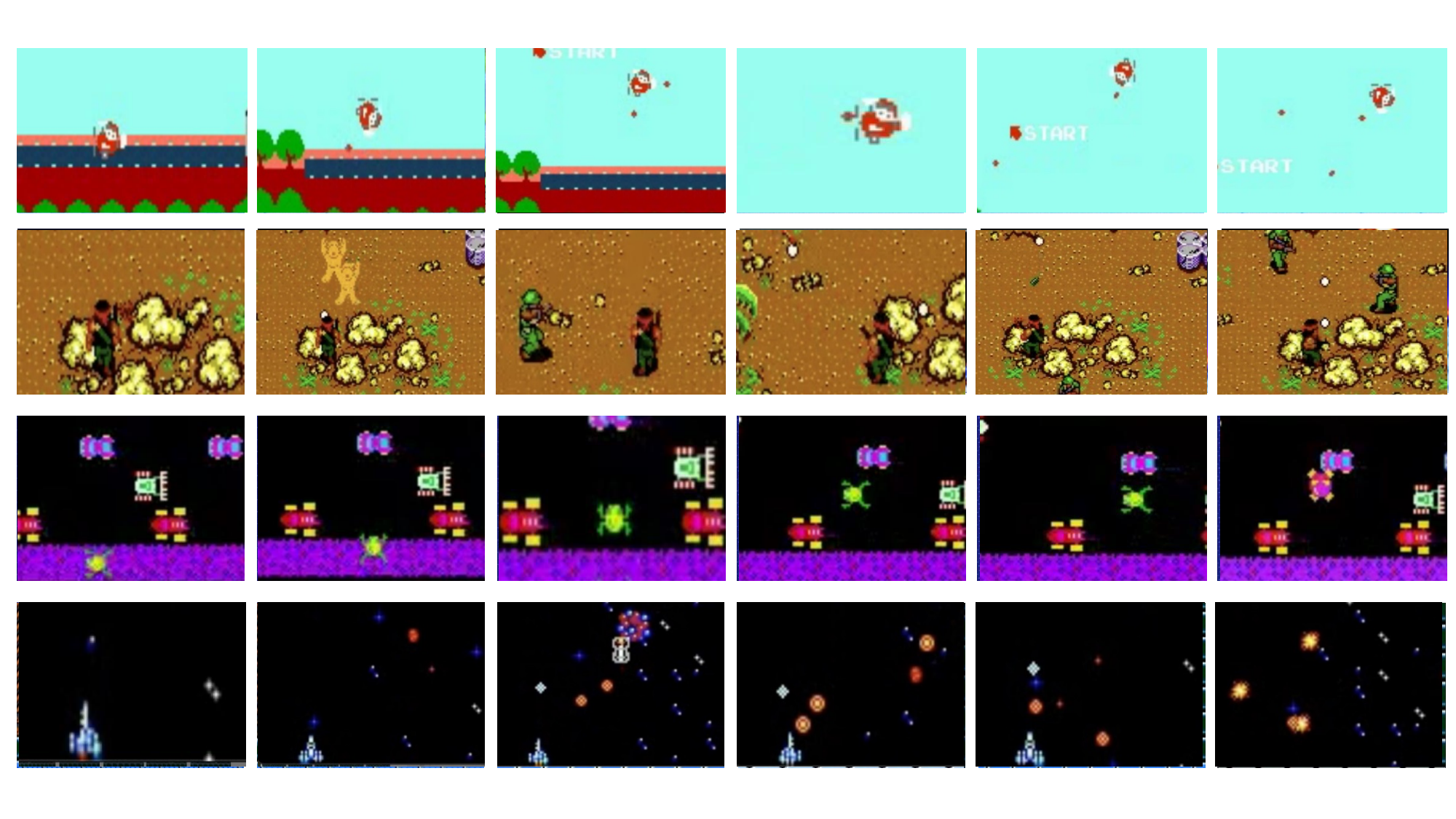}
\caption{
\textbf{PPO Case Study.}
The figure presents four typical behaviors of the PPO agent:
(1) Line~1 demonstrates that PPO can learn effective action sequences, enabling the agent to simultaneously shoot while dodging bullets through rolling maneuvers;
(2) Line~2 illustrates its capacity to not only acquire efficient key-press strategies but also identify primary movement directions that drive game progression;
(3) Line~3 reveals that due to poorly shaped rewards, PPO may exploit design flaws by repeatedly moving forward and backward to farm progression rewards in a degenerate manner; and
(4) Line~4 shows that as the exploration rate decays in later training stages, the agent nearly halts exploration, leading to a performance plateau and frequent failures at identical game states.
}
    \label{fig:ppo}
\end{figure*}

\begin{figure*}
    \centering
    \includegraphics[width=0.9\linewidth]{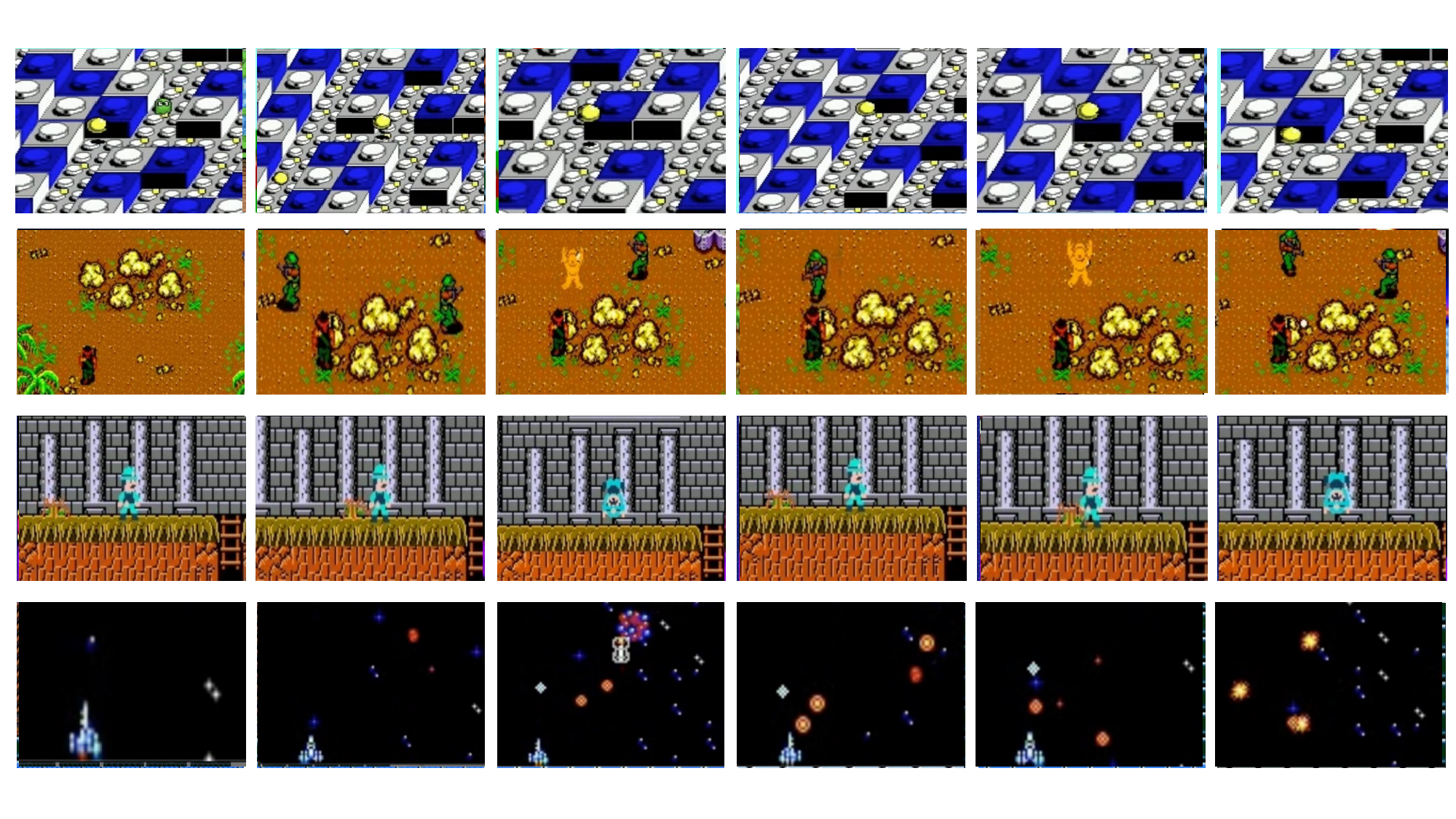}
\caption{
\textbf{DQN Case Study.}
The figure presents four typical behaviors of the DQN agent: 
(1) Line~1 shows that it can correctly identify when specific actions should be executed; 
(2) Line~2 illustrates its ability to detect and exploit advantageous environmental features (\eg, using rocks as cover); 
(3) Line~3 reveals that poorly shaped rewards can lead the agent to adopt degenerate strategies, such as repeatedly ``dying on purpose'' when death yields positive reward; and 
(4) Line~4 shows that the agent may fall into meaningless repetitive actions, such as continuous jumping without purpose.
}
    \label{fig:dqn}
\end{figure*}

Reinforcement learning agents, such as PPO (Figure~\ref{fig:ppo}) and DQN (Figure~\ref{fig:dqn}), demonstrate distinct ability to master complex motor control and identify key environmental interactions. These models excel at discovering effective key combinations for simultaneous maneuvering and attacking, as well as exploiting environmental features like cover to advance game progression. However, their reliance on scalar reward signals leads to significant brittleness. As observed in the case studies, both PPO and DQN are prone to exploiting poorly shaped rewards, leading to degenerate strategies like ``dying on purpose" or repetitive movement loops to farm points. Furthermore, as exploration rates decay, these agents frequently suffer from policy collapse, halting necessary exploration and failing repeatedly at identical game states due to a lack of semantic understanding.

\subsection{Prediction-based Methods}
\begin{figure*}
    \centering
    \includegraphics[width=0.9\linewidth]{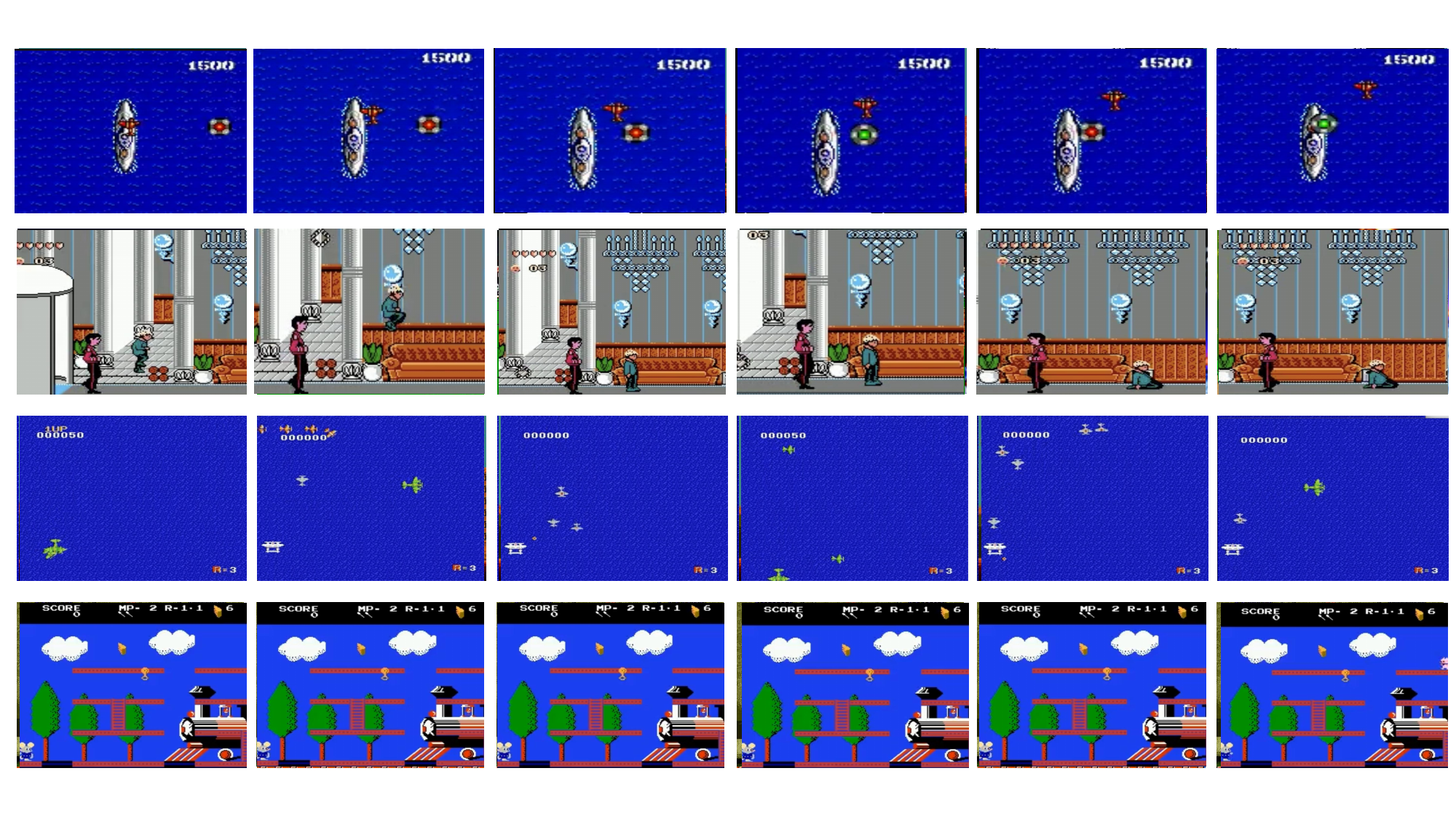}
\caption{
\textbf{DreamerV3 Case Study.}
The figure illustrates four characteristic behaviors of the Dreamer agent:
(1) Line~1 shows that Dreamer reliably exhibits risk-avoiding behavior and tends to choose actions that maximize short-term safety; 
(2) Line~2 demonstrates its strong temporal stability, often producing highly repetitive and consistent action sequences; 
(3) Line~3 reveals a biased policy that over-relies on a single heuristic—such as persistently moving left to evade enemies, and 
(4) Line~4 shows that the agent can easily collapse into inaction, remaining stationary when its value estimates become uncertain.
}
    \label{fig:dreamer}
\end{figure*}

\begin{figure*}
    \centering
    \includegraphics[width=0.9\linewidth]{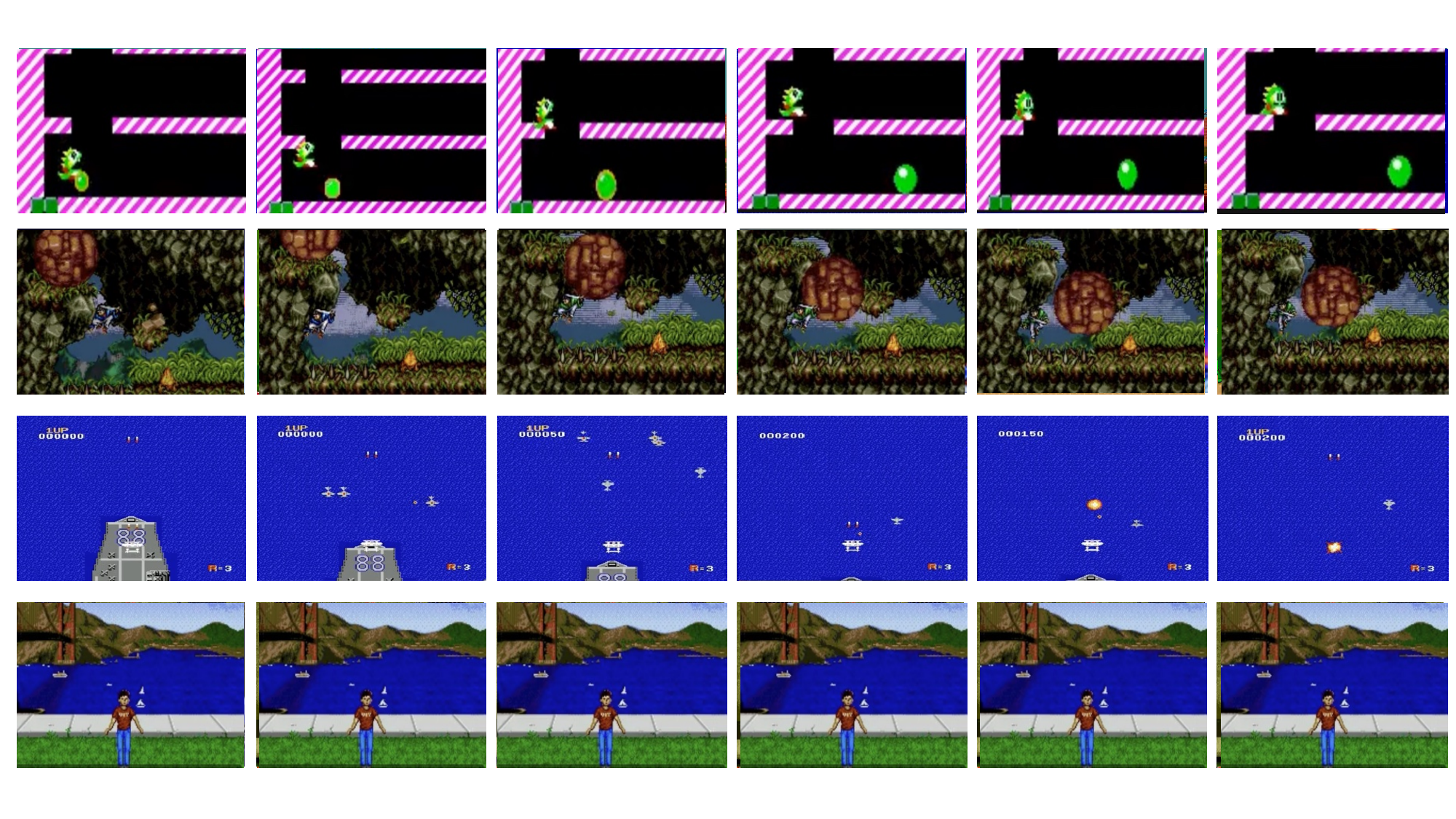}
\caption{
\textbf{V-JEPA2 Case Study.}
The figure illustrates four representative behaviors of the V-JEPA agent: 
(1) Line~1 shows that the agent maintains high action efficiency with minimal redundancy, avoiding the ineffective key combinations often observed in other models; 
(2) Line~2 demonstrates its capacity for strategic environmental exploitation, such as utilizing terrain features (\eg, rocks) to evade hazards; 
(3) Line~3 reveals its vulnerability to collapse into passive idling by outputting zero vectors; and 
(4) Line~4 indicates that this susceptibility to inaction persists when immediate feedback is lacking, resulting in a failure to initiate necessary exploration.
}
    \label{fig:vjepa}
\end{figure*}

\begin{figure*}
    \centering
    \includegraphics[width=0.85\linewidth]{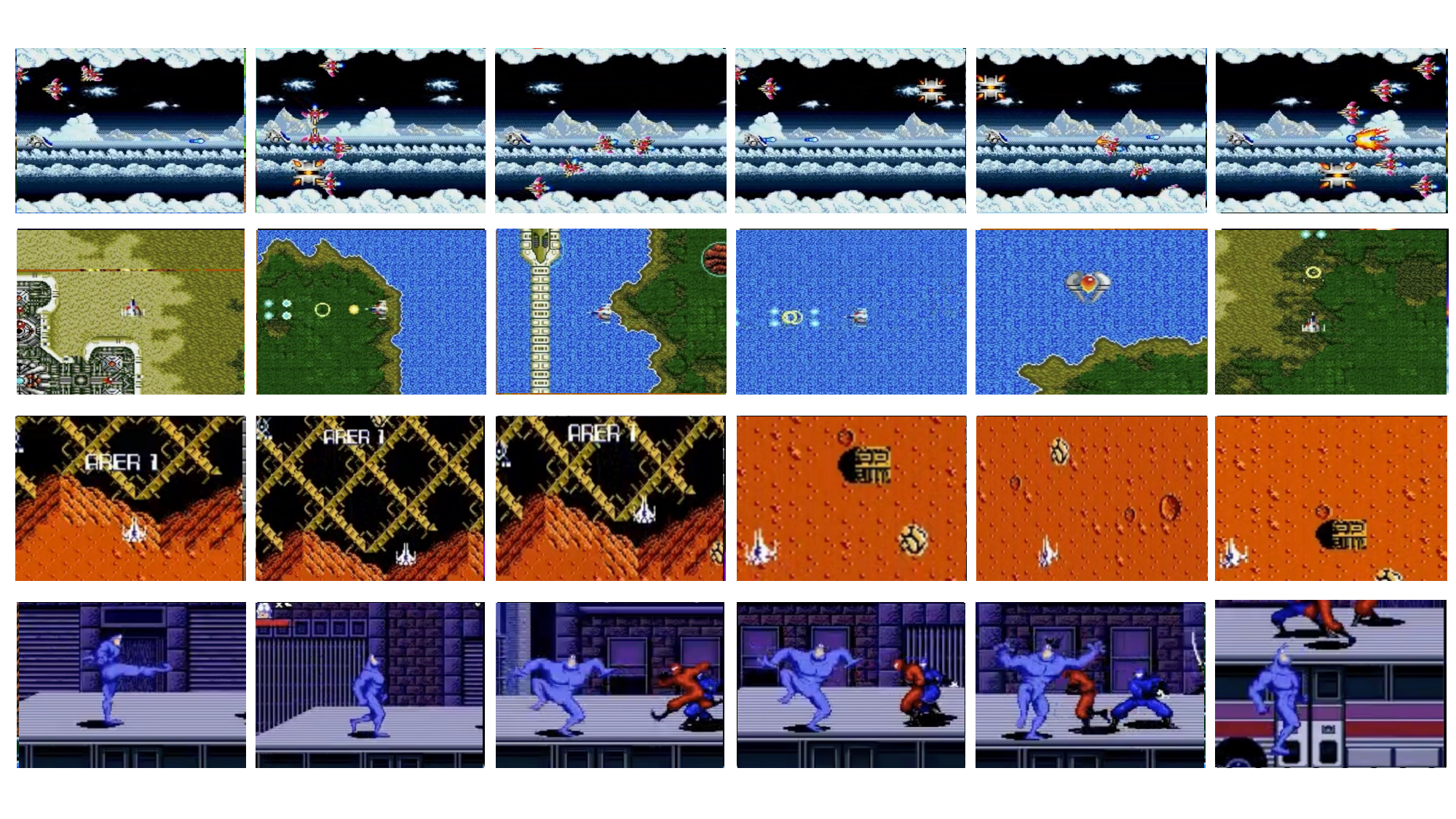}
\caption{
\textbf{Genie Case Study.}
The figure presents four key capabilities and limitations of our Genie-based world model:
(1) Line~1 demonstrates enhanced motion trajectory prediction, enabling the agent to execute preemptive evasion maneuvers;
(2) Line~2 reveals the emergence of strategic path planning, where the agent learns systematic navigation paths beyond reactive bullet avoidance;
(3) Line~3 illustrates a critical model limitation: in environments with poor bullet reconstruction fidelity, the agent fails to develop shooting behaviors and defaults to purely defensive evasion strategies;
(4) Line~4 highlights spatial reasoning deficiencies, where prediction inaccuracies in distance estimation lead to occasional falls from platform edges.
}
    \label{fig:genie}
\end{figure*}

World model approaches, represented by Dreamer (Figure~\ref{fig:dreamer}), Genie (Figure~\ref{fig:genie}), and V-JEPA (Figure~\ref{fig:vjepa}), exhibit strong temporal stability and high action efficiency. These agents are characterized by risk-averse behaviors, prioritizing short-term safety and minimizing redundant inputs. V-JEPA, for instance, shows strategic capacity in utilizing terrain features for evasion. However, a critical limitation shared across these prediction-based methods is susceptibility to passivity. When value estimates become uncertain or immediate feedback is lacking, these models often collapse into inaction—idling or outputting zero vectors rather than initiating exploration. Additionally, they can develop biased policies that over-rely on simple heuristics, such as persistently moving in a single direction, failing to adapt when dynamic hazards require complex, non-linear responses.

\subsection{Reasoning-based Methods}
\begin{figure*}
    \centering
    \includegraphics[width=0.85\linewidth]{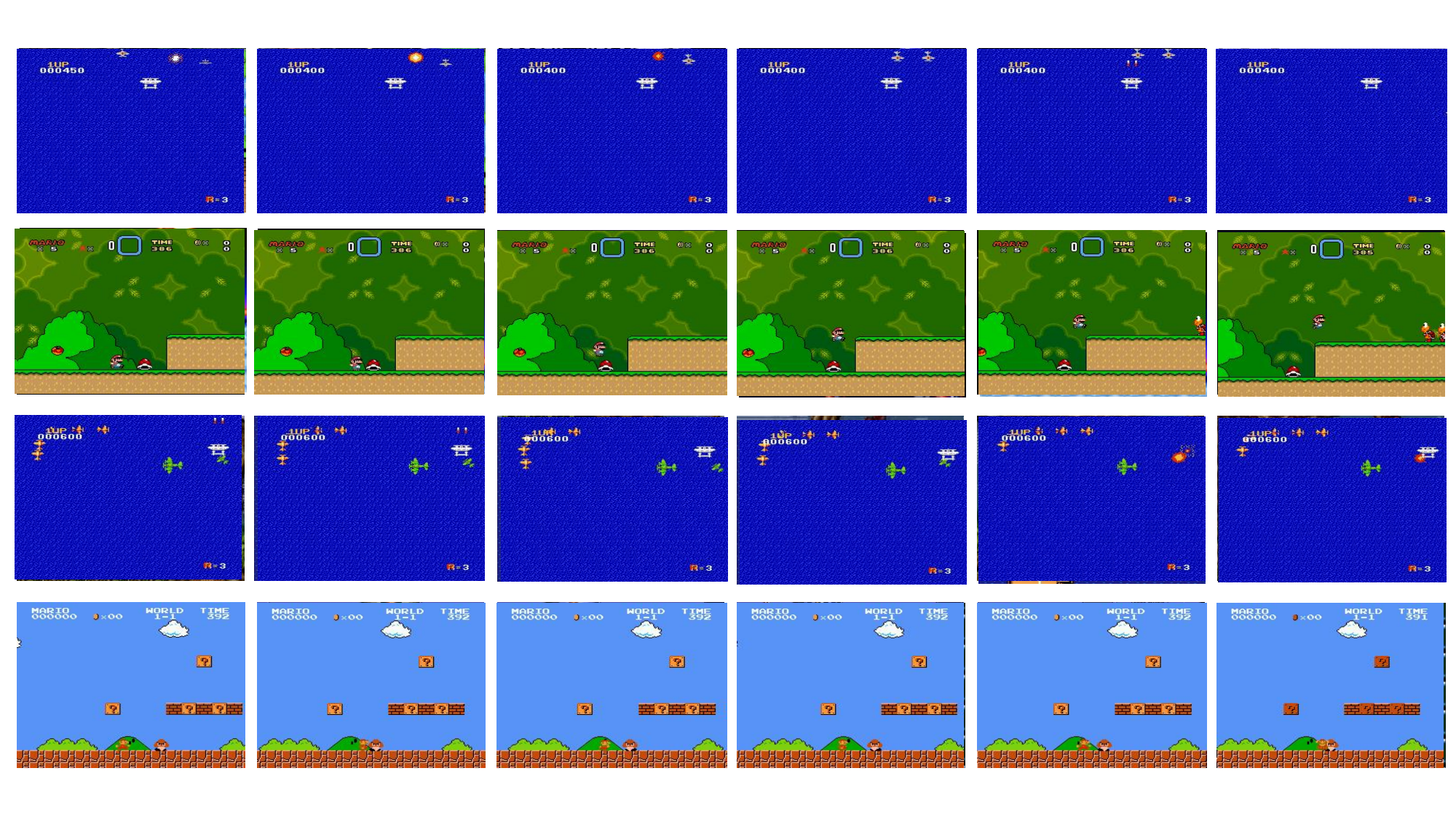}
\caption{
\textbf{GPT-4o Case Study.}
The figure illustrates four char-
acteristic behaviors of the GPT-4o agent. (1) Line 1 shows that the agent demonstrates effective target engagement and reaction speed, discharging projectiles to neutralize an aerial threat; (2) Line 2 highlights its proficiency in precise spatial navigation, executing a controlled jump to successfully land on the target platform; (3) Line 3 reveals a blind spot in situational awareness regarding rear-approaching entities, where the agent fails to evade the trailing aircraft, resulting in a fatal collision; and (4) Line 4 indicates a fundamental failure in basic obstacle avoidance, where the agent initiates a direct collision with a visible ground enemy rather than executing an evasive maneuver.
}
    \label{fig:gpt4o}
\end{figure*}

\begin{figure*}
    \centering
    \includegraphics[width=0.85\linewidth]{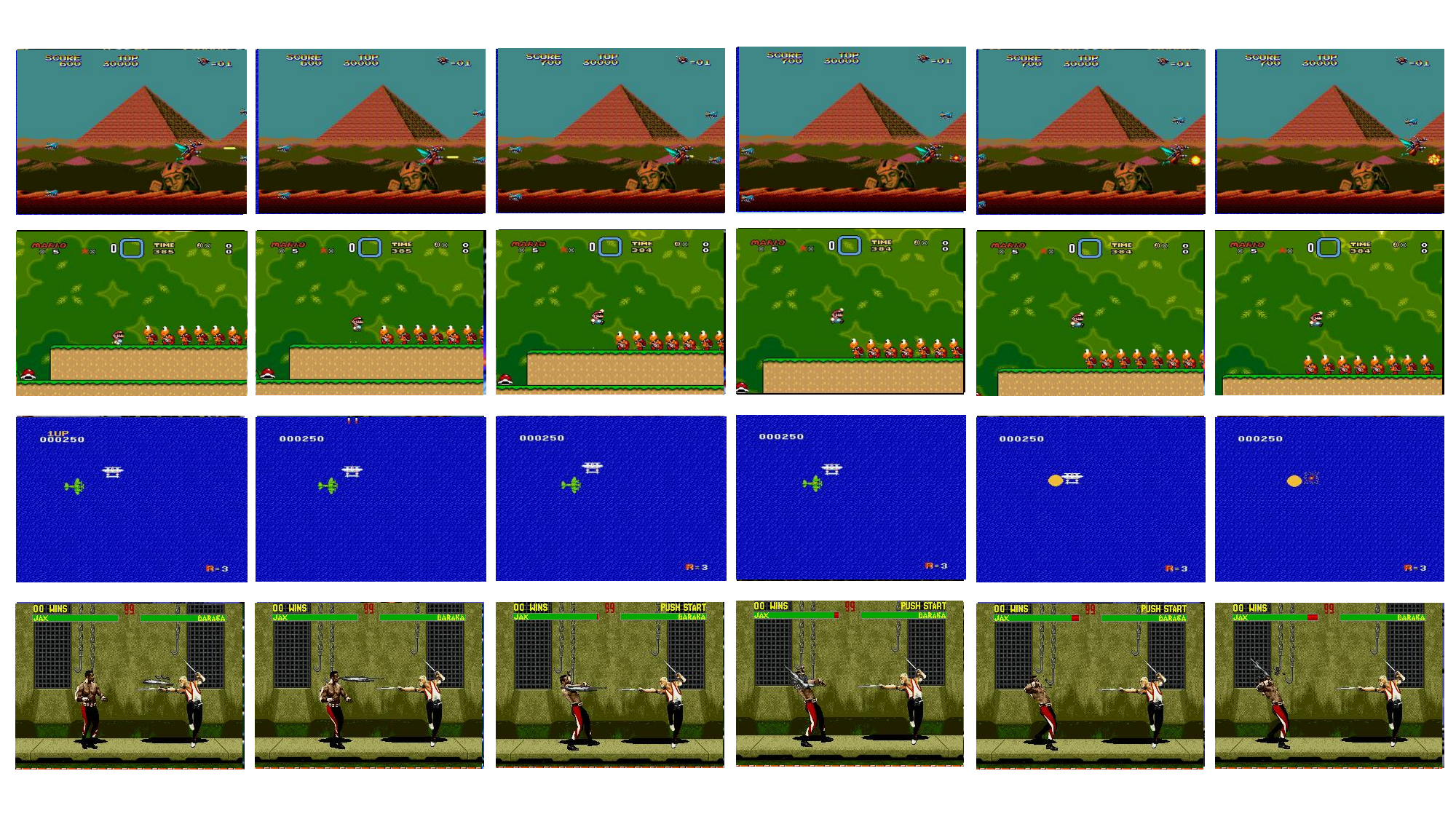}
\caption{
\textbf{GPT-5 Case Study.}
The figure illustrates four characteristic behaviors of the GPT-5 agent. (1) Line 1 shows that the agent demonstrates accurate target acquisition and offensive capability, intercepting an aerial enemy to clear the path; (2) Line 2 highlights its proficiency in precision platforming and spatial navigation, executing a calculated jump to skip the enemies; (3) Line 3 reveals limitation in situational awareness regarding trailing threats, where the agent fails to evade a collision with an enemy approaching from the rear; and (4) Line 4 indicates susceptibility to delayed reaction times in combat scenarios, resulting in a failure to dodge or block an incoming projectile attack.
}
    \label{fig:gpt5}
\end{figure*}

\begin{figure*}
    \centering
    \includegraphics[width=0.85\linewidth]{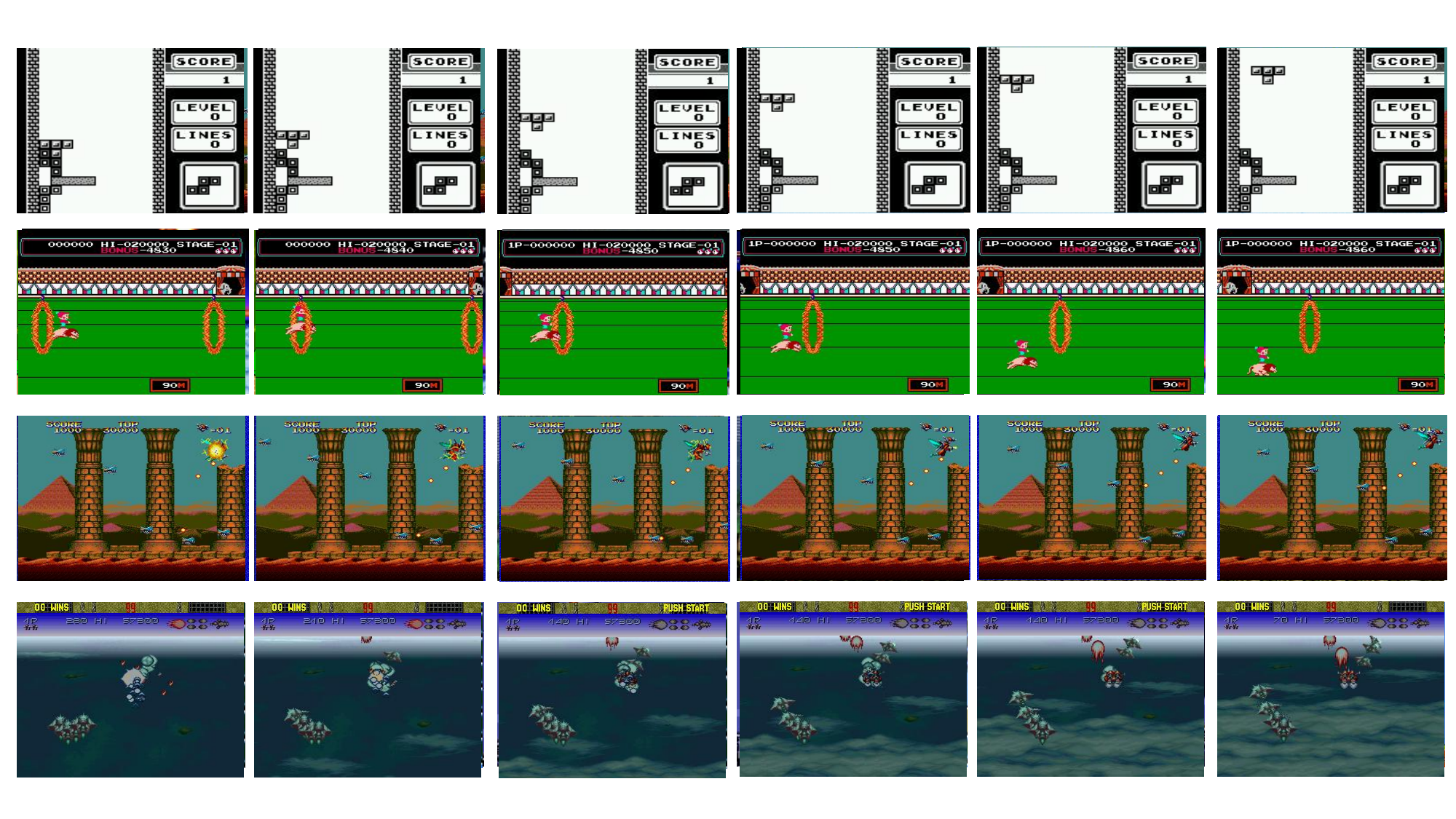}
\caption{
\textbf{Qwen3-VL-30B-A3B Case Study.}
The figure illustrates four representative behaviors of the Qwen3-VL-30B-A3B agent: (1) Line 1 shows that the agent demonstrates spatial reasoning and planning, rotating and tucking the tetromino into a precise gap to maintain clean board; (2) Line 2 highlights its proficiency in high-frequency temporal control, executing a timed jump to pass the obstacle (the fire ring); (3) Line 3 reveals a limitation in processing dense visual clutter, where the agent fails to distinguish between terrain features and enemy projectiles, resulting in a fatal collision; and (4) Line 4 indicates a susceptibility to policy degradation in open environments, where the agent exhibits passive drifting behavior rather than actively targeting enemies or dodging incoming formations.
}
    \label{fig:qwen3}
\end{figure*}

Large Vision-Language Models (VLMs), such as GPT-4o (Figure~\ref{fig:gpt4o}), GPT-5 (Figure~\ref{fig:gpt5}), and Qwen3-VL-30B-A3B (Figure~\ref{fig:qwen3}), introduce strong semantic reasoning capabilities in the control loop. These agents demonstrate proficiency in spatial navigation and target acquisition, successfully executing calculated jumps and neutralizing aerial threats through accurate planning. Despite these strategic strengths, they struggle with real-time situational awareness and reaction latency. The case studies reveal persistent weakness in handling fast-moving dynamic entities, particularly those approaching from behind or requiring rapid reflexes. This suggests that despite competent reasoning, a perception–action latency gap undermines their performance in high-speed adversarial settings.

\subsection{Interactive Physical Reasoner}
\begin{figure*}
    \centering
    \includegraphics[width=0.85\linewidth]{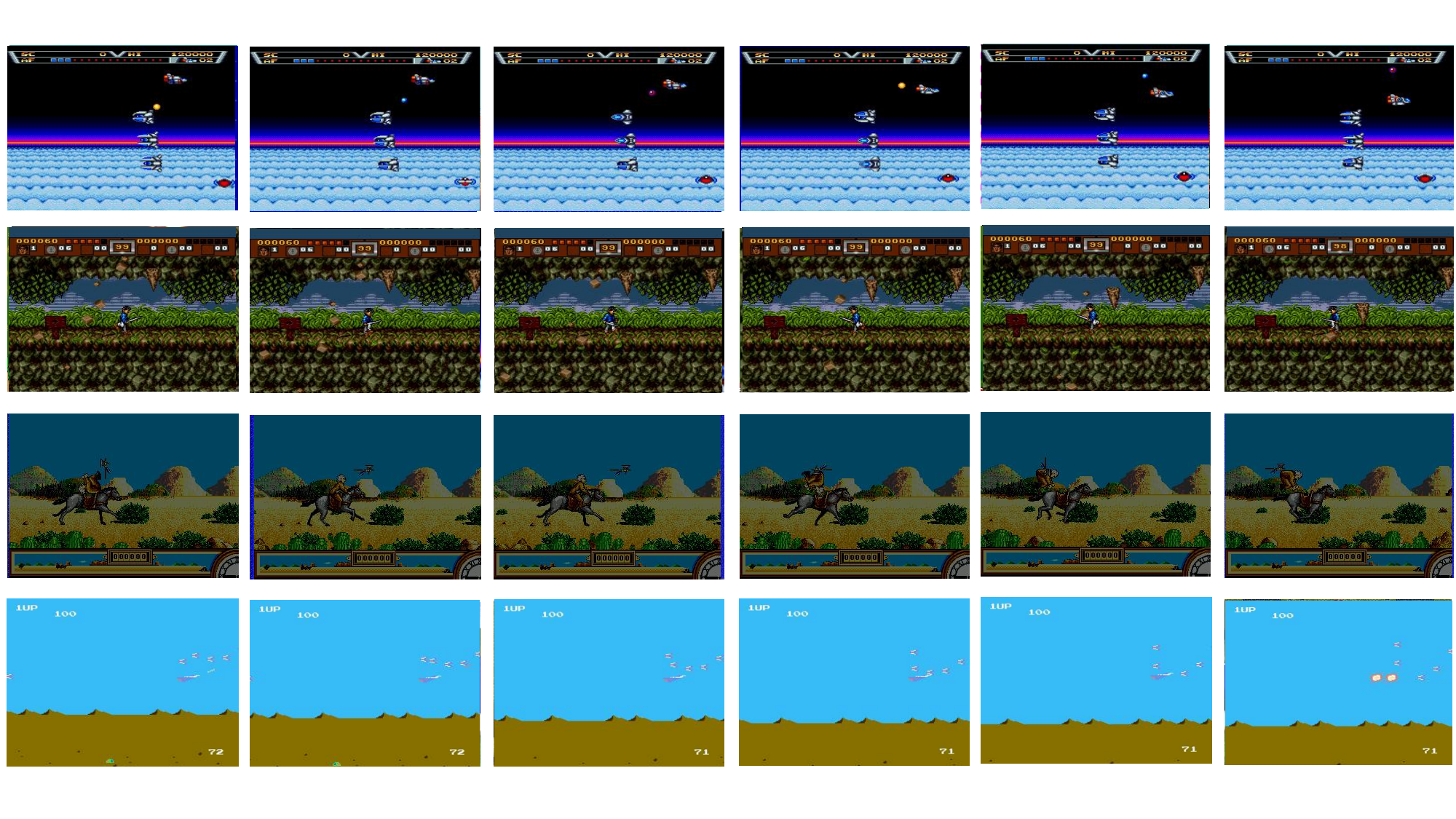}
\caption{
\textbf{IPR Case Study.} The figure illustrates four representative behaviors of the IPR agent: (1) Line 1 shows that the agent demonstrates precise reactive control, maneuvering to evade incoming projectiles; (2) Line 2 highlights its proficiency in dynamic environmental perception, allowing it to anticipate and dodge falling hazards (rocks); (3) Line 3 reveals vulnerability in rapid collision avoidance, where the agent fails to react to aerial threats (bats) (4) Line 4 indicates limitation in handling high-density adversarial environments, leading to an inability to evade when confronted by multiple simultaneous enemies.
}
    \label{fig:ipr}
\end{figure*}

The Interactive Physical Reasoner (IPR) (Figure~\ref{fig:ipr}) agent distinguishes itself by predictive imagination. Unlike the purely reactive RL agents or the prediction-based models, the IPR agent can simulate the trajectories of falling hazards and incoming projectiles, allowing for precise evasion and dynamic maneuvering. However, this imaginative capability is computationally constrained. While effective in one-on-one interactions, the agent reveals vulnerabilities in high-density adversarial environments. When the visual scene becomes cluttered with multiple simultaneous threats, the agent's capacity to ``imagine" can lead to failure. Although there are limitations, IPR can reach high performance through its imagination ability and interactive physical reasoning.

\end{document}